\documentclass[10pt,journal,compsoc]{IEEEtran}

\ifCLASSOPTIONcompsoc
  \usepackage[nocompress]{cite}
\else
  \usepackage{cite}
\fi

\ifCLASSINFOpdf
\else
\fi

\usepackage{stfloats}

\usepackage[table,xcdraw]{xcolor}

\usepackage{amsmath,amssymb,amsfonts}
\usepackage{graphicx}
\usepackage{textcomp}
\usepackage{tabularx}
\usepackage{makecell}
\usepackage{pgfplots}
\usepackage{multirow}
\usepackage{amssymb}
\usepackage{pifont}
\newcommand{\cmark}{\ding{51}}
\newcommand{\xmark}{\ding{55}}
\usepackage{hyperref}
\usepackage{soul}
\usepackage{subcaption}

\usepackage{changepage}
\usepackage{pdflscape}
\usepackage{comment} 
\usepackage{multicol}
\usepackage{algorithm}
\usepackage[noend]{algpseudocode}
\usepackage{longtable}
\usepackage{adjustbox}
\usepackage{arydshln}
\usepackage{lipsum}
\usepackage{supertabular}

\pgfplotsset{compat=1.11,
    /pgfplots/ybar legend/.style={
    /pgfplots/legend image code/.code={%
       \draw[##1,/tikz/.cd,yshift=-0.25em]
        (0cm,0cm) rectangle (3pt,0.8em);},
   },
}

\pgfplotsset{
    compat=1.11,
    legend image code/.code={
    \draw[mark repeat=2,mark phase=2]
    plot coordinates {
        (0cm,0cm)
        (0.1cm,0cm)        %
        (0.2cm,0cm)         %
        };%
    }
}

\def\BibTeX{{\rm B\kern-.05em{\sc i\kern-.025em b}\kern-.08em
    T\kern-.1667em\lower.7ex\hbox{E}\kern-.125emX}}

\algnewcommand{\IfThenElse}[3]{%
  \State \algorithmicif\ #1\ \algorithmicthen\ #2\ \algorithmicelse\ #3}
\algnewcommand{\IfThen}[3]{%
  \State \algorithmicif\ #1\ \algorithmicthen\ #2\ }

\makeatletter
\newcommand*{\addFileDependency}[1]{%
  \typeout{(#1)}
  \@addtofilelist{#1}
  \IfFileExists{#1}{}{\typeout{No file #1.}}
}
\makeatother

\begin{document}
\title{Effectiveness of Moving Target Defenses for Adversarial Attacks in ML-based Malware Detection}

\author{Aqib~Rashid, Jose~Such%
\IEEEcompsocitemizethanks{\IEEEcompsocthanksitem The authors are with the Department of Informatics, King's College London, Strand, London WC2R 2LS, United Kingdom.\protect\\
E-mail: \{aqib.rashid, jose.such\}@kcl.ac.uk}%
}

\IEEEtitleabstractindextext{%

\begin{abstract}
Several moving target defenses (MTDs) to counter adversarial ML attacks have been proposed in recent years. MTDs claim to increase the difficulty for the attacker in conducting attacks by regularly changing certain elements of the defense, such as cycling through configurations. To examine these claims, we study for the first time the effectiveness of several recent MTDs for adversarial ML attacks applied to the malware detection domain. Under different threat models, we show that transferability and query attack strategies can achieve high levels of evasion against these defenses through existing and novel attack strategies across Android and Windows. We also show that fingerprinting and reconnaissance are possible and demonstrate how attackers may obtain critical defense hyperparameters as well as information about how predictions are produced. Based on our findings, we present key recommendations for future work on the development of effective MTDs for adversarial attacks in ML-based malware detection.
\end{abstract}

}

\maketitle

\IEEEdisplaynontitleabstractindextext

\IEEEpeerreviewmaketitle

\vspace{-2mm}
\IEEEraisesectionheading{\section{Introduction}\label{sec:introduction}}

\vspace{-2.5mm}
ML models offer undeniable advantages in several domains  \cite{he2015delving,chio2018machine}, especially in malware detection, where they can classify benign and malicious executables. However, these models are vulnerable to adversarial attacks \cite{szegedy2013intriguing,papernot2017practical,rosenberg2018generic}, which exploit the faults of ML algorithms to exert control over the predictions. In particular, \emph{evasion attacks} are performed by generating an adversarial example that crosses the decision boundary to \emph{evade} the classifier \cite{szegedy2013intriguing,papernot2017practical}, where an attacker can have a malware sample predicted as benign \cite{chakraborty2018adversarial}. Thus, defensive measures are needed to counter such attacks. Recent work has proposed several feature-based, gradient-based, and randomization-based defenses %
(e.g., \cite{chakraborty2018adversarial,szegedy2013intriguing,papernot2016distillation, tramer2017ensemble, xu2017feature, wang2016random, xie2017mitigating, sengupta2018mtdeep, rosenberg2021adversarial}). However, many of these have been proven ineffective for dealing with adversarial attacks in several studies \cite{carlini2019evaluating,athalye2018obfuscated}, including in the malware detection domain \cite{pods}. 

Recently, moving target defenses (MTDs) have been proposed as a remedy \cite{rashid2022stratdef,sengupta2018mtdeep, qian2020ei, amich2021morphence,song2019moving, cho2020toward}. MTDs use several ML models (rather than a single model) in a specific manner to defend against adversarial attacks. They aim to make attackers less effective at reconnaissance and targeted attacks by regularly moving the defense (e.g., constituent models, how predictions are produced) \cite{cho2020toward}. MTDs belong to the family of ensemble defenses, which aim to outperform single models in terms of robustness, increased complexity for the attacker, variance reduction, prediction accuracy, and generalization \cite{lecun2015deep}. Therefore, several MTDs for adversarial ML have been proposed, using techniques such as increasing the heterogeneity of constituent models, diversity training of constituent models, dynamically regenerating the models, query-limiting, and strategically choosing models \cite{rashid2022stratdef,amich2021morphence,qian2020ei,song2019moving,sengupta2018mtdeep,yang2020dverge,pang2019improving,kariyappa2019improving}. 

However, prior to our work, the performance of MTDs against adversarial ML attacks under different threat models has not been evaluated, nor compared with each other, nor extensively compared with other defenses. It is important to assess whether MTDs may be a line worth investigating further to remedy adversarial ML attacks, and if so, what kinds of strategies and ways to design MTDs seem more effective.
In this paper, we present the first evaluation of several recent MTDs applied to the ML-based malware detection domain. To conduct this evaluation, we use transferability and query attack strategies from prior work as well as novel ones that we propose to maximize the evasion of MTDs. Additionally, we offer methods to conduct fingerprinting and reconnaissance to increase the understanding of how the target MTD works to enhance attacks, with minimal knowledge about it initially. Based on our evaluation, we offer recommendations for developing effective future MTDs. The main contributions of our work are:

1) We conduct the first evaluation of several MTDs to defend against adversarial ML attacks applied to the malware detection domain. Across Android and Windows, we show that the MTDs can be evaded with minimal information.

2) We examine the performance of MTDs using existing transferability and query attack strategies as well as novel improvements to these strategies for maximizing the evasion of MTDs. Our novel strategies increase the evasion rate by up to 50\% versus prior attack strategies.

3) We show that it may be possible to fingerprint and recognize MTDs with a set of initial techniques that allow us to discover the predictive nature of the MTDs studied and, in some cases, some of their critical hyperparameters.%

4) Informed by the evidence produced in our evaluation, we derive key insights and make crucial recommendations to help design more effective MTDs against adversarial ML.%

The paper is organized as follows. Sections~\ref{sec:background} and \ref{sec:threatmodel} provide the background and threat models considered. Section~\ref{sec:attackstrategy} details the attack strategies used. %
Section \ref{sec:expsetup} details the experimental setting. Sections \ref{sec:blackboxeval}-\ref{sec:otherfindings} report our results. Section~\ref{sec:discussion} synthesizes our findings and offers recommendations for building MTDs based on them. Section \ref{sec:relatedwork} discusses related work, and we conclude in Section~\ref{sec:conclusion}.

\vspace{-3mm}
\section{Background}
\label{sec:background}
\vspace{-1.5mm}
\noindent{\textbf{ML-based Malware Detection}} offers several advantages over signature-based detection methods, such as rapid assimilation of large datasets and the ability to generalize to unknown threats \cite{rosenberg2018generic}. Typically, deep neural networks (DNNs) are trained on binary feature vectors representing \emph{benign} (i.e., \emph{goodware}) and \emph{malware} software executables. For developing software representations that can be fed into ML models, \emph{feature extraction} parses an executable into a binary feature vector. The quality of ML models depends on the features used during training \cite{grosse2016, al2018adversarial, rosenberg2018generic}. However, adversarial ML has increased the attack surface \cite{szegedy2013intriguing}.

\noindent{\textbf{Adversarial ML.}} 
Our work concerns a type of adversarial ML attack called \emph{evasion attacks}, where an attacker perturbs the features of an input sample to obtain a specific prediction from the model \cite{papernot2018sok}, meaning that a malware sample is predicted as benign. An adversarial example is a perturbed version of the feature vector representing the executable. Even if the attacker does not have direct access to the target model, an attack can still be performed. Due to \emph{transferability}, adversarial examples developed for one model may evade other models because of weaknesses shared by classifiers \cite{szegedy2013intriguing}. In a \emph{transferability attack}, the attacker relies on the transferability property of adversarial examples to hold between the target model and a substitute model, which is an estimation of the target model for which a single DNN is commonly used \cite{papernot2017practical, rosenberg2018generic, brendel2017decision}. Meanwhile, \emph{query attacks} do not use substitute models \cite{chen2020stateful, brendel2017decision, chen2020hopskipjumpattack, rosenberg2020query, ilyas2018black, li2020qeba, pierazzi2020problemspace, Bhagoji_2018_ECCV} but instead generate adversarial examples by iteratively perturbing a malware sample based on queries to the target model. In other domains, techniques for this include gradient and decision boundary estimation \mbox{\cite{ilyas2018black, brendel2017decision,chen2020hopskipjumpattack}}. However, these techniques do not cater to the constraints of the malware detection domain regarding discrete features or functionality preservation. Instead, \emph{software transplantation-based approaches} \mbox{\cite{rosenberg2020query, pierazzi2020problemspace, yang2017malware}} can be used for query attacks. This involves perturbing a malware sample with benign ``donor'' features.%

\noindent{\textbf{Defenses Against Adversarial ML.}} Numerous defensive approaches for adversarial ML attacks have been proposed. These include gradient-based \cite{papernot2016distillation, tramer2017ensemble}, feature-based \cite{xu2017feature, wang2016random} and randomization-based \cite{xie2017mitigating, sengupta2018mtdeep} approaches, as well as techniques like (ensemble) adversarial training \cite{szegedy2013intriguing, tramer2017ensemble}. Most approaches are typically single-model defenses, where one model is made robust \cite{pang2019improving}. However, several recent surveys \cite{carlini2019evaluating,athalye2018obfuscated, pods} have found these defenses to be ineffective. 

\noindent{\textbf{Moving Target Defenses (MTDs)}} have been deployed in different security areas \mbox{\cite{cho2020toward, paruchuri2008playing, tambe2011security, Ward2018SurveyOC, DBLP:series/ais/100, 6673500}} and applied in various fields, such as industrial control systems \mbox{\cite{8085954}}, network intrusion detection systems \mbox{\cite{10.1145/2995272.2995281, 6623770}}, distributed systems \mbox{\cite{10.1145/2995272.2995283}}, web applications \mbox{\cite{10.5555/3091125.3091155}}, and cloud security \mbox{\cite{Bardas2017MTDCM}}. MTDs regularly change their configuration and move system components to increase uncertainty and make it more difficult to understand their behavior \mbox{\cite{cho2020toward}}. 

\noindent{\textbf{MTDs Against Adversarial ML Attacks. }}
MTDs have been proposed to defend against adversarial ML attacks \mbox{\cite{cho2020toward, dasgupta2019survey,rashid2022stratdef, sengupta2018mtdeep, qian2020ei, song2019moving, amich2021morphence, additional1, additional2}} by using several ML models in a specific manner to produce a prediction. In this setting, MTDs %
claim to boost prediction accuracy, robustness, generalization, and to reduce variance compared with other defenses through techniques that ``move'' the defense's configuration. Techniques include model regeneration \cite{song2019moving, amich2021morphence}, game-theoretic movement strategies \cite{sengupta2018mtdeep, rashid2022stratdef, additional1}, using adversarially-trained student models with different model selections \cite{qian2020ei, amich2021morphence} and combinatorially-boosted ensembles \cite{additional2}. Within the context of adversarial ML, MTDs can be categorized as either dynamic or hybrid. For an input sample \(X\), a dynamic MTD may return the prediction \(y\) at first, before returning the prediction \(y'\) later, such that \(y \neq y'\). Meanwhile, a hybrid MTD continues to return the same prediction \(y\) until a specific condition is reached, causing it to alter itself (e.g., regenerating models if a query budget is reached). Then, the prediction for \(X\) may be \(y'\) such that \(y \neq y'\). %
Contrarily, other ensemble defenses (such as voting \cite{shahzad2013comparative}) are static, where the same prediction for an input is returned. The non-static predictive nature of MTDs introduces yet another layer of complexity for attackers that is absent in other defenses.

\noindent{\textbf{Novelty \& Contributions.}} While prior work has shown the ineffectiveness of single-model defenses for adversarial ML \cite{pods,shahzad2013comparative,athalye2018obfuscated,papernot2016limitations,papernot2018sok,goodfellow2014explaining}, MTDs for adversarial ML have not been widely evaluated, nor compared with each other, nor extensively compared with other defenses. 
It is essential to determine if MTDs are a worthwhile research direction to protect against adversarial ML attacks and, if so, which approaches for designing MTDs appear more effective. Therefore, in this paper, we present the first evaluation of several MTDs using a range of attacks and threat models.

\vspace{-2.5mm}
\section{Threat Model}
\label{sec:threatmodel}
\vspace{-1.5mm}
In our work, attackers aim to evade a feature-based ML model to have malware samples predicted as benign.

\begin{figure}[!ht]
\vspace{-3mm}
\centering
    \scalebox{0.55}{
        \includegraphics{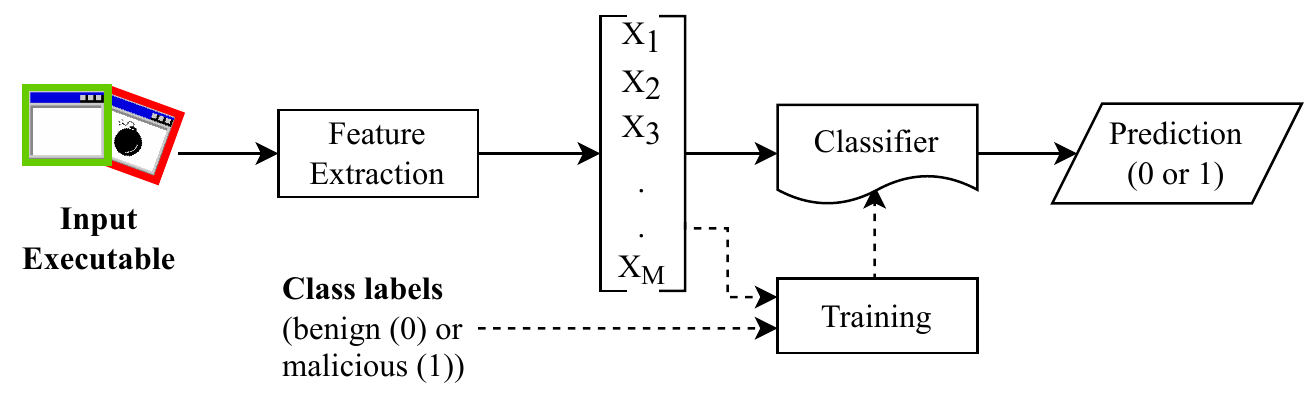}
        
    }
\vspace{-2.5mm}
\caption{Overview of a malware detection classifier.}
  \label{figure:malwareclassifierschema}
\vspace{-3mm}
\end{figure}

\noindent\textbf{Target Model.} The target model is an MTD with several constituent ML models. To train a constituent model, software executables are represented as binary feature vectors based on raw extracted features. Following prior work on ML-based malware detection \cite{rosenberg2020query, grosse2017adversarial, li2021framework}, the features we use include the libraries, API calls, permissions, or network addresses used, among others, which  are provided by the datasets we use,  detailed in Section~\ref{sec:expsetup}. %
With the features \(1...M\), a vector \(X\) can be constructed for each input sample such that \(X \in \{{0,1}\}^M \). Here, \(X_i = 1\) or \(X_i = 0\) indicates the presence or absence of feature \(i\) respectively. The feature vectors and their associated class labels are used to train the constituent binary classification models, as shown in Figure~\ref{figure:malwareclassifierschema}. The constituent models are then used in an MTD known as the \emph{oracle} \(O\). A user requests a prediction from \(O\), which uses the constituent models in a specific manner to return a prediction (e.g., a probability distribution to \emph{move} between, i.e., choose, the constituent models \cite{sengupta2018mtdeep}). The oracle provides scoreless feedback \cite{ilyas2018black}, only returning the prediction to limit the user's understanding of its behavior.

\noindent{\textbf{Attacker's Goal.}}
The attacker's goal is to generate an adversarial example \(X'\) to evade \(O\). Suppose \(O\colon X \in \{{0,1}\}^M \) and we have some function \(check()\) to check the functionality of an input sample. This goal can then be summarized as:
\begin{equation}
    check(X) = check(X'); O(X) = 1; O(X') = 0
\vspace{-2mm}
\end{equation}

\noindent{\(X'\)} must be functionally-equivalent to the original sample \(X\) but result in a \emph{benign} prediction from \(O\). For predictions, 1 represents the malware class and 0 the benign class.

\noindent{\textbf{Attacker Capabilities \& Knowledge.}} We model two types of attackers with different levels of knowledge, as commonly featured in prior work \cite{ilyas2018black, laskov2014practical, papernot2018sok}. Neither attacker knows that the target model is an MTD. The limited-knowledge \emph{gray-box attacker} has access to the same training data as the target model and has knowledge of the feature representation as well as the statistical representation of the features across the dataset. However, they have no knowledge of the parameters, configurations, or constituent models of the target model. This represents when some sensitive model information may have been leaked. Therefore, for transferability attacks, they train substitute models using the training data and attack them with the aim of generating adversarial examples that transfer to the oracle \cite{szegedy2013intriguing, laskov2014practical, papernot2017practical}. To conduct query attacks, the gray-box attacker uses their knowledge of the features to apply suitable perturbations using a software transplantation-based approach in a heuristically-driven manner. Meanwhile, the \emph{black-box attacker} can only observe the predicted outputs for their queries. They have no knowledge about the target system but have some information pertaining to the kind of feature extraction performed (e.g., the static analysis that a malware detection classifier may consider). %
Therefore, they conduct a chosen-plaintext attack \cite{chakraborty2018adversarial} for transferability attacks. This is achieved by querying the oracle sufficiently to train a substitute model, which is an estimation of the oracle, and then attacking it to generate adversarial examples that transfer to the oracle \cite{szegedy2013intriguing, laskov2014practical, papernot2017practical} (unlike the gray-box attacker, who can use the training data). For query attacks, the black-box attacker also uses a transplantation-based approach but without any extra information to guide the attack \cite{rosenberg2020query, pierazzi2020problemspace}.

\vspace{-2mm}
\section{Attack Strategies}
\label{sec:attackstrategy}
\vspace{-1mm}

\subsection{Transferability Attacks}
\label{sec:transfattack}
\vspace{-1.5mm}
In transferability attacks, attackers construct substitute models that are approximations of the oracle. Adversarial examples are then generated for these substitute models, in anticipation that they will transfer to the oracle \mbox{\cite{goodfellow2014explaining, papernot2017practical, papernot2016limitations, szegedy2013intriguing, rosenberg2018generic, brendel2017decision,liu2016delving}}. To evaluate MTDs against transferability attacks, we use practical strategies from prior work as well as our own. Attack strategies from prior work are included as baselines %
and to show MTDs' performance with already existing, general strategies that are not tailored to MTDs. These include the Single DNN strategy \mbox{\cite{papernot2017practical}}, where a single DNN is used as the substitute model \mbox{\cite{papernot2017practical, rosenberg2018generic, brendel2017decision}}, as well as the Ensemble DNN strategy \mbox{\cite{liu2016delving}}, where several DNNs are used as the substitute models. 

Additionally, we propose a novel attack strategy that specifically considers that the target model \emph{may be} an MTD. %
For this, we present two main improvements on previous strategies. First, our transferability attack strategy utilizes an ensemble of diverse substitute models (including different ML families). Second, as a novel technique to increase attack success against MTDs, our attack strategy aims to maximize the transferability of adversarial examples across substitute models prior to evaluating them on the oracle by checking transferability across (part of) the substitute models constructed. We include these two additions because MTDs may change the model used for predictions dynamically, and so ensuring a degree of transferability of adversarial examples between substitute models maximizes success on the oracle, as we confirm later on experimentally.

\begin{figure}[!htbp]
\vspace{-2.5mm}
\centering
    \scalebox{0.8}{
        \includegraphics{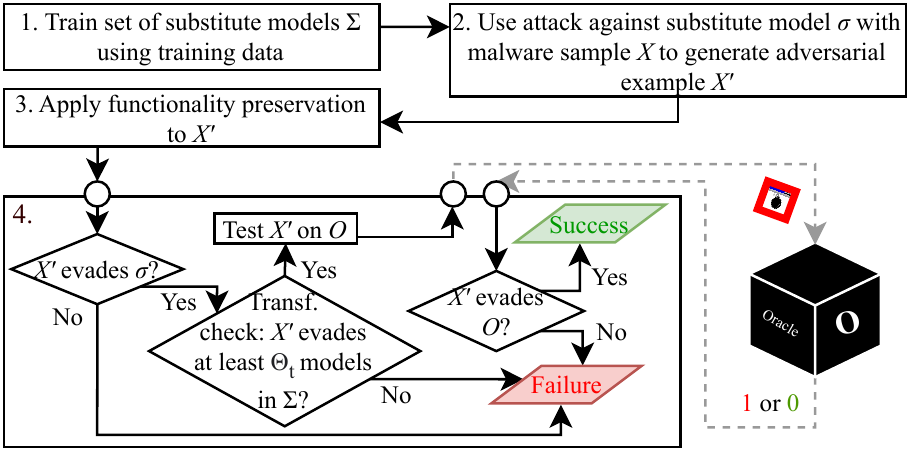}
   }
\vspace{-2.5mm}
    \caption{Overview of our transferability attack strategy.}
  \label{figure:birdattackpipeline}
\vspace{-2.5mm}
\end{figure}

Figure~\ref{figure:birdattackpipeline} summarizes the key steps of our transferability attack strategy, with each step described in detail next:

\noindent{\textbf{Step 1) Developing Substitute Models.}} The aim here is to train substitute models that are an estimation of \(O\) (the oracle). It is well-established that a model can be evaded by utilizing different model architectures and a dataset independent of it \cite{szegedy2013intriguing}. As detailed in Section~\ref{sec:expsetup}, we particularly favor an ensemble of substitute models as diverse as possible to increase the chances of finding highly-transferable adversarial examples that will transfer to the oracle. %

For each attacker, the procedure for constructing the substitute models differs due to their capabilities. 
The \emph{black-box attacker} has no capabilities beyond observing the predictions for their queries. Therefore, under this scenario, we develop a synthetic dataset (\(\Delta\)) that is an estimation of the input-output relations of the oracle. To develop \(\Delta\), we use a set \(B_{train}\) (containing benign and malware input samples) and query \(O\) with each input sample \(X \in B_{train}\), recording the prediction. For example, if \(O(X) = 0\), then the input-output relation \(X \mapsto 0\) is stored in \(\Delta\). \(\Delta\) is then used to train the ensemble of diverse substitute models (\(\Sigma\)) for our attack strategy. This provides white-box access to models that are an estimation of the oracle. For the other attack strategies that we test (that is, the Single DNN \cite{papernot2017practical} and Ensemble DNN \cite{liu2016delving} strategies), \(\Delta\) is used to train the substitute model(s) according to their procedures. Prior work in other domains has suggested that substitute models require a large training dataset \cite{papernot2017practical} and, while unrealistic, an attacker could create a replica of the oracle with infinite queries. However, we show later that a high evasion rate can be achieved with a small set of input samples (i.e., \(|B_{train}| \leq 100\)). With fewer queries to the oracle, an attacker can remain stealthier, with lower chances of their behavior being detected \cite{chen2020stateful,grosse2017statistical}.

Conversely, the \emph{gray-box attacker} has access to the training data of the defense. Therefore, we train an ensemble of diverse substitute models (\(\Sigma\)) using the available training data. The advantage is that no queries to the oracle are necessary, reducing the time cost and the risk of adversarial behavior being detected. However, as we demonstrate later in Section~\ref{sec:grayboxeval}, using the training data of the target models may not always lead to the best attack performance. Because they are based on direct queries to each oracle, the substitute models developed by the black-box attacker may reflect their characteristics and behavior better. Also, given that the gray-box attacker has knowledge of the dataset, an alternative technique is to use universal adversarial perturbations (UAPs) \cite{moosavi2017universal, labaca2021universal}. With UAPs, the same sets of perturbations are reused and applied to several malware samples to generate adversarial examples. However, we show later in Section~\ref{sec:grayboxeval} that this attack approach is less successful.%

\noindent{\textbf{Step 2) Generating an Adversarial Example.}} As usual in all transferability attacks, once the substitute models have been constructed, we have full access to them. Then, we can use an attack against a substitute model with existing white-box attacks from the literature to generate an adversarial example \(X'\) from a malware sample \(X\) (e.g., FGSM \cite{goodfellow2014explaining}, JSMA \cite{papernot2016limitations}--- see Section~\ref{sec:expsetup} for the full list of white-box attacks used in the evaluation). The idea is that this adversarial example is likely to \emph{transfer} to the target model (the oracle). %

\noindent{\textbf{Step 3) Applying Functionality Preservation.}} We then %
validate the perturbations that have been used to generate \(X'\). This is done in order to preserve the original functionality of \(X\) within the feature space as a lower bound, which is essential in this domain. Otherwise, one may have an adversarial example that evades the oracle but has lost its functionality. In this process, any invalid perturbations found are reverted to their original values. %
Note that the perturbations that are valid or invalid and the way in which functionality preservation works in practice depend on the particular target platform for the malware. We detail valid and invalid perturbations for each of the cases we explore in the evaluation (Android and Windows) later in Section~\ref{sec:expsetup}.

\noindent{\textbf{Step 4) Transferability Across Substitutes \& Evaluation.}} We then ensure that the adversarial example \(X'\) still evades the substitute model used to build it, as the process for validating the perturbations may have reversed some perturbations used to cross the decision boundary.
If so, after this, we include a crucial new step where we look at how successful \(X'\) is in evading the other substitute models (i.e., a local transferability check is performed before actually testing \(X'\) on the oracle \(O\)). 
As the attack success relies on the transferability property holding between \(\Sigma\) and \(O\), our hypothesis is that adversarial examples that transfer better across all \(\sigma\) in the ensemble of substitute models \(\Sigma\) are more likely to evade \(O\). For this, we check whether \(X'\) transfers across a proportion of substitute models (\(\Theta_{t}\)) before testing it on \(O\). For example, \(\Theta_{t} = 0.75\) means that \(X'\) must evade 75\% of substitute models. %
If \(X'\) adequately transfers across the substitute models and then evades \(O\), it is counted as a success, whereas a failure is when \(O\) is not evaded by \(X'\).

\vspace{-2mm}
\subsection{Query Attacks}
\label{sec:queryattackstrategydetail}
\vspace{-1.5mm}

Query attacks generate adversarial examples by iteratively perturbing an input sample \cite{chen2020stateful, brendel2017decision, chen2020hopskipjumpattack, rosenberg2020query, ilyas2018black, li2020qeba,Bhagoji_2018_ECCV}, rather than using substitute models. Most query attacks, however, are designed for the image recognition domain and therefore perturb features continuously, meaning they are less effective in our domain due to its constraints, such as discrete features and functionality preservation. For example, as explained in \mbox{\cite{rosenberg2020query}}, a feature for an API call (e.g., $WriteFile()$) cannot be perturbed continuously (e.g., $WriteFile()+0.001$). For this, an entirely new feature is required, offering the same functionality.

To overcome these problems, we can use software transplantation-based approaches. This means that features from benign samples are used to perturb a malware sample (e.g., a feature is added to a malware sample), which can be conducted in a scenario with less \cite{rosenberg2020query, yang2017malware} or more \cite{pierazzi2020problemspace} information  about the target model. Overall, this allows malware samples to cross the decision boundary of the oracle while catering to the constraints of this domain. When conducting this attack, limiting the number of queries to the oracle is critical, as adversarial behavior can be detected when analyzing queries for abnormalities \cite{chen2020stateful}. Moreover, some MTDs use query budgets, which may hinder the construction of adversarial examples. Hence, we use the parameter \(n_{max}\) to govern the maximum number of allowed queries. We offer two approaches for performing query attacks in black-box and gray-box scenarios.

\noindent{\textbf{Black-box Query Attack.}} Under the black-box scenario, the attacker has no knowledge of the target model besides the predicted output. This means that the attack is conducted in a non-heuristic manner, where randomly-chosen features are perturbed accordingly. Our black-box query attack strategy is inspired by Rosenberg et al.'s %
\mbox{\cite{rosenberg2020query}}. However, our threat model considers \emph{less information} about the target model under the black-box condition. We assume no access to prediction confidence scores or usage of sliding windows, so we use a modified version of their decision-based attack without these assumptions. Furthermore, if allowed by the dataset (see Section~\ref{sec:expsetup}), our black-box attack strategy makes use of both feature addition \emph{and} feature removal to increase the possible avenues for evasion. In our black-box query attack, a malware sample \(X\) is perturbed using features from a set of benign donor samples (\(\mathcal{B}\)) to generate an adversarial example \(X'\) using a transplantation-based approach. However, this is performed in a non-heuristic manner because of this attacker's weaker capabilities. This means that the feature to perturb during each iteration of the attack is chosen \emph{randomly}, which can be added to (0 to 1) or removed from (1 to 0) the feature vector (as permitted by the dataset). Feature removal can be performed when features absent in a benign sample (but perhaps functionally insignificant in \(X\)) are removed to cross the decision boundary. The attack process takes constraints regarding functionality preservation into consideration by only making perturbations that are allowed (see Section~\ref{sec:expsetup} later). The transplantation of the features continues until \(O\) is evaded, \(n_{max}\) is reached, or the possible features of the benign sample are exhausted (in which case another benign sample may be used).

\noindent{\textbf{Gray-box Query Attack.}} 
The gray-box attacker has access to the training data and knowledge of the statistical representation of the features across the dataset. Therefore, features can be added to a malware sample in a heuristically-driven manner using their frequency in benign samples. 

\begin{figure}[!htbp]
\vspace{-4mm}
\centering
    \scalebox{0.8}{
        \includegraphics{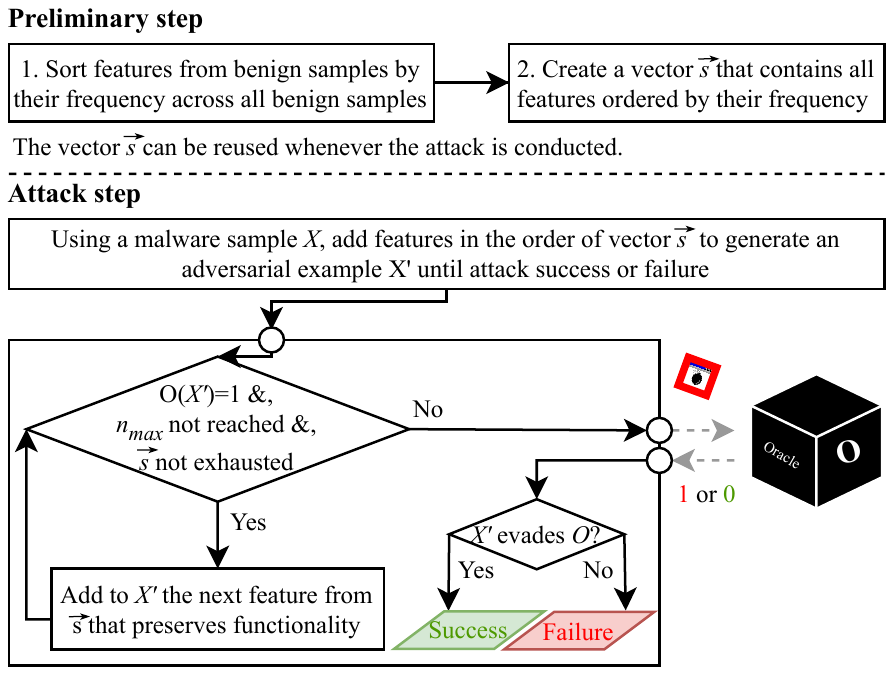}
   }
\vspace{-2mm}
    \caption{Overview of our gray-box query attack strategy.}
  \label{figure:grayboxattackfigu}
\vspace{-3mm}
\end{figure}

In other words, features that are more frequent in benign samples are added to the malware sample \(X\) before others to promote traversal of the decision boundary as soon as possible. We show later that this significantly increases attack success and reduces queries to the oracle. Hence, as per Figure~\ref{figure:grayboxattackfigu}, from the data available to the attacker, the features from benign samples are sorted by their frequency across all benign samples into a vector \(\vec{s}\) as a preliminary step to the attack --- \(\vec{s}\) can be reused whenever the attack is conducted. Then, given the ordered vector of features \(\vec{s}\) (which contains features ordered by their frequency in benign samples), we can generate an adversarial example \(X'\). Using a malware sample \(X\), the next feature from \(\vec{s}\) that preserves the original functionality of \(X\) is added to generate \(X'\). Recall from before that only certain perturbations for each target platform will preserve the functionality of the malware sample (platform-specific details of valid perturbations in Section~\ref{sec:expsetup}). As before, perturbations are validated for functionality preservation before being tested on \(O\). The transplantation of features continues until the generated adversarial example \(X'\) evades \(O\), \(n_{max}\) is reached, or the possible features are exhausted.

\vspace{-3mm}
\section{Experimental Setup}
\label{sec:expsetup}
\vspace{-1.5mm}

\noindent{\textbf{Datasets.}} In many security applications, sampling from the true distribution is challenging and sometimes impossible~\cite{arp2022and, 8949524}. Particularly, the number of publicly-available, up-to-date datasets in our domain is a well-known problem, which limits the remits and conclusions of academic work in this domain~\cite{arp2022and, 8949524}. %
To partially mitigate this, we use three datasets that cover different platforms and collection methods that have been used previously (e.g., \mbox{\cite{wangrobust, grosse2017adversarial, 10.1145/3503463, labaca2021universal, pierazzi2020problemspace, 203684, barbero2022transcending, 10.5555/3361338.3361389}}): DREBIN for Android malware \mbox{\cite{arp2014drebin}}, SLEIPNIR for Windows malware \mbox{\cite{al2018adversarial}}, and AndroZoo for Android malware \mbox{\cite{Allix:2016:ACM:2901739.2903508}}. %

DREBIN has 123,453 benign samples and 5,560 malware samples, based on extracted static features. Hence, we use 5,560 samples from each class (benign and malware). SLEIPNIR consists of 19,696 benign samples and 34,994 malware samples. The features of this dataset are derived from Windows API calls in PE files parsed by the LIEF library\cite{LIEF} into a vector representation. SLEIPNIR is used to represent Windows out of simplicity, with a convenient binary feature-space, enabling a clearer comparison between the Android and Windows datasets as our work is in the feature-space. To keep the dataset balanced, we use 19,696 samples from each class. %
AndroZoo \mbox{\cite{Allix:2016:ACM:2901739.2903508}} is a large-scale dataset with timestamped Android apps from between January 2017 and December 2018. This dataset offers apps from different stores and markets, with VirusTotal summary reports for each. Similar to prior work \mbox{\cite{pierazzi2020problemspace, 10.5555/3361338.3361389, miller2016reviewer}}, we consider an app malicious if it has 4 or more VirusTotal detections, and benign if it has 0 VirusTotal detections (with apps that have 1-3 detections discarded). Thus, the final dataset contains $\approx$ 150K recent applications, with 135,859 benign apps and 15,778 malicious apps. We use 15,778 samples from each class (benign and malware).
As in recent publications \mbox{\cite{demontis2017yes, grosse2017adversarial}}, and for completeness, we use a large number of features for each dataset, i.e., 58,975 for DREBIN, 22,761 for SLEIPNIR, and 10,000 for AndroZoo. Note, however, that we provide an experiment in Appendix~\ref{appendix:lessfeaturesexp} with far fewer features for some datasets (500 features), suggesting that the number of features does not play a role in the results.

The datasets are split randomly for each class according to the Pareto principle, with an 80:20 ratio for training and test sets. Subsequently, using the same ratio, this training set is further partitioned into the final training and validation sets for constructing the models and defenses we evaluate. Effectively, this produces the 64:16:20 split that has been commonly used before (e.g., \cite{li2019few, yao2019hierarchically}). The validation set is used to tune the models during development. The remaining test set is used in the attacks and is further split into training, validation, and the final test set. The training set here is used in the black-box transferability attack and corresponds to \(B_{train}\), which is used to obtain the input-output relations of \(O\) for developing \(\Delta\). For DREBIN, \(|B_{train}| = 1423\). As SLEIPNIR and AndroZoo are larger, we perform additional splits to reduce \(|B_{train}|\) to 1513 and 1494 respectively. Meanwhile, the malware samples in the final test set are the input samples for transferability and query attacks. For DREBIN, SLEIPNIR, and AndroZoo, there are 229, 230 and 234 such samples respectively.%

We consider established guidelines for conducting malware experiments \mbox{\cite{6234405}}. For example, as the models in our evaluation decide whether an input sample is benign or malicious, retaining benign samples in the datasets is necessary. This also means that we do not need to strictly balance datasets over malware families. Instead, we balance datasets across the positive and negative classes, and randomly select unique samples from each class to appear in the training and test sets (without any chance of repetition) \mbox{\cite{pods, al2018adversarial, rosenberg2018generic}}.

\noindent{\textbf{Moving Target Defenses.}} We evaluate four MTDs that are configured to provide maximal robustness using the parameters suggested in their original papers. As we show in the next subsection, these MTDs offer sound performance in non-adversarial conditions for malware detection, so they are adequate for use in this domain. That is, these MTDs are domain-agnostic. We present the configuration for each defense in \mbox{Appendix~\ref{appendix:evaldef}}. \textbf{\emph{DeepMTD}} is a hybrid MTD \cite{song2019moving}. \(n\) student models are generated by perturbing weights of a base DNN, with \(w\) controlling the amount of perturbations. To make a prediction, if more than \(T\) x 100\% of the outputs from the student models are the same, the input sample is considered non-adversarial (and the majority class is the final prediction). Otherwise, it is classified as adversarial. To keep the system ``moving'', new models are generated when the system is idle. \textbf{\emph{Morphence}} is a hybrid MTD \cite{amich2021morphence}. It generates \(n\) student models from a base model by shuffling weights randomly. Then, \(p\) student models are adversarially-trained (where \(p \leq n\)). For a user's query, the prediction of the most confident student model is returned. Morphence uses a query budget; student models are regenerated if the number of queries exceeds \(Q_{max}\). \textbf{\emph{MTDeep}} \cite{sengupta2018mtdeep} is a dynamic MTD. This defense models the interactions between attackers and defenders as a Bayesian Stackelberg game to produce a strategy vector to choose models at prediction-time based on attack intensity. The resulting strategy can be pure (where only a single model is chosen) or mixed (where one of several models can be chosen). \textbf{\emph{StratDef}} \cite{rashid2022stratdef} is a dynamic MTD that provides a 
framework to systematically construct, select, and strategize a set of models. %
It gives particular consideration to the heterogeneity of its constituent models to minimize transferability, and it uses different optimizers to choose the best strategy for selecting a model at prediction time.

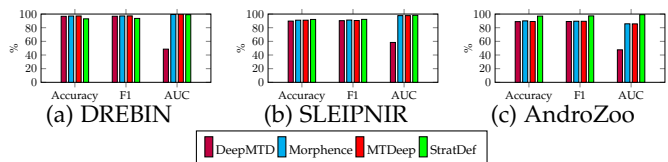
\begin{figure}[!htbp]
\vspace{-2.5mm}
\hspace{-0.25cm} 
\centering
\begin{subfigure}[b]{0.15\textwidth}
\begin{tikzpicture}[scale=0.45]
    \begin{axis}[
        ybar,
        enlarge y limits=false,
		bar width=0.15cm,
        ylabel={\%},
		ymin=0,
        ymax=100,
        symbolic x coords={Accuracy, F1, AUC},
        xtick=data,
        ytick={0,20,40,60,80,100},
        nodes near coords align={horizontal},
        legend pos=north east,
        legend style={nodes={scale=0.5, transform shape}},
        legend to name=mtdvaluesalphazero,
        legend columns=4,
		enlarge x limits={abs=1cm},
        width=6.6cm,
        ylabel style=
        {
            yshift=-2mm, 
        },
        height=3.6cm
        ]
    \addplot[fill=purple] coordinates {(Accuracy, 96.7) (F1, 96.8) (AUC, 48.6) };
    \addplot[fill=cyan] coordinates {(Accuracy, 97.1) (F1, 97.2) (AUC, 99.4) };
    \addplot[fill=red] coordinates {(Accuracy, 97.2) (F1, 97.3) (AUC, 99.8)};
    \addplot[fill=green] coordinates {(Accuracy, 93) (F1, 93.6) (AUC, 99.1) };
    
    \legend{DeepMTD, Morphence, MTDeep, StratDef}
    \end{axis}
   
\end{tikzpicture}
\vspace{-6.5mm}
\caption{DREBIN}
\end{subfigure}
\hspace{0.1cm}
\begin{subfigure}[b]{0.15\textwidth}
\begin{tikzpicture}[scale=0.45]
    \begin{axis}[
        ybar,
        enlarge y limits=false,
		bar width=0.15cm,
        ylabel={\%},
		ymin=0,
        ymax=100,
        symbolic x coords={Accuracy, F1, AUC},
        xtick=data,
        ytick={0,20,40,60,80,100},
        nodes near coords align={horizontal},
        legend pos=north east,
        legend style={nodes={scale=0.5, transform shape}},
        legend to name=mtdvaluesalphazero,
        legend columns=4,
		enlarge x limits={abs=1cm},
        width=6.6cm,
        ylabel style=
        {
            yshift=-2mm, 
        },
        height=3.6cm
        ]
    \addplot[fill=purple] coordinates {(Accuracy, 89.7) (F1, 90.3) (AUC, 58.4) };
    \addplot[fill=cyan] coordinates {(Accuracy, 91) (F1, 91.2) (AUC, 97.7)};
    \addplot[fill=red] coordinates {(Accuracy, 91) (F1, 90.5) (AUC, 97.7) };
    \addplot[fill=green] coordinates {(Accuracy, 92.1) (F1, 92.2) (AUC, 98.2)};
    
    \legend{DeepMTD, Morphence, MTDeep, StratDef}
    \end{axis}
   
\end{tikzpicture}
\vspace{-6.5mm}
\caption{SLEIPNIR}
\end{subfigure}
\hspace{0.1cm}
\begin{subfigure}[b]{0.15\textwidth}
\begin{tikzpicture}[scale=0.45]
    \begin{axis}[
        ybar,
        enlarge y limits=false,
		bar width=0.15cm,
        ylabel={\%},
		ymin=0,
        ymax=100,
        symbolic x coords={Accuracy, F1, AUC},
        xtick=data,
        ytick={0,20,40,60,80,100},
        nodes near coords align={horizontal},
        legend pos=north east,
        legend style={nodes={scale=0.5, transform shape}},
        legend to name=mtdvaluesalphazero,
        legend columns=4,
		enlarge x limits={abs=1cm},
            width=6.6cm,
            ylabel style=
            {
                yshift=-2mm, 
            },
        height=3.6cm
        ]
    \addplot[fill=purple] coordinates {(Accuracy, 88.9) (F1, 89.1) (AUC, 47.6) };
    \addplot[fill=cyan] coordinates {(Accuracy, 90) (F1, 89.5) (AUC, 85.7) };
    \addplot[fill=red] coordinates {(Accuracy, 89.1) (F1, 89.4) (AUC, 85.6)};
    \addplot[fill=green] coordinates {(Accuracy, 97) (F1, 97.3) (AUC, 99.1)};
    
    \legend{DeepMTD, Morphence, MTDeep, StratDef}
    \end{axis}
   
\end{tikzpicture}
\vspace{-6.5mm}
\caption{AndroZoo}
\end{subfigure}

\ref{mtdvaluesalphazero}

\vspace{-2.5mm}
\caption{Malware detection performance of evaluated MTDs.}
\label{figure:avgmtdvalues}
\vspace{-2.5mm}
\end{figure}

\noindent{\textbf{Performance of MTDs in Non-adversarial Conditions.}} Some MTDs in this paper have not been applied to the malware detection domain and instead have been first applied in the image recognition domain. However,  similar to defenses such as adversarial training (which was first applied to images and then to malware detection \mbox{\cite{grosse2017adversarial}}), MTDs can operate effectively across a range of domains. As we show in \mbox{Figure~\ref{figure:avgmtdvalues}}, the MTDs that we choose to evaluate offer sound predictions (90+\% accuracy) on test sets in the malware detection domain, thereby showcasing their ability to work well in this domain too. DeepMTD seems to offer moderate performance considering AUC, but we still choose to include it in the evaluation just in case this is a result of the mechanism used against adversarial ML attacks, so we can compare it with other MTDs. %
 
\vspace{-1mm}

\noindent{\textbf{Other Evaluated Defenses.}} As MTDs are a type of \emph{ensemble} defense --- that utilize several models --- we compare their performance with other types of defenses that use ensembles (in some manner) as baselines. \textbf{\emph{Voting}} is a static ensemble defense that has previously been applied to malware detection \cite{shahzad2013comparative, yerima2018droidfusion}. Multiple models are involved in making the prediction, which is determined by \emph{majority} or \emph{veto} voting. In majority voting, each constituent model makes a prediction, and the class with the most predictions is returned. In veto voting, if any constituent model returns a ``malware'' prediction, the entire system returns this prediction. %
\textbf{\emph{Ensemble adversarial training}} is a defense that trains a single model by using adversarial examples for other models \cite{tramer2017ensemble}. The adversarially-trained model is then used to produce predictions. We use an adversarially-trained neural network (NN-AT) that %
is constructed using the training data described previously. Then, it is boosted by training with a quantity of adversarial examples that is 25\% of the size of the training data, which is a sizeable amount to promote high adversarial robustness. Using a set of vanilla models (see \mbox{Appendix~\ref{appendix:arcsubmodels}}), we generate adversarial examples by applying a range of attacks (listed in the following section) and conducting the functionality preservation procedure (described later). A proportion of these adversarial examples are then used for adversarially-training the neural network model (NN-AT).

\noindent{\textbf{Substitute Models (Transferability Attack).}}
For our transferability attack (Section~\ref{sec:transfattack}), we use an ensemble of diverse substitute models to maximize evasion. With minimal tuning of hyperparameters to demonstrate the simplicity of our attack strategy, we construct four substitute models: a decision tree (DT), a neural network (NN), a random forest (RF) and a support vector machine (SVM) --- see Appendix~\ref{appendix:arcsubmodels} for architectures. We use the scikit-learn\cite{scikit-learn}, Keras\cite{chollet2015keras} and Tensorflow\cite{tensorflow2015-whitepaper} libraries for training. We use \(\Theta_{t} = 0.75\) to ensure that an adversarial example evades the majority of substitute models before testing it on the oracle. We also test other values in Sections~\ref{sec:blackboxeval} and \ref{sec:grayboxeval}. %

\noindent{\textbf{Generating Adversarial Examples.}} For the \emph{transferability attack} (Section~\ref{sec:transfattack}), we generate adversarial examples for substitute models using a variety of attacks: the Basic Iterative Method \cite{kurakin2016adversarial}, Decision Tree attack \cite{grosse2017statistical}, Fast Gradient Sign Method \cite{goodfellow2014explaining}, Jacobian Saliency Map Approach \cite{papernot2016limitations} and SVM attack \cite{grosse2017statistical}. These attacks produce continuous feature vectors and do not consider functionality preservation. %
Therefore, after applying these attacks, we round the values in the generated continuous feature vectors to produce discrete feature vectors, representing the presence or absence of a feature (e.g., usage of a particular library). For example, if an attack changes the value of a particular feature to \(\le 0.5\), it is set to 0 in the feature vector; meanwhile, if the value is \( \geq 0.5\), it is set to 1 in the feature vector. We then check for invalid perturbations to preserve functionality within the feature-space. Only after invalid perturbations are reverted does an adversarial example proceed further in the attack pipeline according to the attack strategy. 

For the \emph{query attack} (Section~\ref{sec:queryattackstrategydetail}), we apply the attack strategies under the black-box and gray-box scenarios. In both scenarios, a malware sample is perturbed by transplanting features from benign samples~\cite{rosenberg2020query, yang2017malware,pierazzi2020problemspace}. For example, if a particular feature is enabled in benign samples (i.e., its value is 1 in the feature vectors), it is added to the malware sample (changed from 0 to 1 in the feature vector for the malware sample) in order to move closer to crossing the decision boundary. The difference between the black-box and gray-box attack strategies lies in the choice of which features to perturb first. The gray-box attacker perturbs features based on their frequency in benign samples based on their knowledge of the dataset. Meanwhile, the black-box attacker chooses which features to perturb randomly, as in \cite{rosenberg2020query}, as no further information is available.

In both transferability and query  attacks, the \emph{permitted (valid) perturbations} (either feature addition or removal) for each dataset are determined by consulting with industry documentation, previous work \mbox{\cite{li2021framework, al2018adversarial, pierazzi2020problemspace, labaca2021universal}}, and the feature representation for each dataset. %
DREBIN and AndroZoo allow for both feature addition and removal \cite{li2021framework, 8171381} --- see \mbox{Appendix~\ref{appendix:allowedperturbationsdrebin}} for a summary of the allowed perturbations. In contrast, due to the encapsulation by LIEF's feature extraction process when developing the dataset, we can only perform feature addition on SLEIPNIR. This results in a more \emph{constrained attack surface} for SLEIPNIR, as there is a greater restriction on the perturbations that can be applied, leading to, as we will see later on, less effective attacks. The procedure we use offers a lower bound of functionality preservation within the feature-space, similar to prior work \mbox{\cite{severi2021explanation, grosse2017adversarial, li2021framework}}. While we remain in the feature-space, the perturbations we perform could be translated to the problem-space as well. For example, feature addition could be achieved by adding dead-code or by using opaque predicates \mbox{\cite{moser2007limits}}. Feature removal --- which is more complex but still achievable --- could be performed by rewriting the dexcode, encrypting API calls and network addresses (e.g., removing the features but retaining functionality).

\noindent{\textbf{Evaluation Metrics.}}
We use several metrics in our work, similar to previous work \cite{papernot2017practical}. The \emph{evasion rate} is defined as the number of adversarial examples that evade the oracle over the number of adversarial examples that evade the substitute models. Furthermore, as MTDs may employ different models at prediction-time, an adversarial example may not always evade the oracle. Therefore, we include the \emph{repeat evasion rate (RER)} as an additional metric for evaluating the oracle. This measures the number of times an adversarial example evades the oracle out of 100 attempts. We also use standard ML metrics such as accuracy, F1, AUC, and false positive rate (FPR) to evaluate the models and defenses.

\vspace{-3mm}
\section{Black-box Results}
\label{sec:blackboxeval}
\vspace{-1.5mm}

\subsection{Transferability Attack}
\vspace{-1.5mm}

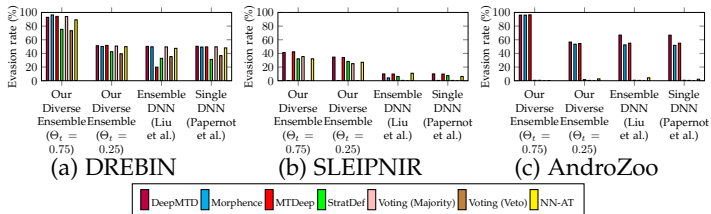
\begin{figure}[!htbp]
\vspace{-2.5mm}
\hspace{-0.4cm}
\centering
\begin{subfigure}[b]{0.16\textwidth}
	\begin{tikzpicture}[scale=0.45]
		\begin{axis}[
			ybar,
			enlarge y limits=false,
			width=7.2cm, height=3.6cm,
			bar width=0.07cm,
			ymin=0, ymax=100,
			ylabel={Evasion rate (\%)},
			symbolic x coords={Our Diverse Ensemble (\(\Theta_{t} = 0.75\)), Our Diverse Ensemble (\(\Theta_{t} = 0.25\)), Ensemble DNN, Single DNN},
			xtick=data,
			xticklabels={Our Diverse Ensemble (\(\Theta_{t} = 0.75\)), Our Diverse Ensemble (\(\Theta_{t} = 0.25\)), Ensemble DNN (Liu et al.), Single DNN (Papernot et al.)},
			xticklabel style={text width=1.5cm,align=center},
			nodes near coords align={horizontal},
			legend pos=north east,
			legend style={nodes={scale=0.4, transform shape}},
			legend to name=evasionratemainleg,
			legend columns=8,
			enlarge x limits={abs=0.6cm},
			ylabel style=
            {
                yshift=-2mm, 
            }
			]
    \addplot [fill=purple]
    coordinates {
        (Our Diverse Ensemble (\(\Theta_{t} = 0.75\)), 92.902)
        (Our Diverse Ensemble (\(\Theta_{t} = 0.25\)), 51.264)
        (Ensemble DNN,50.339)
        (Single DNN, 50.448)
    };
    \addplot [fill=cyan]
    coordinates {
        (Our Diverse Ensemble (\(\Theta_{t} = 0.75\)), 96.344)
        (Our Diverse Ensemble (\(\Theta_{t} = 0.25\)), 50.223)
        (Ensemble DNN,49.663)
        (Single DNN, 49.342)

    };
    \addplot [fill=red]
    coordinates {
        (Our Diverse Ensemble (\(\Theta_{t} = 0.75\)), 94.243)
        (Our Diverse Ensemble (\(\Theta_{t} = 0.25\)), 51.756)
        (Ensemble DNN,19.872)
        (Single DNN, 49.552)

    };
    \addplot [fill=green]
    coordinates {
        (Our Diverse Ensemble (\(\Theta_{t} = 0.75\)), 75.29)
        (Our Diverse Ensemble (\(\Theta_{t} = 0.25\)), 42.46)
        (Ensemble DNN,32.81)
        (Single DNN, 31.2)

    };
    \addplot [fill=pink]
    coordinates {
        (Our Diverse Ensemble (\(\Theta_{t} = 0.75\)), 93.978)
        (Our Diverse Ensemble (\(\Theta_{t} = 0.25\)), 50.814)
        (Ensemble DNN,49.55)
        (Single DNN, 49.661)

    };
    \addplot [fill=brown]
    coordinates {
        (Our Diverse Ensemble (\(\Theta_{t} = 0.75\)), 73.265)
        (Our Diverse Ensemble (\(\Theta_{t} = 0.25\)), 39.494)
        (Ensemble DNN,35.359)
        (Single DNN, 36.717)

    };
    \addplot [fill=yellow]
    coordinates {
        (Our Diverse Ensemble (\(\Theta_{t} = 0.75\)), 89.243)
        (Our Diverse Ensemble (\(\Theta_{t} = 0.25\)), 49.889)
        (Ensemble DNN,47.393)
        (Single DNN, 48.085)
    };

    \legend{DeepMTD, Morphence, MTDeep, StratDef, Voting (Majority), Voting (Veto)}%
    \end{axis}
   
  \end{tikzpicture}
\vspace{-6.5mm}
\caption{DREBIN}
\end{subfigure}
\hspace{0.05cm} 
\begin{subfigure}[b]{0.16\textwidth}
	\begin{tikzpicture}[scale=0.45]
			\begin{axis}[
			ybar,
			enlarge y limits=false,
			width=7.2cm, height=3.6cm,
			bar width=0.07cm,
			ymin=0, ymax=100,
			ylabel={Evasion rate (\%)},
			symbolic x coords={Our Diverse Ensemble (\(\Theta_{t} = 0.75\)), Our Diverse Ensemble (\(\Theta_{t} = 0.25\)), Ensemble DNN, Single DNN},
			xtick=data,
			xticklabels={Our Diverse Ensemble (\(\Theta_{t} = 0.75\)), Our Diverse Ensemble (\(\Theta_{t} = 0.25\)), Ensemble DNN (Liu et al.), Single DNN (Papernot et al.)},
			xticklabel style={text width=1.5cm,align=center},
			nodes near coords align={horizontal},
			legend pos=north east,
			legend style={nodes={scale=0.4, transform shape}},
			legend to name=evasionratemainleg,
			legend columns=8,
			enlarge x limits={abs=0.6cm},
			ylabel style=
            {
                yshift=-2mm, 
            }
			]
    \addplot [fill=purple]
    coordinates {
        (Our Diverse Ensemble (\(\Theta_{t} = 0.75\)), 41.315)
        (Our Diverse Ensemble (\(\Theta_{t} = 0.25\)), 34.589)
        (Ensemble DNN, 9.932)
        (Single DNN, 9.977)
    };
    \addplot [fill=cyan]
    coordinates {
        (Our Diverse Ensemble (\(\Theta_{t} = 0.75\)), 0.3)
        (Our Diverse Ensemble (\(\Theta_{t} = 0.25\)), 0.228)
        (Ensemble DNN,4.093)
        (Single DNN, 0.445)
    };
    \addplot [fill=red]
    coordinates {
        (Our Diverse Ensemble (\(\Theta_{t} = 0.75\)), 42.308)
        (Our Diverse Ensemble (\(\Theta_{t} = 0.25\)), 33.756)
        (Ensemble DNN,9.773)
        (Single DNN, 9.773)

    };
    \addplot [fill=green]
    coordinates {
        (Our Diverse Ensemble (\(\Theta_{t} = 0.75\)), 32.1)
        (Our Diverse Ensemble (\(\Theta_{t} = 0.25\)), 28.26)
        (Ensemble DNN,6.27)
        (Single DNN, 7.416)
    };
    \addplot [fill=pink]
    coordinates {
        (Our Diverse Ensemble (\(\Theta_{t} = 0.75\)), 35.39)
        (Our Diverse Ensemble (\(\Theta_{t} = 0.25\)), 25)
        (Ensemble DNN,0.515)
        (Single DNN, 0.455)

    };
    \addplot [fill=brown]
    coordinates {
        (Our Diverse Ensemble (\(\Theta_{t} = 0.75\)), 0)
        (Our Diverse Ensemble (\(\Theta_{t} = 0.25\)), 0.114)
        (Ensemble DNN,0)
        (Single DNN, 0)
    };
    \addplot [fill=yellow]
    coordinates {
        (Our Diverse Ensemble (\(\Theta_{t} = 0.75\)), 32.101)
        (Our Diverse Ensemble (\(\Theta_{t} = 0.25\)), 26.991)
        (Ensemble DNN,10.944)
        (Single DNN, 6.276)
    };

    \legend{DeepMTD, Morphence, MTDeep, StratDef, Voting (Majority), Voting (Veto), NN-AT}%
    \end{axis}
   
\end{tikzpicture}
\vspace{-6.5mm}
\caption{SLEIPNIR}
\end{subfigure}
\hspace{0.05cm} 
\begin{subfigure}[b]{0.16\textwidth}
	\begin{tikzpicture}[scale=0.45]
			\begin{axis}[
			ybar,
			enlarge y limits=false,
			width=7.2cm, height=3.6cm,
			bar width=0.07cm,
			ymin=0, ymax=100,
			ylabel={Evasion rate (\%)},
			symbolic x coords={Our Diverse Ensemble (\(\Theta_{t} = 0.75\)), Our Diverse Ensemble (\(\Theta_{t} = 0.25\)), Ensemble DNN, Single DNN},
			xtick=data,
			xticklabels={Our Diverse Ensemble (\(\Theta_{t} = 0.75\)), Our Diverse Ensemble (\(\Theta_{t} = 0.25\)), Ensemble DNN (Liu et al.), Single DNN (Papernot et al.)},
			xticklabel style={text width=1.5cm,align=center},
			nodes near coords align={horizontal},
			legend pos=north east,
			legend style={nodes={scale=0.4, transform shape}},
			legend to name=evasionratemainleg,
			legend columns=8,
			enlarge x limits={abs=0.6cm},
			ylabel style=
            {
                yshift=-2mm, 
            }
			]
    \addplot [fill=purple]
    coordinates {
        (Our Diverse Ensemble (\(\Theta_{t} = 0.75\)), 96.1)
        (Our Diverse Ensemble (\(\Theta_{t} = 0.25\)), 56.6)
        (Ensemble DNN,66.9)
        (Single DNN, 66.7)
    };
    \addplot [fill=cyan]
    coordinates {
        (Our Diverse Ensemble (\(\Theta_{t} = 0.75\)), 96.1)
        (Our Diverse Ensemble (\(\Theta_{t} = 0.25\)), 53.5)
        (Ensemble DNN,52.5)
        (Single DNN, 51.8)

    };
    \addplot [fill=red]
    coordinates {
        (Our Diverse Ensemble (\(\Theta_{t} = 0.75\)), 96.5)
        (Our Diverse Ensemble (\(\Theta_{t} = 0.25\)), 54.4)
        (Ensemble DNN,55)
        (Single DNN, 55)

    };
    \addplot [fill=green]
    coordinates {
        (Our Diverse Ensemble (\(\Theta_{t} = 0.75\)), 0.6)
        (Our Diverse Ensemble (\(\Theta_{t} = 0.25\)), 1.6)
        (Ensemble DNN,0.7)
        (Single DNN, 0.7)

    };
    \addplot [fill=pink]
    coordinates {
        (Our Diverse Ensemble (\(\Theta_{t} = 0.75\)), 0.6)
        (Our Diverse Ensemble (\(\Theta_{t} = 0.25\)), 0.5)
        (Ensemble DNN,0.5)
        (Single DNN, 0.5)

    };
    \addplot [fill=brown]
    coordinates {
        (Our Diverse Ensemble (\(\Theta_{t} = 0.75\)), 0)
        (Our Diverse Ensemble (\(\Theta_{t} = 0.25\)), 0)
        (Ensemble DNN,0)
        (Single DNN, 0)

    };
    \addplot [fill=yellow]
    coordinates {
        (Our Diverse Ensemble (\(\Theta_{t} = 0.75\)), 0.2)
        (Our Diverse Ensemble (\(\Theta_{t} = 0.25\)), 2.7)
        (Ensemble DNN,4.3)
        (Single DNN, 2.2)
    };

    \legend{DeepMTD, Morphence, MTDeep, StratDef, Voting (Majority), Voting (Veto), NN-AT}%
    \end{axis}
   
  \end{tikzpicture}
\vspace{-6.5mm}
\caption{AndroZoo}
\end{subfigure}
\ref{evasionratemainleg}
\vspace{-2.5mm}
\caption{Evasion rate of attack strategies.}
\label{figure:evasionratemain}
\vspace{-2.5mm}
\end{figure}

\noindent{\textbf{Evasion rate.}} In general, we observe that attacks against the majority of MTDs are effective regardless of the attack strategy used. Figure~\ref{figure:evasionratemain} shows that MTDs seem easily evaded, with the average evasion rate across all attack strategies sitting at 56\% for DREBIN, 12.9\% for SLEIPNIR, and 50.3\% for AndroZoo. However, the best attack performance is achieved by our attack strategy (Diverse Ensemble with \(\Theta_{t} = 0.75\)), with peak evasion rates of 96.3\% for DREBIN, 42.3\% for SLEIPNIR, and 96.1\% for AndroZoo. This demonstrates that the majority of MTDs possess insufficient movement mechanisms and do not adequately prevent attackers from developing sufficient representations of them. This is especially true in less constrained environments, such as with DREBIN and AndroZoo. For these datasets, the attack surface is greater due to %
the ability to perform more perturbations (i.e., both feature addition and removal), which leads to greater evasion.

For SLEIPNIR, there are greater restrictions on the perturbations that can be applied by attacks. Hence, this represents a more constrained environment for the attacker, where MTDs have it easier to defend themselves. The overall lower average evasion rate implies that MTDs can work well in these conditions. In particular, Morphence and veto voting are evaded the least, with an average evasion rate across all strategies sitting at \(\approx 1\%\) for Morphence and 0\% for veto voting. However, although veto voting offers good protection, it exhibits a high FPR (see Section~\ref{sec:widerpicture} later). Interestingly, StratDef seems least evaded across all the datasets --- for AndroZoo, the evasion rate of the attack sits at below 1\%. Also, although transferability attacks seem to be less effective for SLEIPNIR, we show later that query attacks are more effective and manage to extensively evade all defenses even with this dataset.

\noindent{\textbf{Comparing attack strategies.}} 
In terms of how the different strategies perform, Figure~\ref{figure:evasionratemain} shows that our attack strategy (Diverse Ensemble with $\Theta_{t} = 0.75$) achieves the highest evasion rate. For DREBIN and AndroZoo, the evasion rate sits at 65+\% against most defenses (including single-model defenses), which is considerably greater than the \(\approx 50\%\) evasion rate of the baseline attack strategies (Ensemble DNN and Single DNN). This is because, unlike other attack strategies, our attack strategy maximizes the transferability of adversarial examples locally across an ensemble of diverse substitute models before evaluating them on the oracle, thereby reducing attack failures. Even in a more constrained environment (with SLEIPNIR), our attack strategies surpass the Single DNN and Ensemble DNN attack strategies with a peak evasion rate of 40+\% (versus \(\approx 10\%\)). As our attack strategy (Diverse Ensemble with $\Theta_{t} = 0.75$) offers the highest evasion rate across all datasets, we use this approach for the rest of the black-box evaluation.

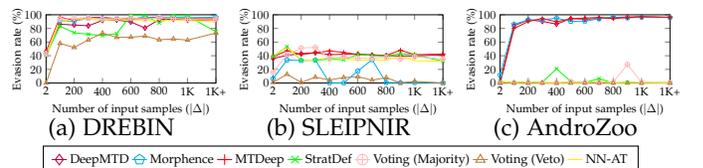
\begin{figure}[!htbp]
\vspace{-2.5mm}
\hspace{-0.25cm}  
\centering
\begin{subfigure}[b]{0.15\textwidth}
    \begin{tikzpicture}[scale=0.45]
        \begin{axis}[
            xlabel={Number of input samples (\(|\Delta|\))},
            ylabel={Evasion rate (\%)}, 
            xmin=2, xmax=1200,
            ymin=0, ymax=100,
            xtick={2,200,400,600,800,1000,1200},
            xticklabels={2,200,400,600,800,1K,1K+},
            ymajorgrids=false,
            legend pos= south west,
            legend style={nodes={scale=0.5, transform shape}},
            cycle list name=color list,
            legend to name=evasionratevariedleg,
            legend columns=8,
            height=3.6cm,
            width=6.6cm,
            ylabel style=
            {
                yshift=-2mm, 
            }
        ]

		\addplot[color=purple,mark=diamond]
            coordinates {
				(2,45.876)(100,86.147)(200,85.053)(300,83.918)(400,91.741)(500,92.428)(600,89.429)(700,80.645)(800,93.157)(900,93.612)(1000,92.157)(1200,92.902)
            };
        \addlegendentry{DeepMTD}
		
		\addplot[color=cyan,mark=pentagon]
            coordinates {
				(2,42.365)(100,91.314)(200,90.618)(300,94.702)(400,94.36)(500,94.336)(600,94.218)(700,96.272)(800,96.718)(900,96.312)(1000,95.699)(1200,96.344)
            };
        \addlegendentry{Morphence}

		\addplot[color=red,mark=+]
            coordinates {
				(2,45.946)(100,96.403)(200,91.957)(300,95.862)(400,96.145)(500,94.407)(600,93.86)(700,94.57)(800,96.28)(900,94.181)(1000,94.36)(1200,94.243)
            };
        \addlegendentry{MTDeep}

        \addplot[color=green,mark=x]
            coordinates {
				(2,40.722)(100,83.292)(200,73.747)(300,71.501)(400,69.615)(500,72.33)(600,99.497)(700,98.519)(800,88.106)(900,99.007)(1000,98.305)(1200,75.291)
            };
        \addlegendentry{StratDef}

     	\addplot[color=pink,mark=oplus]
            coordinates {
				(2,43.005)(100,93.333)(200,95.139)(300,95.381)(400,93.541)(500,94.676)(600,95.74)(700,93.261)(800,95.982)(900,94.725)(1000,93.534)(1200,93.978)
            };
        \addlegendentry{Voting (Majority)}

		\addplot[color=brown,mark=triangle]
            coordinates {
				(2,0)(100,58.009)(200,52.174)(300,63.257)(400,73.14)(500,66.878)(600,67.152)(700,68.763)(800,63.374)(900,64.73)(1000,62.887)(1200,73.265)
            };
        \addlegendentry{Voting (Veto)}

		\addplot[color=yellow]
            coordinates {
			(2, 40.5)(100, 90.625)(200, 92.473)(300, 93.548)(400, 91.476)(500, 93.067)(600, 92.276)(700, 92.144)(800, 90.515)(900, 92.585)(1000, 91.841)(1200,89.243)
            };
        \addlegendentry{NN-AT}

        \end{axis}
    \end{tikzpicture}
    \vspace{-6.5mm}
\caption{DREBIN}
\end{subfigure}
\hspace{0.1cm} 
\begin{subfigure}[b]{0.15\textwidth}
    \begin{tikzpicture}[scale=0.45]
        \begin{axis}[
            xlabel={Number of input samples (\(|\Delta|\))},
            ylabel={Evasion rate (\%)}, 
            xmin=2, xmax=1200,
            ymin=0, ymax=100,
            xtick={2,200,400,600,800,1000,1200},
            xticklabels={2,200,400,600,800,1K,1K+},
            ymajorgrids=false,
            legend pos= south west,
            legend style={nodes={scale=0.5, transform shape}},
            cycle list name=color list,
            legend to name=evasionratevariedleg,
            legend columns=8,
            height=3.6cm,
            width=6.6cm,
            ylabel style=
            {
                yshift=-2mm, 
            }
        ]
		\addplot[color=purple,mark=diamond]
            coordinates {
				(2,35.798)(100,40.63)(200,43.008)(300,44.511)(400,41.548)(500,42.147)(600,42.289)(700,40.491)(800,40)(900,39.783)(1000,39.722)(1200,41.315)
            };
        \addlegendentry{DeepMTD}
		
		\addplot[color=cyan,mark=pentagon]
            coordinates {
				(2,6.439)(100,33.76)(200,33.017)(300,33.655)(400,0.338)(500,0.168)(600,17.722)(700,34.146)(800,0.647)(900,0.549)(1000,0.639)(1200,0.3)
            };
        \addlegendentry{Morphence}

		\addplot[color=red,mark=+]
            coordinates {
				(2,38.743)(100,47.377)(200,41.903)(300,44.271)(400,47.08)(500,44.84)(600,41.824)(700,41.475)(800,40.524)(900,48.066)(1000,41.587)(1200,42.308)
            };
        \addlegendentry{MTDeep}

         \addplot[color=green,mark=x]
            coordinates {
				(2,38.861)(100,53.295)(200,33.55)(300,33.442)(400,35.034)(500,34.206)(600,41.693)(700,40.524)(800,39.755)(900,43.023)(1000,41.229)(1200,32.053)
            };
        \addlegendentry{StratDef}
        
        		\addplot[color=pink,mark=oplus]
            coordinates {
				(2,15.594)(100,41.766)(200,51.25)(300,52.258)(400,35.774)(500,38.204)(600,36.227)(700,36.06)(800,37.937)(900,34.992)(1000,38.462)(1200,35.39)
            };
        \addlegendentry{Voting (Majority)}

		\addplot[color=brown,mark=triangle]
            coordinates {
				(2,0.99)(100,12.89)(200,0.936)(300,8.607)(400,4.466)(500,8.103)(600,9.423)(700,4.244)(800,7.971)(900,0.183)(1000,2.427)(1200,0)
            };
        \addlegendentry{Voting (Veto)}

		\addplot[color=yellow]
            coordinates {
			(2, 38.119)(100, 53.295)(200, 33.387)(300, 33.819)(400, 37.818)(500, 35.932)(600, 33.072)(700, 33.945)(800, 33.639)(900, 39.427)(1000, 32.73)(1200,32.101)

            };
        \addlegendentry{NN-AT}

        \end{axis}
    \end{tikzpicture}
    \vspace{-6.5mm}
\caption{SLEIPNIR}
\end{subfigure}
\hspace{0.1cm} 
\begin{subfigure}[b]{0.15\textwidth}
    \begin{tikzpicture}[scale=0.45]
        \begin{axis}[
            xlabel={Number of input samples (\(|\Delta|\))},
            ylabel={Evasion rate (\%)}, 
            xmin=2, xmax=1200,
            ymin=0, ymax=100,
            xtick={2,200,400,600,800,1000,1200},
            xticklabels={2,200,400,600,800,1K,1K+},
            ymajorgrids=false,
            legend pos= south west,
            legend style={nodes={scale=0.5, transform shape}},
            cycle list name=color list,
            legend to name=androzooevasionratevariedleg,
            legend columns=8,
            height=3.6cm,
            width=6.6cm,
            ylabel style=
            {
                yshift=-2mm, 
            }
        ]

		\addplot[color=purple,mark=diamond]
            coordinates {
				(2,1.5)(100,83.8)(200,92.6)(300,90.4)(400,86.3)(500,94.4)(600,93.5)(700,94.6)(800,96.2)(900,96.7)(1000,97.1)(1200,96.1)
            };
        \addlegendentry{DeepMTD}
		
		\addplot[color=cyan,mark=pentagon]
            coordinates {
				(2,11.8)(100,86.1)(200,92.4)(300,92.7)(400,95.1)(500,90.3)(600,90.7)(700,94.2)(800,96.2)(900,95.4)(1000,96.7)(1200,96.1)
            };
        \addlegendentry{Morphence}

		\addplot[color=red,mark=+]
            coordinates {
				(2,0)(100,79.9)(200,90.2)(300,94.5)(400,89.7)(500,94.2)(600,95.1)(700,96.2)(800,94.4)(900,96)(1000,97.1)(1200,96.5)
            };
        \addlegendentry{MTDeep}

        \addplot[color=green,mark=x]
            coordinates {
				(2,2.3)(100,0.7)(200,1.8)(300,1.1)(400,20.9)(500,2.3)(600,0.7)(700,6.5)(800,0)(900,1.1)(1000,0.4)(1200,0.6)
            };
        \addlegendentry{StratDef}

     	\addplot[color=pink,mark=oplus]
            coordinates {
				(2,0)(100,0.2)(200,0.4)(300,0.5)(400,0.4)(500,2)(600,0.7)(700,0.7)(800,1.3)(900,27.2)(1000,0.9)(1200,0.6)
            };
        \addlegendentry{Voting (Majority)}

		\addplot[color=brown,mark=triangle]
            coordinates {
				(2,0)(100,0.2)(200,0)(300,0.2)(400,0)(500,0)(600,0.2)(700,0.2)(800,0.2)(900,0)(1000,0)(1200,0)
            };
        \addlegendentry{Voting (Veto)}

		\addplot[color=yellow]
            coordinates {
				(2,2.3)(100,0.7)(200,1.6)(300,0.9)(400,0.6)(500,0.9)(600,0.6)(700,2.6)(800,0.4)(900,0.9)(1000,0.9)(1200,0.2)
            };
        \addlegendentry{NN-AT}

        \end{axis}
    \end{tikzpicture}
    \vspace{-6.5mm}
\caption{AndroZoo}
\end{subfigure}
\ref{evasionratevariedleg}
\vspace{-2.5mm}
\caption{Evasion rate vs. number of input samples.}

\label{figure:evasionratevaried}
\vspace{-2.5mm}
\end{figure}

\noindent{\textbf{Varying the number of input samples (\(|\Delta|\)).}} We now vary the input samples to construct substitute models. This allows us to explore whether an effective transferability attack can be performed against MTDs with fewer samples to construct substitute models. Previously, the maximum number of input samples was used to develop \(\Delta\) (i.e., \(|B_{train}|\)). We therefore cap \(|\Delta|\) and query each defense with different numbers of input samples from each class, up to the maximum available (1K+, i.e., \(|B_{train}|\)). As \(|\Delta|\) increases, it should produce a better representation of the oracle (up to a point), while fewer input samples should be less detectable. Recall that we use our attack strategy (Diverse Ensemble with $\Theta_{t} = 0.75$) for this experiment. Figure~\ref{figure:evasionratevaried} shows that for DREBIN and AndroZoo, as \(|\Delta|\) increases past two (where the evasion rate sits at \(\approx40\%\)), the evasion rate increases and reaches up to 90+\% after which it stabilizes. Generally, veto voting and StratDef are more robust defenses with less evasion for DREBIN these datasets. Fluctuations in the performance of some defenses are attributable to several reasons, such as model regeneration (Morphence) and dynamic model selection (StratDef). Moreover, \(\Delta\) can become noisier due to inaccurate data, resulting in substitute models with a poorer approximation of the oracle's behavior at some intervals, explaining the dips in evasion rates for some defenses (DeepMTD and veto voting). Interestingly, most defenses are no better than the single-model defense, NN-AT.%

Meanwhile, in the more constrained environment with SLEIPNIR, attack performance varies more, with the evasion rate peaking at 53\%. However, there is significantly less evasion compared with DREBIN and AndroZoo due to a smaller attack surface being available to exploit. %
Morphence's model regeneration procedure causes it to exhibit greater fluctuations in its performance under this more constrained dataset. %
Despite this, evasion is achievable even if its query budget is exceeded. %
For AndroZoo, we observe that nearly all MTDs are evaded in the same manner as DREBIN. As $|\Delta|$ increases past 100, the evasion rate against these defenses sits at 80+\%. Interestingly, though, StratDef and the non-MTDs exhibit significant robustness; in some cases, the attack fails to achieve significant evasion.

\begin{figure}[!htbp]
\vspace{-2.5mm}
\hspace{-0.25cm} 
\centering
\begin{subfigure}[b]{0.15\textwidth}
    \begin{tikzpicture}[scale=0.45]
        \begin{axis}[
            xlabel={Number of input samples (\(|\Delta|\))},
            ylabel={Avg. RER (\%)}, 
            xmin=2, xmax=1200,
            ymin=0, ymax=100,
            xtick={2,200,400,600,800,1000,1200},
            xticklabels={2,200,400,600,800,1K,1K+},
            ymajorgrids=false,
            legend pos= south west,
            legend style={nodes={scale=0.5, transform shape}},
            cycle list name=color list,
            legend to name=reratevariedleg,
            legend columns=8,
            height=3.6cm,
            width=6.6cm,
            ylabel style=
            {
                yshift=-2mm, 
            }
        ]

		\addplot[color=purple,mark=diamond]
            coordinates {
				(2,100)(100,100)(200,100)(300,100)(400,100)(500,100)(600,100)(700,100)(800,100)(900,100)(1000,100)(1200,100)

            };
        \addlegendentry{DeepMTD}
		
		\addplot[color=cyan,mark=pentagon]
            coordinates {
				(2,100)(100,100)(200,100)(300,100)(400,100)(500,100)(600,100)(700,100)(800,100)(900,100)(1000,100)(1200,100)

            };
        \addlegendentry{Morphence}

		\addplot[color=red,mark=+]
            coordinates {
				(2,100)(100,100)(200,100)(300,100)(400,100)(500,100)(600,100)(700,100)(800,100)(900,100)(1000,100)(1200,100)

            };
        \addlegendentry{MTDeep}

        \addplot[color=green,mark=x]
            coordinates {
				(2,100)(100,100)(200,100)(300,100)(400,100)(500,100)(600,90.136)(700,87.639)(800,88.223)(900,87.729)(1000,87.416)(1200,100)

            };
        \addlegendentry{StratDef}

     	\addplot[color=pink,mark=oplus]
            coordinates {
				(2,100)(100,100)(200,100)(300,100)(400,100)(500,100)(600,100)(700,100)(800,100)(900,100)(1000,100)(1200,100)

            };
        \addlegendentry{Voting (Majority)}

		\addplot[color=brown,mark=triangle]
            coordinates {
				(2,0)(100,100)(200,100)(300,100)(400,100)(500,100)(600,100)(700,100)(800,100)(900,100)(1000,100)(1200,100)

            };
        \addlegendentry{Voting (Veto)}

		\addplot[color=yellow]
            coordinates {
				(2,100)(100,100)(200,100)(300,100)(400,100)(500,100)(600,100)(700,100)(800,100)(900,100)(1000,100)(1200,100)

            };
        \addlegendentry{NN-AT}

        \end{axis}
    \end{tikzpicture}
    \vspace{-6.5mm}
\caption{DREBIN}
\end{subfigure}
\hspace{0.1cm}
\begin{subfigure}[b]{0.15\textwidth}
    \begin{tikzpicture}[scale=0.45]
        \begin{axis}[
            xlabel={Number of input samples (\(|\Delta|\))},
            ylabel={Avg. RER (\%)}, 
            xmin=2, xmax=1200,
            ymin=0, ymax=100,
            xtick={2,200,400,600,800,1000,1200},
            xticklabels={2,200,400,600,800,1K,1K+},
            ymajorgrids=false,
            legend pos= south west,
            legend style={nodes={scale=0.5, transform shape}},
            cycle list name=color list,
            legend to name=reratevariedleg,
            legend columns=8,
            height=3.6cm,
            width=6.6cm,
            ylabel style=
            {
                yshift=-2mm, 
            }
        ]
		\addplot[color=purple,mark=diamond]
            coordinates {
				(2,100)(100,100)(200,100)(300,100)(400,100)(500,100)(600,100)(700,100)(800,100)(900,100)(1000,100)(1200,100)

            };
        \addlegendentry{DeepMTD}
		
		\addplot[color=cyan,mark=pentagon]
            coordinates {
				(2,100)(100,100)(200,100)(300,100)(400,100)(500,100)(600,100)(700,100)(800,100)(900,100)(1000,100)(1200,100)

            };
        \addlegendentry{Morphence}

		\addplot[color=red,mark=+]
            coordinates {
				(2,100)(100,100)(200,100)(300,100)(400,100)(500,100)(600,100)(700,100)(800,100)(900,100)(1000,100)(1200,100)

            };
        \addlegendentry{MTDeep}

         \addplot[color=green,mark=x]
            coordinates {
				(2,61.325)(100,61.353)(200,57.311)(300,49.918)(400,49.816)(500,60.923)(600,51.38)(700,46.589)(800,51.781)(900,52.71)(1000,43.982)(1200,50.853)

            };
        \addlegendentry{StratDef}
        
        		\addplot[color=pink,mark=oplus]
            coordinates {
				(2,100)(100,100)(200,100)(300,100)(400,100)(500,100)(600,100)(700,100)(800,100)(900,100)(1000,100)(1200,100)

            };
        \addlegendentry{Voting (Majority)}

		\addplot[color=brown,mark=triangle]
            coordinates {
				(2,100)(100,100)(200,100)(300,100)(400,100)(500,100)(600,100)(700,100)(800,100)(900,100)(1000,100)(1200,100)

            };
        \addlegendentry{Voting (Veto)}

		\addplot[color=yellow]
            coordinates {
				(2,100)(100,100)(200,100)(300,100)(400,100)(500,100)(600,100)(700,100)(800,100)(900,100)(1000,100)(1200,100)

            };
        \addlegendentry{NN-AT}

        \end{axis}
    \end{tikzpicture}
\vspace{-6.5mm}
\caption{SLEIPNIR}
\end{subfigure}
\hspace{0.1cm}
\begin{subfigure}[b]{0.15\textwidth}
    \begin{tikzpicture}[scale=0.45]
        \begin{axis}[
            xlabel={Number of input samples (\(|\Delta|\))},
            ylabel={Avg. RER (\%)}, 
            xmin=2, xmax=1200,
            ymin=0, ymax=100,
            xtick={2,200,400,600,800,1000,1200},
            xticklabels={2,200,400,600,800,1K,1K+},
            ymajorgrids=false,
            legend pos= south west,
            legend style={nodes={scale=0.5, transform shape}},
            cycle list name=color list,
            legend to name=reratevariedleg,
            legend columns=8,
            height=3.6cm,
            width=6.6cm,
            ylabel style=
            {
                yshift=-2mm, 
            }
        ]
		\addplot[color=purple,mark=diamond]
            coordinates {
				(2,100)(100,100)(200,100)(300,100)(400,100)(500,100)(600,100)(700,100)(800,100)(900,100)(1000,100)(1200,100)

            };
        \addlegendentry{DeepMTD}
		
		\addplot[color=cyan,mark=pentagon]
            coordinates {
				(2,100)(100,100)(200,100)(300,100)(400,100)(500,100)(600,100)(700,100)(800,100)(900,100)(1000,100)(1200,100)

            };
        \addlegendentry{Morphence}

		\addplot[color=red,mark=+]
            coordinates {
				(2,100)(100,100)(200,100)(300,100)(400,100)(500,100)(600,100)(700,100)(800,100)(900,100)(1000,100)(1200,100)

            };
        \addlegendentry{MTDeep}

         \addplot[color=green,mark=x]
            coordinates {
				(2,81.2)(100,62.6667)(200,83.25)(300,67)(400,51.396)(500,59.3)(600,100)(700,49.7742)(800,0)(900,81.6)(1000,100)(1200,49.3333)

            };
        \addlegendentry{StratDef}
        
        		\addplot[color=pink,mark=oplus]
            coordinates {
				(2,100)(100,100)(200,100)(300,100)(400,100)(500,100)(600,100)(700,100)(800,100)(900,100)(1000,100)(1200,100)

            };
        \addlegendentry{Voting (Majority)}

		\addplot[color=brown,mark=triangle]
            coordinates {
				(2,100)(100,100)(200,100)(300,100)(400,100)(500,100)(600,100)(700,100)(800,100)(900,100)(1000,100)(1200,100)

            };
        \addlegendentry{Voting (Veto)}

		\addplot[color=yellow]
            coordinates {
				(2,100)(100,100)(200,100)(300,100)(400,100)(500,100)(600,100)(700,100)(800,100)(900,100)(1000,100)(1200,100)

            };
        \addlegendentry{NN-AT}

        \end{axis}
    \end{tikzpicture}
\vspace{-6.5mm}
\caption{AndroZoo}
\end{subfigure}

\ref{reratevariedleg}
\vspace{-2.5mm}
\caption{Average RER vs. number of input samples. Lines appear merged as most defenses exhibit 100\% RER.}
\label{figure:reratevaried}
\vspace{-2.5mm}
\end{figure}
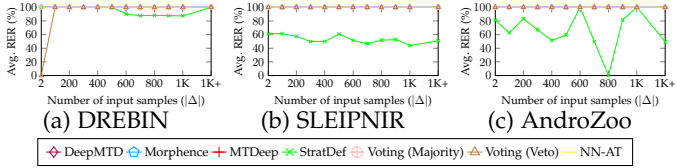

\noindent{\textbf{Repeat Evasion Rate.}} Figure~\ref{figure:reratevaried} shows the average repeat evasion rate (RER) versus the number of input samples. This measures how many times (out of 100) an adversarial example evades the oracle, averaged across the attack. For StratDef, the average RER is 95\% for DREBIN (minimum of 87\%), 53\% for SLEIPNIR (minimum of 44\%), and 65.4\% for AndroZoo (minimum of 0\%). Meanwhile, for DeepMTD, Morphence, MTDeep, and voting, we achieve 100\% RER, which means that adversarial examples have a higher chance of repeatedly evading the same oracle. This is due to the predictive nature of each defense, the characteristics of their \emph{movement} mechanisms, or the lack of diversity among their constituent models. As single-model defenses such as NN-AT %
are completely static, they experience 100\% RER.

\vspace{-2mm}
\subsection{Query Attack}
\label{sec:blackboxqueryattack}
\vspace{-1.5mm}
Unlike the transferability attack, the black-box query attack does not use substitute models. Instead, a malware sample is perturbed using features from benign samples until it evades the oracle or resources are exhausted (e.g., number of queries, available features to perturb).  

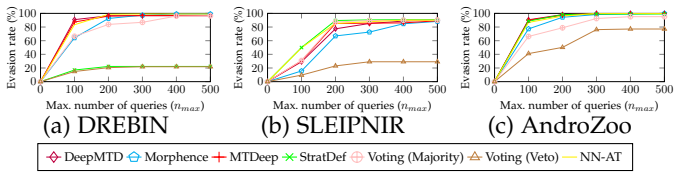
\begin{figure}[!htbp]
\vspace{-2.5mm}
\hspace{-0.25cm} 
\centering
\begin{subfigure}[b]{0.15\textwidth}
    \begin{tikzpicture}[scale=0.45]
        \begin{axis}[
            xlabel={Max. number of queries (\(n_{max}\))},
            ylabel={Evasion rate (\%)}, 
            xmin=0, xmax=500,
            ymin=0, ymax=100,
            xtick={0,100,200,300,400,500},
            ymajorgrids=false,
            legend pos= south west,
            legend style={nodes={scale=0.5, transform shape}},
            cycle list name=color list,
            legend to name=evasionrateblackboxqbasedleg,
            legend columns=8,
            height=3.6cm,
            width=6.6cm,
            ylabel style=
            {
                yshift=-2mm, 
            }
        ]

		\addplot[color=purple,mark=diamond]
            coordinates {
				(0,0)(100, 90.83)(200, 96.507)(300, 96.507)(400, 96.507)(500, 96.507)

            };
        \addlegendentry{DeepMTD}
		
		\addplot[color=cyan,mark=pentagon]
            coordinates {
				(0,0)(100, 64.192)(200, 92.576)(300, 97.38)(400, 99.127)(500, 99.127)

		};
        \addlegendentry{Morphence}

		\addplot[color=red,mark=+]
            coordinates {
				(0,0)(100, 87.336)(200, 96.07)(300, 96.943)(400, 96.943)(500, 96.943)

            };
        \addlegendentry{MTDeep}

        \addplot[color=green,mark=x]
            coordinates {
				(0,0)(100, 17.031)(200, 22.271)(300, 22.271)(400, 22.271)(500, 22.271)

            };
        \addlegendentry{StratDef}

     	\addplot[color=pink,mark=oplus]
            coordinates {
				(0,0)(100, 66.376)(200, 83.843)(300, 86.9)(400, 96.07)(500, 96.507)

            };
        \addlegendentry{Voting (Majority)}

		\addplot[color=brown,mark=triangle]
            coordinates {
				(0,0)(100, 14.847)(200, 20.524)(300, 21.834)(400, 21.834)(500, 21.834)

            };
        \addlegendentry{Voting (Veto)}

		\addplot[color=yellow]
            coordinates {
				(0,0)(100, 83.406)(200, 98.69)(300, 99.127)(400, 99.127)(500, 99.127)

            };
        \addlegendentry{NN-AT}

        \end{axis}
    \end{tikzpicture}
\vspace{-6.5mm}
\caption{DREBIN}
\end{subfigure}
\hspace{0.1cm} 
\begin{subfigure}[b]{0.15\textwidth}
    \begin{tikzpicture}[scale=0.45]
        \begin{axis}[
            xlabel={Max. number of queries (\(n_{max}\))},
            ylabel={Evasion rate (\%)}, 
            xmin=0, xmax=500,
            ymin=0, ymax=100,
            xtick={0,100,200,300,400,500},
            ymajorgrids=false,
            legend pos= south west,
            legend style={nodes={scale=0.5, transform shape}},
            cycle list name=color list,
            legend to name=evasionrateblackboxqbasedleg,
            legend columns=8,
            height=3.6cm,
            width=6.6cm,
            ylabel style=
            {
                yshift=-2mm, 
            }
        ]
		\addplot[color=purple,mark=diamond]
            coordinates {
				(0,0)(100, 29.13)(200, 77.391)(300, 85.217)(400, 88.696)(500, 90.435)

            };
        \addlegendentry{DeepMTD}
		
		\addplot[color=cyan,mark=pentagon]
            coordinates {
				(0,0)(100, 15.652)(200, 66.957)(300, 72.609)(400, 84.783)(500, 88.261)

            };
        \addlegendentry{Morphence}

		\addplot[color=red,mark=+]
            coordinates {
				(0,0)(100, 30.435)(200, 85.217)(300, 85.652)(400, 86.087)(500, 89.13)

            };
        \addlegendentry{MTDeep}

         \addplot[color=green,mark=x]
            coordinates {
				(0,0)(100, 50)(200, 89.565)(300, 90.435)(400, 90.87)(500, 90.87)

            };
        \addlegendentry{StratDef}
        
        		\addplot[color=pink,mark=oplus]
            coordinates {
				(0,0)(100, 30.87)(200, 87.391)(300, 89.565)(400, 90)(500, 90)

            };
        \addlegendentry{Voting (Majority)}

		\addplot[color=brown,mark=triangle]
            coordinates {
				(0,0)(100, 9.565)(200, 23.043)(300, 29.13)(400, 29.13)(500, 29.13)

            };
        \addlegendentry{Voting (Veto)}

		\addplot[color=yellow]
            coordinates {
				(0,0)(100, 49.565)(200, 85.217)(300, 87.826)(400, 90)(500, 90)

            };
        \addlegendentry{NN-AT}

        \end{axis}
    \end{tikzpicture}
\vspace{-6.5mm}
\caption{SLEIPNIR}
\end{subfigure}
\hspace{0.1cm} 
\begin{subfigure}[b]{0.15\textwidth}
    \begin{tikzpicture}[scale=0.45]
        \begin{axis}[
            xlabel={Max. number of queries (\(n_{max}\))},
            ylabel={Evasion rate (\%)}, 
            xmin=0, xmax=500,
            ymin=0, ymax=100,
            xtick={0,100,200,300,400,500},
            ymajorgrids=false,
            legend pos= south west,
            legend style={nodes={scale=0.5, transform shape}},
            cycle list name=color list,
            legend to name=evasionrategrayboxqbasedlegandrozoo,
            legend columns=8,
            height=3.6cm,
            width=6.6cm,
            ylabel style=
            {
                yshift=-2mm, 
            }
        ]

		\addplot[color=purple,mark=diamond]
            coordinates {
				(0,0)(100,90.6)(200,98.2)(300,100)(400,100)(500,100)

            };
        \addlegendentry{DeepMTD}
        
		\addplot[color=cyan,mark=pentagon]
            coordinates {
				(0,0)(100,77.3)(200,94.3)(300,98.3)(400,99.1)(500,100)
		};
        \addlegendentry{Morphence}

		\addplot[color=red,mark=+]
            coordinates {
				(0,0)(100,90)(200,97.3)(300,99.5)(400,100)(500,100)

            };
        \addlegendentry{MTDeep}

        \addplot[color=green,mark=x]
            coordinates {
				(0,0)(100,88.6)(200,98.2)(300,99.1)(400,99.1)(500,99.1)

            };
        \addlegendentry{StratDef}

     	\addplot[color=pink,mark=oplus]
            coordinates {
				(0,0)(100,66.2)(200,78.9)(300,92.5)(400,95.2)(500,95.2) 

            };
        \addlegendentry{Voting (Majority)}

		\addplot[color=brown,mark=triangle]
            coordinates {
				(0,0)(100,41.1)(200,50.2)(300,76.2)(400,77.1)(500,77.1)

            };
        \addlegendentry{Voting (Veto)}

		\addplot[color=yellow]
            coordinates {
				(0,0)(100,86.7)(200,96)(300,100)(400,100)(500,100)

            };
        \addlegendentry{NN-AT}

        \end{axis}
    \end{tikzpicture}
\vspace{-6.5mm}
\caption{AndroZoo}
\end{subfigure}
\ref{evasionrateblackboxqbasedleg}
\vspace{-2.5mm}
\caption{Evasion rate vs. maximum number of queries.}
\label{figure:evasionrateblackboxqbased}
\vspace{-2.5mm}
\end{figure}

Figure~\ref{figure:evasionrateblackboxqbased} shows the results of the black-box query attack. This attack clearly outperforms the transferability attack, as models that were hardly evaded by the transferability attack (especially for SLEIPNIR and AndroZoo) are evaded by the query attack with fewer queries. The average evasion rate of the query attack across all defenses sits at 72\% for DREBIN, 69\% for SLEIPNIR, and 90.3\% for AndroZoo (with peak evasion rates of 99\%, 91\%, and 100\% respectively). For DREBIN, veto voting and StratDef exhibit the greatest robustness, with the attack only achieving \(\approx 20\%\) evasion rate. For SLEIPNIR, although the environment is still more constrained in terms of the attack surface (i.e., fewer possible perturbations), veto voting and Morphence can be evaded still, unlike in the transferability attack. As before, the evasion rate is relatively lower for SLEIPNIR because there are fewer perturbations that can be applied to the malware samples, limiting the evasion opportunities. However, as the perturbations used by the attack are tailored to this domain, we observe that the attack yields better attack performance. This is evident in the results of the attack against StratDef for AndroZoo, where the black-box transferability attack yielded insignificant evasion. Veto voting remains robust across the datasets because a single model influences the prediction and the perturbations selected by the attack are insufficient to evade it. However, it exhibits a higher FPR (see Section~\ref{sec:widerpicture} later). Defenses such as DeepMTD, MTDeep, and majority voting --- as well as the single-model defense, NN-AT --- offer minimal protection against this attack.

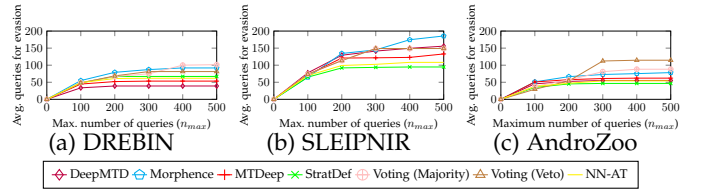
\begin{figure}[!htbp]
\vspace{-2.5mm}
\hspace{-0.25cm} 
\centering
\begin{subfigure}[b]{0.15\textwidth}
    \begin{tikzpicture}[scale=0.45]
        \begin{axis}[
            xlabel={Max. number of queries (\(n_{max}\))},
            ylabel={Avg. queries for evasion}, 
            xmin=0, xmax=500,
            ymin=0, ymax=200,
            xtick={0,100,200,300,400,500},
            ymajorgrids=false,
            legend pos= south west,
            legend style={nodes={scale=0.5, transform shape}},
            cycle list name=color list,
            legend to name=numqueriesblackboxleg,
            legend columns=8,
            height=3.6cm,
            width=6.6cm,
            ylabel style=
            {
                yshift=-2mm, 
            }
        ]

		\addplot[color=purple,mark=diamond]
            coordinates {
            (0,0)
				(100, 33.683)
				(200, 39.086)
				(300, 39.086)
				(400, 39.086)
				(500, 39.086)
				
            };
        \addlegendentry{DeepMTD}
		
		\addplot[color=cyan,mark=pentagon]
            coordinates {
            (0,0)
				(100, 54.993)
				(200, 79.085)
				(300, 87.179)
				(400, 91.969)
				(500, 91.969)

		};
        \addlegendentry{Morphence}

		\addplot[color=red,mark=+]
            coordinates {
            (0,0)
				(100, 44.78)
				(200, 52.009)
				(300, 53.387)
				(400, 53.387)
				(500, 53.387)
				
            };
        \addlegendentry{MTDeep}

        \addplot[color=green,mark=x]
            coordinates {
            (0,0)
				(100, 48.436)
(200, 67.078)
(300, 67.078)
(400, 67.078)
(500, 67.078)

            };
        \addlegendentry{StratDef}

     	\addplot[color=pink,mark=oplus]
            coordinates {
            (0,0)
				(100, 49.23)
(200, 67.047)
(300, 74.653)
(400, 100.105)
(500, 101.579)

            };
        \addlegendentry{Voting (Majority)}

		\addplot[color=brown,mark=triangle]
            coordinates {
            (0,0)
				(100, 47.912)
(200, 68.702)
(300, 81.34)
(400, 81.34)
(500, 81.34)

            };
        \addlegendentry{Voting (Veto)}

		\addplot[color=yellow]
            coordinates {
				
            (0,0)
				(100, 48.822)
				(200, 61.009)
				(300, 61.74)
				(400, 61.74)
				(500, 61.74)
				
            };
        \addlegendentry{NN-AT}

        \end{axis}
    \end{tikzpicture}
\vspace{-6.5mm}
\caption{DREBIN}
\end{subfigure}
\hspace{0.1cm}
\begin{subfigure}[b]{0.15\textwidth}
    \begin{tikzpicture}[scale=0.45]
        \begin{axis}[
            xlabel={Max. number of queries (\(n_{max}\))},
            ylabel={Avg. queries for evasion}, 
            xmin=0, xmax=500,
            ymin=0, ymax=200,
            xtick={0,100,200,300,400,500},
            ymajorgrids=false,
            legend pos= south west,
            legend style={nodes={scale=0.5, transform shape}},
            cycle list name=color list,
            legend to name=numqueriesblackboxleg,
            legend columns=8,
            height=3.6cm,
            width=6.6cm,
            ylabel style=
            {
                yshift=-2mm, 
            }
        ]
		\addplot[color=purple,mark=diamond]
            coordinates {
            (0,0)
				(100, 77.522)
(200, 129.157)
(300, 141.413)
(400, 149.456)
(500, 155.817)

            };
        \addlegendentry{DeepMTD}
		
		\addplot[color=cyan,mark=pentagon]
            coordinates {
            (0,0)
				(100, 64.694)
(200, 134.364)
(300, 144.898)
(400, 174.605)
(500, 185.611)

            };
        \addlegendentry{Morphence}

		\addplot[color=red,mark=+]
            coordinates {
            (0,0)
				(100, 71.8)
(200, 120.954)
(300, 121.675)
(400, 122.864)
(500, 132.956)

            };
        \addlegendentry{MTDeep}

         \addplot[color=green,mark=x]
            coordinates {
            (0,0)
				(100, 65.052)
(200, 92.083)
(300, 93.625)
(400, 94.794)
(500, 94.794)

            };
        \addlegendentry{StratDef}
        
        		\addplot[color=pink,mark=oplus]
            coordinates {
            (0,0)
				(100, 73.773)
(200, 113.453)
(300, 148.881)
(400, 148.881)
(500, 148.881)

            };
        \addlegendentry{Voting (Majority)}

		\addplot[color=brown,mark=triangle]
            coordinates {
            (0,0)
				(100, 73.773)
(200, 113.453)
(300, 148.881)
(400, 148.881)
(500, 148.881)

            };
        \addlegendentry{Voting (Veto)}

		\addplot[color=yellow]
            coordinates {
            (0,0)
				(100, 69.535)
				(200, 99.199)
				(300, 102.53)
				(400, 108.3)
				(500, 108.3)

            };
        \addlegendentry{NN-AT}

        \end{axis}
    \end{tikzpicture}
\vspace{-6.5mm}
\caption{SLEIPNIR}
\end{subfigure}
\hspace{0.1cm}
\begin{subfigure}[b]{0.15\textwidth}
    \begin{tikzpicture}[scale=0.45]
        \begin{axis}[
            xlabel={Maximum number of queries (\(n_{max}\))},
            ylabel={Avg. queries for evasion}, 
            xmin=0, xmax=500,
            ymin=0, ymax=200,
            xtick={0,100,200,300,400,500},
            ymajorgrids=false,
            legend pos= south west,
            legend style={nodes={scale=0.5, transform shape}},
            cycle list name=color list,
            legend to name=avgqueriesgrayboxqbasedleg,
            legend columns=8,
            height=3.6cm,
            width=6.6cm,
            ylabel style=
            {
                yshift=-2mm, 
            }
        ]
        
		\addplot[color=purple,mark=diamond]
            coordinates {
				(0,0)(100,45.4)(200,52.5)(300,55.3)(400,55.3)(500,55.3)

            };
        \addlegendentry{DeepMTD}
        
		\addplot[color=cyan,mark=pentagon]
            coordinates {
				(0,0)(100,51.2)(200,66.4)(300,72.8)(400,75.3)(500,78.3)
		};
        \addlegendentry{Morphence}

		\addplot[color=red,mark=+]
            coordinates {
				(0,0)(100,50.9)(200,57.6)(300,61)(400,62.1)(500,62.1)

            };
        \addlegendentry{MTDeep}

        \addplot[color=green,mark=x]
            coordinates {
				(0,0)(100,34.9)(200,45.2)(300,46.6)(400,46.6)(500,46.6)

            };
        \addlegendentry{StratDef}

     	\addplot[color=pink,mark=oplus]
            coordinates {
				(0,0)(100,37.8)(200,56.7)(300,81.1)(400,88.2)(500,88.2)

            };
        \addlegendentry{Voting (Majority)}

		\addplot[color=brown,mark=triangle]
            coordinates {
				(0,0)(100,29.5)(200,53.8)(300,112.3)(400,114.6)(500,114.6)

            };
        \addlegendentry{Voting (Veto)}

		\addplot[color=yellow]
            coordinates {
				(0,0)(100,37.7)(200,47.2)(300,54.1)(400,54.1)(500,54.1)

            };
        \addlegendentry{NN-AT}

        \end{axis}
    \end{tikzpicture}
\vspace{-6.5mm}
\caption{AndroZoo}
\end{subfigure}
\ref{numqueriesblackboxleg}
\vspace{-2.5mm}
\caption{Avg. queries for evasion vs. max. queries.}
\label{figure:numqueriesblackbox}
\vspace{-2.5mm}
\end{figure}

There is a strong correlation between \(n_{max}\) (the maximum number of queries allowed) and the evasion rate. In fact, we also examine the number of queries in Figure~\ref{figure:numqueriesblackbox}, which shows the average number of queries required to evade each defense versus \(n_{max}\). Most defenses can be evaded with less than 100 queries for DREBIN, 150 for SLEIPNIR, and 100 for AndroZoo --- see Appendix~\ref{appendix:extendedresults} for extended results. For other domains that use continuous features (such as image recognition), query attacks may require substantially more queries \mbox{\cite{chen2020hopskipjumpattack}} to achieve attack success compared with domains that use discrete features. This is because perturbations in a discrete feature-space (e.g., 0 to 1) have a greater effect on the final prediction, meaning that fewer of them may be required to achieve evasion, compared with smaller perturbations made per query (e.g., $+0.01$) in a continuous feature-space. For all datasets, attacks against static defenses (or those behaving statically) need fewer queries for evasion. 

Furthermore, experiments for the transferability attack show that all defenses, besides StratDef, exhibit 100\% average RER. For the query attack, the same is true for DREBIN --- even for StratDef --- where adversarial examples achieve 100\% average RER against the defenses. For SLEIPNIR, the average RER for StratDef is 53.9\% as the adversarial examples are not as effective as the transferability attack, while all other defenses exhibit 100\% average RER. For AndroZoo, the average RER is 100\% for all defenses for both attacks, except for StratDef, whose average RER sits at 53.5\%. We demonstrate that our attack strategy remains effective despite the black-box scenario. %

\vspace{-3mm}
\section{Gray-box Results}
\label{sec:grayboxeval}
\vspace{-1.5mm}
Under this attack scenario, we have access to the training data of defenses, which we use to develop substitute models for the transferability attack. Moreover, we have knowledge of the statistical representation of the features, which can be used to enhance the query attack. With greater information about the dataset, we also evaluate each defense against Universal Adversarial Perturbations (UAPs) \cite{moosavi2017universal, labaca2021universal}. 

\vspace{-2mm}

\subsection{Transferability Attack}
\label{sec:grayboxevaltransfattack}
\vspace{-1.5mm}
We evaluate each defense against our attack strategy (Diverse Ensemble with $\Theta_{t} = 0.75$), having shown its superior performance. Our attack strategy, using the Diverse Ensemble but with \(\Theta_{t} = 0.25\) is also included as an alternative, as is the Single DNN strategy \cite{papernot2017practical} as a baseline for comparison. The Ensemble DNN is not included as it has been proven less effective than our attack strategy, and its performance is on-par with the Single DNN strategy. Recall that, under the gray-box scenario, the substitute models for each oracle are equivalent for each attack strategy as they are trained on the same training data as the defenses, rather than a synthetic dataset that is an estimation of \(O\)'s input-output relations.

\begin{figure}[!htbp]
\vspace{-3mm}
\hspace{-0.25cm} 
	\centering
	\begin{subfigure}[b]{0.15\textwidth}
		\begin{tikzpicture}[scale=0.45]
			\begin{axis}[
					ybar,
					enlarge y limits=false,
					width=6.6cm,
            ylabel style=
            {
                yshift=-2mm, 
            }, height=3.6cm,
					ymin=0, ymax=100,
					ylabel={Evasion rate (\%)},
					symbolic x coords={Our Diverse Ensemble (\(\Theta_{t} = 0.75\)), Our Diverse Ensemble (\(\Theta_{t} = 0.25\)), Single DNN},
					xtick=data,
        			xticklabels={Our Diverse Ensemble (\(\Theta_{t} = 0.75\)), Our Diverse Ensemble (\(\Theta_{t} = 0.25\)), Single DNN (Papernot et al.)},
					xticklabel style={text width=2cm,align=center},
					nodes near coords align={horizontal},
					legend pos=north east,
					legend style={nodes={scale=0.4, transform shape}},
					legend to name=grayboxevasionrateleg,
					legend columns=8,
					bar width=0.1cm,
					enlarge x limits={abs=0.6cm}
				]
				\addplot [fill=purple]
				coordinates {
					(Our Diverse Ensemble (\(\Theta_{t} = 0.75\)), 95.1)
(Our Diverse Ensemble (\(\Theta_{t} = 0.25\)), 54.9)
(Single DNN, 55.6)	
				};
				\addplot [fill=cyan]
				coordinates {
					(Our Diverse Ensemble (\(\Theta_{t} = 0.75\)), 91.3)
(Our Diverse Ensemble (\(\Theta_{t} = 0.25\)), 50.5)
(Single DNN, 50.4)	
				};
				\addplot [fill=red]
				coordinates {
					(Our Diverse Ensemble (\(\Theta_{t} = 0.75\)), 92.7)
(Our Diverse Ensemble (\(\Theta_{t} = 0.25\)), 51.4)
(Single DNN, 51.8)
				};
				
				\addplot [fill=green]
				coordinates {
				(Our Diverse Ensemble (\(\Theta_{t} = 0.75\)), 64.5)
(Our Diverse Ensemble (\(\Theta_{t} = 0.25\)), 39.5)
(Single DNN, 43.5)	
				};
				\addplot [fill=pink]
				coordinates {
					(Our Diverse Ensemble (\(\Theta_{t} = 0.75\)), 94.1)
(Our Diverse Ensemble (\(\Theta_{t} = 0.25\)), 52.3)
(Single DNN, 52.7)	
				};
				\addplot [fill=brown]
				coordinates {
					(Our Diverse Ensemble (\(\Theta_{t} = 0.75\)), 57.8)
(Our Diverse Ensemble (\(\Theta_{t} = 0.25\)), 32)
(Single DNN, 34.8)	
				};

                \addplot [fill=yellow]
				coordinates {
					(Our Diverse Ensemble (\(\Theta_{t} = 0.75\)), 90.467)
(Our Diverse Ensemble (\(\Theta_{t} = 0.25\)), 50.056)
(Single DNN, 49.777)

				};

				\legend{DeepMTD, Morphence, MTDeep, StratDef, Voting (Majority), Voting (Veto), NN-AT}%
			\end{axis}
				   
		\end{tikzpicture}
\vspace{-6.5mm}
		\caption{DREBIN}
	\end{subfigure}
	\hspace{0.1cm} 
	\begin{subfigure}[b]{0.15\textwidth}
		\begin{tikzpicture}[scale=0.45]
			\begin{axis}[
					ybar,
					enlarge y limits=false,
					width=6.6cm,
            ylabel style=
            {
                yshift=-2mm, 
            }, height=3.6cm,
					ymin=0, ymax=100,
					ylabel={Evasion rate (\%)},
					symbolic x coords={Our Diverse Ensemble (\(\Theta_{t} = 0.75\)), Our Diverse Ensemble (\(\Theta_{t} = 0.25\)), Single DNN},
					xtick=data,
        			xticklabels={Our Diverse Ensemble (\(\Theta_{t} = 0.75\)), Our Diverse Ensemble (\(\Theta_{t} = 0.25\)), Single DNN (Papernot et al.)},
					xticklabel style={text width=2cm,align=center},
					nodes near coords align={horizontal},
					legend pos=north east,
					legend style={nodes={scale=0.5, transform shape}},
					legend to name=grayboxevasionrateleg,
					legend columns=8,
					bar width=0.1cm,
					enlarge x limits={abs=0.6cm}
				]
				\addplot [fill=purple]
				coordinates {
				(Our Diverse Ensemble (\(\Theta_{t} = 0.75\)), 96.8)
                (Our Diverse Ensemble (\(\Theta_{t} = 0.25\)), 55.3)
                (Single DNN, 57.3)
				};
				\addplot [fill=cyan]
				coordinates {
				(Our Diverse Ensemble (\(\Theta_{t} = 0.75\)), 0)
                (Our Diverse Ensemble (\(\Theta_{t} = 0.25\)), 0)
                (Single DNN, 0)
				};
				\addplot [fill=red]
				coordinates {
				(Our Diverse Ensemble (\(\Theta_{t} = 0.75\)), 96.5)
                (Our Diverse Ensemble (\(\Theta_{t} = 0.25\)), 55.4)
                (Single DNN, 56.3)
				};
				\addplot [fill=green]
				coordinates {
				(Our Diverse Ensemble (\(\Theta_{t} = 0.75\)), 0)
                (Our Diverse Ensemble (\(\Theta_{t} = 0.25\)), 0)
                (Single DNN, 0)
				};
				\addplot [fill=pink]
				coordinates {
					(Our Diverse Ensemble (\(\Theta_{t} = 0.75\)), 0)
                    (Our Diverse Ensemble (\(\Theta_{t} = 0.25\)), 0.2)
                    (Single DNN, 0)
				};
				\addplot [fill=brown]
				coordinates {
				(Our Diverse Ensemble (\(\Theta_{t} = 0.75\)), 0)
                (Our Diverse Ensemble (\(\Theta_{t} = 0.25\)), 0)
                (Single DNN, 0)
				};
				
                \addplot [fill=yellow]
				coordinates {
					(Our Diverse Ensemble (\(\Theta_{t} = 0.75\)), 0)
                    (Our Diverse Ensemble (\(\Theta_{t} = 0.25\)), 0)
                    (Single DNN, 0)

				};

				\legend{DeepMTD, Morphence, MTDeep, StratDef, Voting (Majority), Voting (Veto), NN-AT}%
			\end{axis}
				   
		\end{tikzpicture}
\vspace{-6.5mm}
		\caption{SLEIPNIR}
	\end{subfigure}
	\hspace{0.1cm} 
    \begin{subfigure}[b]{0.15\textwidth}
		\begin{tikzpicture}[scale=0.45]
			\begin{axis}[
					ybar,
					enlarge y limits=false,
					width=6.6cm,
            ylabel style=
            {
                yshift=-2mm, 
            }, height=3.6cm,
					ymin=0, ymax=100,
					ylabel={Evasion rate (\%)},
					symbolic x coords={Our Diverse Ensemble (\(\Theta_{t} = 0.75\)), Our Diverse Ensemble (\(\Theta_{t} = 0.25\)), Single DNN},
					xtick=data,
        			xticklabels={Our Diverse Ensemble (\(\Theta_{t} = 0.75\)), Our Diverse Ensemble (\(\Theta_{t} = 0.25\)), Single DNN (Papernot et al.)},
					xticklabel style={text width=2cm,align=center},
					nodes near coords align={horizontal},
					legend pos=north east,
					legend style={nodes={scale=0.5, transform shape}},
					legend to name=grayboxevasionrateleg,
					legend columns=8,
					bar width=0.1cm,
					enlarge x limits={abs=0.6cm}
				]
				\addplot [fill=purple]
				coordinates {
					(Our Diverse Ensemble (\(\Theta_{t} = 0.75\)), 95.1)
(Our Diverse Ensemble (\(\Theta_{t} = 0.25\)), 75.3)
(Single DNN,64.8)
				};
				\addplot [fill=cyan]
				coordinates {
					(Our Diverse Ensemble (\(\Theta_{t} = 0.75\)), 86.7)
					(Our Diverse Ensemble (\(\Theta_{t} = 0.25\)), 63.5)
					(Single DNN,47.3)					
				};
				\addplot [fill=red]
				coordinates {
					(Our Diverse Ensemble (\(\Theta_{t} = 0.75\)), 93.4)
(Our Diverse Ensemble (\(\Theta_{t} = 0.25\)), 69.6)
(Single DNN,56.4)
				};
				
				\addplot [fill=green]
				coordinates {
					(Our Diverse Ensemble (\(\Theta_{t} = 0.75\)), 0)
					(Our Diverse Ensemble (\(\Theta_{t} = 0.25\)), 0)
					(Single DNN,0)
					
				};
				\addplot [fill=pink]
				coordinates {
					(Our Diverse Ensemble (\(\Theta_{t} = 0.75\)), 0)
					(Our Diverse Ensemble (\(\Theta_{t} = 0.25\)), 0)
					(Single DNN,0)
					
				};
				\addplot [fill=brown]
				coordinates {
					(Our Diverse Ensemble (\(\Theta_{t} = 0.75\)), 0)
					(Our Diverse Ensemble (\(\Theta_{t} = 0.25\)), 0)
					(Single DNN,0)
					
				};

                \addplot [fill=yellow]
				coordinates {
					(Our Diverse Ensemble (\(\Theta_{t} = 0.75\)), 0)
					(Our Diverse Ensemble (\(\Theta_{t} = 0.25\)), 0)
					(Single DNN,0)

				};

				\legend{DeepMTD, Morphence, MTDeep, StratDef, Voting (Majority), Voting (Veto), NN-AT}%
			\end{axis}
				   
		\end{tikzpicture}
\vspace{-6.5mm}
		\caption{AndroZoo}
	\end{subfigure}
	\ref{grayboxevasionrateleg}
	
\vspace{-2.5mm}
	\caption{Evasion rate of attack strategies.}
	\label{figure:grayboxevasionrate}
\vspace{-3mm}
\end{figure}
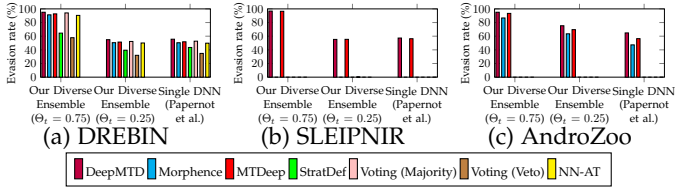

Figure~\ref{figure:grayboxevasionrate} shows the evasion rate for the attack strategies under this threat model. In the less constrained environments with DREBIN and AndroZoo, our attack strategy can achieve a 55+\% evasion rate across the MTDs for DREBIN and mostly 80+\% for AndroZoo. Interestingly, when compared with the black-box transferability attack, the trends and the performance of each defense are similar. However, under a smaller attack surface and hence a more constrained environment with SLEIPNIR, there is greater difficulty in evading the MTDs in general. For the defenses that can be evaded, similar results are observable as in the black-box attack, with a 95+\% evasion rate against the weaker defenses. In contrast, the vast majority of MTDs and defenses offer effective resilience against the attack, with defenses such as Morphence and veto voting remaining quite resilient to adversarial examples. The attack is also unable to evade StratDef, while the evasion rate for majority voting is non-zero, with less than 10 successful adversarial examples in total (which is why some results seem missing --- see Appendix~\ref{appendix:extendedresults} for full results).

In general, 
weaker defenses suffer more against the gray-box attack than the black-box attack, while for attacks against less static defenses, having access to the training data (as in the gray-box threat model) is not as useful in conducting attacks as it may seem. This phenomenon can be attributed to several factors. Firstly, substitute models for the black-box attack are more representative of each oracle, as they are based on direct queries to them when the synthetic dataset \(\Delta\) is constructed. This suggests that, in these circumstances, the substitute models capture the oracle's traits and behavior better, even with a smaller dataset. Consequently, adversarial examples for the black-box substitute models may transfer better to the oracle. Moreover, a larger training set for substitute models --- as in the case of the gray-box transferability attack --- may increase overfitting in the substitute models, resulting in inferior attack performance (see Appendix~\ref{appendix:additionalgrayboxtransf} for experiments evaluating attack performance versus training set size). %

\vspace{-2.5mm}
\subsection{Query Attack}
\label{sec:grayboxqueryattack}
\vspace{-1.5mm}
The gray-box query attack uses the frequency of features in benign samples to determine the order of transplantation. We hypothesize that this should reduce the number of queries needed and improve attack performance.

\begin{figure}[!htbp]
\vspace{-2.5mm}
\hspace{-0.25cm}
\centering
\begin{subfigure}[b]{0.15\textwidth}
    \begin{tikzpicture}[scale=0.45]
        \begin{axis}[
            xlabel={Max. number of queries (\(n_{max}\))},
            ylabel={Evasion rate (\%)}, 
            xmin=0, xmax=500,
            ymin=0, ymax=100,
            xtick={0,100,200,300,400,500},
            ymajorgrids=false,
            legend pos= south west,
            legend style={nodes={scale=0.5, transform shape}},
            cycle list name=color list,
            legend to name=evasionrategrayboxqbasedleg,
            legend columns=8,
            height=3.6cm,
            width=6.6cm,
            ylabel style=
            {
                yshift=-2mm, 
            }
        ]

		\addplot[color=purple,mark=diamond]
            coordinates {
				(0,0)(100, 37.118)(200, 85.59)(300, 96.507)(400, 96.507)(500, 96.507)

            };
        \addlegendentry{DeepMTD}
		
		\addplot[color=cyan,mark=pentagon]
            coordinates {
				(0,0)(100, 43.668)(200, 81.223)(300, 98.253)(400, 99.127)(500, 99.127)

		};
        \addlegendentry{Morphence}

		\addplot[color=red,mark=+]
            coordinates {
				(0,0)(100, 39.301)(200, 79.913)(300, 96.943)(400, 96.943)(500, 96.943)

            };
        \addlegendentry{MTDeep}

        \addplot[color=green,mark=x]
            coordinates {
				(0,0)(100, 87.336)(200, 87.336)(300, 87.336)(400, 87.336)(500, 87.336)

            };
        \addlegendentry{StratDef}

     	\addplot[color=pink,mark=oplus]
            coordinates {
				(0,0)(100, 93.886)(200, 96.507)(300, 96.507)(400, 96.507)(500, 96.507)

            };
        \addlegendentry{Voting (Majority)}

		\addplot[color=brown,mark=triangle]
            coordinates {
				(0,0)(100, 15.284)(200, 63.755)(300, 79.476)(400, 84.716)(500, 84.716)

            };
        \addlegendentry{Voting (Veto)}

		\addplot[color=yellow]
            coordinates {
				(0,0)(100, 95.633)(200, 99.127)(300, 99.127)(400, 99.127)(500, 99.127)

            };
        \addlegendentry{NN-AT}

        \end{axis}
    \end{tikzpicture}
\vspace{-6.5mm}
\caption{DREBIN}
\end{subfigure}
\hspace{0.1cm} 
\begin{subfigure}[b]{0.15\textwidth}
    \begin{tikzpicture}[scale=0.45]
        \begin{axis}[
             xlabel={Max. number of queries (\(n_{max}\))},
            ylabel={Evasion rate (\%)}, 
            xmin=0, xmax=500,
            ymin=0, ymax=100,
            xtick={0,100,200,300,400,500},
            ymajorgrids=false,
            legend pos= south west,
            legend style={nodes={scale=0.5, transform shape}},
            cycle list name=color list,
            legend to name=evasionrategrayboxqbasedleg,
            legend columns=8,
            height=3.6cm,
            width=6.6cm,
            ylabel style=
            {
                yshift=-2mm, 
            }
        ]
		\addplot[color=purple,mark=diamond]
            coordinates {
				(0,0)(100, 29.13)(200, 77.391)(300, 85.217)(400, 88.696)(500, 90.435)

            };
        \addlegendentry{DeepMTD}
		
		\addplot[color=cyan,mark=pentagon]
            coordinates {
				(0,0)(100, 15.652)(200, 66.957)(300, 72.609)(400, 84.783)(500, 88.261)

            };
        \addlegendentry{Morphence}

		\addplot[color=red,mark=+]
            coordinates {
				(0,0)(100, 30.435)(200, 85.217)(300, 85.652)(400, 86.087)(500, 89.13)

            };
        \addlegendentry{MTDeep}

         \addplot[color=green,mark=x]
            coordinates {
				(0,0)(100, 50)(200, 89.565)(300, 90.435)(400, 90.87)(500, 90.87)

            };
        \addlegendentry{StratDef}
        
        		\addplot[color=pink,mark=oplus]
            coordinates {
				(0,0)(100, 30.87)(200, 87.391)(300, 89.565)(400, 90)(500, 90)

            };
        \addlegendentry{Voting (Majority)}

		\addplot[color=brown,mark=triangle]
            coordinates {
				(0,0)(100, 9.565)(200, 23.043)(300, 29.13)(400, 29.13)(500, 29.13)

            };
        \addlegendentry{Voting (Veto)}

		\addplot[color=yellow]
            coordinates {
				(0,0)(100, 49.565)(200, 85.217)(300, 87.826)(400, 90)(500, 90)

            };
        \addlegendentry{NN-AT}

        \end{axis}
    \end{tikzpicture}
\vspace{-6.5mm}
\caption{SLEIPNIR}
\end{subfigure}
\hspace{0.1cm} 
\begin{subfigure}[b]{0.15\textwidth}
    \begin{tikzpicture}[scale=0.45]
        \begin{axis}[
            xlabel={Maximum number of queries (\(n_{max}\))},
            ylabel={Evasion rate (\%)}, 
            xmin=0, xmax=500,
            ymin=0, ymax=100,
            xtick={0,100,200,300,400,500},
            ymajorgrids=false,
            legend pos= south west,
            legend style={nodes={scale=0.5, transform shape}},
            cycle list name=color list,
            legend to name=evasionrategrayboxqbasedleg,
            legend columns=8,
            height=3.6cm,
            width=6.6cm,
            ylabel style=
            {
                yshift=-2mm, 
            }
        ]

		\addplot[color=purple,mark=diamond]
            coordinates {
				(0,0)(100,59.2)(200,64.1)(300,82.1)(400,95.1)(500,100)

            };
        \addlegendentry{DeepMTD}
        
		\addplot[color=cyan,mark=pentagon]
            coordinates {
				(0,0)(100,43)(200,53.5)(300,68.4)(400,81.1)(500,94.3)
		};
        \addlegendentry{Morphence}

		\addplot[color=red,mark=+]
            coordinates {
				(0,0)(100,62.4)(200,65.6)(300,76.5)(400,94.1)(500,97.3)

            };
        \addlegendentry{MTDeep}

        \addplot[color=green,mark=x]
            coordinates {
				(0,0)(100,89.5)(200,91.7)(300,93)(400,94.7)(500,95.6)

            };
        \addlegendentry{StratDef}

     	\addplot[color=pink,mark=oplus]
            coordinates {
				(0,0)(100,75.4)(200,84.2)(300,87.7)(400,89)(500,89)

            };
        \addlegendentry{Voting (Majority)}

		\addplot[color=brown,mark=triangle]
            coordinates {
				(0,0)(100,56.6)(200,64.3)(300,69.7)(400,69.7)(500,69.7)

            };
        \addlegendentry{Voting (Veto)}

		\addplot[color=yellow]
            coordinates {
				(0,0)(100,71.4)(200,79.3)(300,89.4)(400,93.8)(500,94.7)

            };
        \addlegendentry{NN-AT}

        \end{axis}
    \end{tikzpicture}
\vspace{-6.5mm}
\caption{AndroZoo}
\end{subfigure}
\ref{evasionrategrayboxqbasedleg}
\vspace{-2.5mm}
\caption{Evasion rate vs. max number of queries.}
\label{figure:evasionrategrayboxqbased}
\vspace{-2.5mm}
\end{figure}
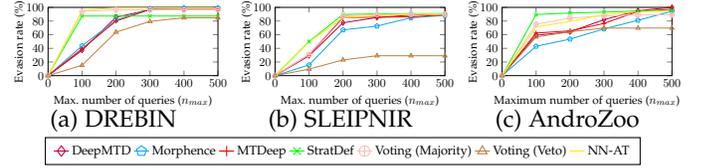

Figure~\ref{figure:evasionrategrayboxqbased} shows the evasion rate versus \(n_{max}\). Similar to the black-box query attack, there is a strong correlation between \(n_{max}\) and the evasion rate, which reaches up to 99\% for DREBIN, 94\% for SLEIPNIR, and 100\% for AndroZoo (with an average evasion rate of 84.6\%, 91\%, and 79.6\% respectively, across all defenses and maximum query sizes). However, the average evasion rate is also higher than the black-box query attack (by \(\approx 20\%\)) as well as the gray-box transferability attack, with nearly all defenses being evaded even in the more constrained environment of SLEIPNIR. As has been a common theme, defenses exhibiting mostly static behavior perform worse, with an evasion rate of 90+\%. This shows that a heuristically-driven query attack method specifically designed for this domain can achieve better evasion against the MTDs, even in more constrained environments. Interestingly, veto voting still offers robustness for SLEIPNIR, although we achieve a higher evasion rate with this attack than previously.

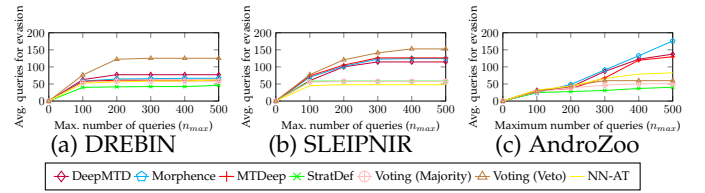
\begin{figure}[!htbp]
\vspace{-2.5mm}
\hspace{-0.25cm} 
\centering
\begin{subfigure}[b]{0.15\textwidth}
    \begin{tikzpicture}[scale=0.45]
        \begin{axis}[
            xlabel={Max. number of queries (\(n_{max}\))},
            ylabel={Avg. queries for evasion}, 
            xmin=0, xmax=500,
            ymin=0, ymax=200,
            xtick={0,100,200,300,400,500},
            ymajorgrids=false,
            legend pos= south west,
            legend style={nodes={scale=0.5, transform shape}},
            cycle list name=color list,
            legend to name=numqueriesgrayboxleg,
            legend columns=8,
            height=3.6cm,
            width=6.6cm,
            ylabel style=
            {
                yshift=-2mm, 
            }
        ]

		\addplot[color=purple,mark=diamond]
            coordinates {
            (0,0)
			(100, 62.8)
			(200, 77.11)
			(300, 77.11)
			(400, 77.11)
			(500, 77.11)

            };
        \addlegendentry{DeepMTD}
		
		\addplot[color=cyan,mark=pentagon]
            coordinates {
            (0,0)
			(100, 58.53)
(200, 64.49)
(300, 65.51)
(400, 66.65)
(500, 66.65)

		};
        \addlegendentry{Morphence}

		\addplot[color=red,mark=+]
            coordinates {
            (0,0)
			(100, 57.57)
(200, 60.91)
(300, 60.91)
(400, 60.91)
(500, 60.91)

            };
        \addlegendentry{MTDeep}

        \addplot[color=green,mark=x]
            coordinates {
            (0,0)
			(100, 40)
			(200, 41.58)
			(300, 42.4)
			(400, 42.4)
			(500, 45.98)
			
            };
        \addlegendentry{StratDef}

     	\addplot[color=pink,mark=oplus]
            coordinates {
            (0,0)
			(100, 54.55)
(200, 58.01)
(300, 58.01)
(400, 58.01)
(500, 58.01)

            };
        \addlegendentry{Voting (Majority)}

		\addplot[color=brown,mark=triangle]
            coordinates {
            (0,0)
			(100, 76.38)
			(200, 122.43)
			(300, 125.33)
			(400, 125.33)
			(500, 125.33)

            };
        \addlegendentry{Voting (Veto)}

		\addplot[color=yellow]
            coordinates {
				
            (0,0)
			(100, 52.46)
			(200, 55.99)
			(300, 58.59)
			(400, 58.59)
			(500, 62.03)

            };
        \addlegendentry{NN-AT}

        \end{axis}
    \end{tikzpicture}
\vspace{-6.5mm}
\caption{DREBIN}
\end{subfigure}
\hspace{0.1cm} 
\begin{subfigure}[b]{0.15\textwidth}
    \begin{tikzpicture}[scale=0.45]
        \begin{axis}[
            xlabel={Max. number of queries (\(n_{max}\))},
            ylabel={Avg. queries for evasion}, 
            xmin=0, xmax=500,
            ymin=0, ymax=200,
            xtick={0,100,200,300,400,500},
            ymajorgrids=false,
            legend pos= south west,
            legend style={nodes={scale=0.5, transform shape}},
            cycle list name=color list,
            legend to name=numqueriesgrayboxleg,
            legend columns=8,
            height=3.6cm,
            width=6.6cm,
            ylabel style=
            {
                yshift=-2mm, 
            }
        ]
		\addplot[color=purple,mark=diamond]
            coordinates {
            (0,0)
			(100, 60.02)
			(200, 100.23)
			(300, 114.54)
			(400, 114.54)
			(500, 114.54)
			
            };
        \addlegendentry{DeepMTD}
		
		\addplot[color=cyan,mark=pentagon]
            coordinates {
            (0,0)
			(100, 70.16)
(200, 101.72)
(300, 122.74)
(400, 124.4)
(500, 124.4)

            };
        \addlegendentry{Morphence}

		\addplot[color=red,mark=+]
            coordinates {
            (0,0)
			(100, 73.76)
			(200, 105.44)
			(300, 126.41)
			(400, 126.41)
			(500, 126.41)
			
            };
        \addlegendentry{MTDeep}

         \addplot[color=green,mark=x]
            coordinates {
            (0,0)
			(100, 58.2)
			(200, 58.2)
			(300, 58.2)
			(400, 58.2)
			(500, 58.2)

            };
        \addlegendentry{StratDef}
        
        		\addplot[color=pink,mark=oplus]
            coordinates {
            (0,0)
			(100, 55.66)
			(200, 57.42)
			(300, 57.42)
			(400, 57.42)
			(500, 57.42)

            };
        \addlegendentry{Voting (Majority)}

		\addplot[color=brown,mark=triangle]
            coordinates {
            (0,0)
			(100, 77.34)
			(200, 120.68)
			(300, 141.14)
			(400, 152.75)
			(500, 152.75)

            };
        \addlegendentry{Voting (Veto)}

		\addplot[color=yellow]
            coordinates {
            (0,0)
			(100, 45.21)
			(200, 47.96)
			(300, 47.96)
			(400, 47.96)
			(500, 47.96)

            };
        \addlegendentry{NN-AT}

        \end{axis}
    \end{tikzpicture}
\vspace{-6.5mm}
\caption{SLEIPNIR}
\end{subfigure}
\hspace{0.1cm} 
\begin{subfigure}[b]{0.15\textwidth}
    \begin{tikzpicture}[scale=0.45]
        \begin{axis}[
            xlabel={Maximum number of queries (\(n_{max}\))},
            ylabel={Avg. queries for evasion}, 
            xmin=0, xmax=500,
            ymin=0, ymax=200,
            xtick={0,100,200,300,400,500},
            ymajorgrids=false,
            legend pos= south west,
            legend style={nodes={scale=0.5, transform shape}},
            cycle list name=color list,
            legend to name=avgqueriesgrayboxqbasedleg,
            legend columns=8,
            height=3.6cm,
            width=6.6cm,
            ylabel style=
            {
                yshift=-2mm, 
            }
        ]
        
		\addplot[color=purple,mark=diamond]
            coordinates {
				(0,0)(100,31.1)(200,41.3)(300,86.8)(400,122.7)(500,137.4)

            };
        \addlegendentry{DeepMTD}
        
		\addplot[color=cyan,mark=pentagon]
            coordinates {
				(0,0)(100,24.6)(200,48.4)(300,91.1)(400,132.7)(500,176.1)
		};
        \addlegendentry{Morphence}

		\addplot[color=red,mark=+]
            coordinates {
				(0,0)(100,30.6)(200,36.7)(300,67.7)(400,119.5)(500,129.7)

            };
        \addlegendentry{MTDeep}

        \addplot[color=green,mark=x]
            coordinates {
				(0,0)(100,24.9)(200,27.7)(300,31)(400,36.7)(500,40.5)

            };
        \addlegendentry{StratDef}

     	\addplot[color=pink,mark=oplus]
            coordinates {
				(0,0)(100,27.3)(200,38.6)(300,46.7)(400,50.7)(500,50.7)

            };
        \addlegendentry{Voting (Majority)}

		\addplot[color=brown,mark=triangle]
            coordinates {
				(0,0)(100,28.1)(200,44.5)(300,59.7)(400,59.7)(500,59.7)

            };
        \addlegendentry{Voting (Veto)}

		\addplot[color=yellow]
            coordinates {
				(0,0)(100,32.3)(200,43.7)(300,66.1)(400,79)(500,82.5)

            };
        \addlegendentry{NN-AT}

        \end{axis}
    \end{tikzpicture}
\vspace{-6.5mm}
\caption{AndroZoo}
\end{subfigure}
\ref{numqueriesgrayboxleg}
\vspace{-2.5mm}
\caption{Avg. queries for evasion vs. max. queries.}
\label{figure:numqueriesgraybox}
\vspace{-2.5mm}
\end{figure}

The gray-box query attack also performs better than the black-box query attack with respect to the queries required to achieve this level of evasion. From Figure~\ref{figure:numqueriesgraybox}, we observe that fewer than 80 queries for DREBIN are enough to cripple most defenses. For SLEIPNIR, there is a further 44\% decrease in the average number of queries compared with the black-box attack. Meanwhile, AndroZoo also presents a reduction in the number of queries required to achieve evasion. Therefore, on the balance of numbers, the gray-box query attack outperforms the black-box attack considering the datasets and the number of queries permitted. Moreover, the average RER for the gray-box query attack for DREBIN is 100\% for all defenses but 0.2\% higher than the black-box query attack on StratDef for SLEIPNIR. The average RER for AndroZoo similar to its black-box counterpart, with a 0.4\% reduction for StratDef, but 100\% for other defenses.

\vspace{-2mm}
\subsection{Universal Adversarial Perturbations}
\label{sec:grayboxuaps}
\vspace{-1.5mm}
Recent research has identified universal adversarial perturbations (UAPs) as a cost-effective technique for producing adversarial examples \cite{moosavi2017universal, labaca2021universal}. With UAPs, a set of perturbations can be applied to multiple malware samples to generate adversarial examples. In other words, sets of perturbations that are known to cause input samples to cross the decision boundary are reused across several malware samples. We evaluate each defense against adversarial examples generated with UAPs to compare with our attack strategies. The UAPs are derived from the dataset that the gray-box attacker has access to. Using the adversarial examples produced by the gray-box transferability attack, we examine if a set of perturbations has been reused exactly to produce adversarial examples from its original samples. In such cases, the adversarial examples are deemed to have been generated with UAPs.

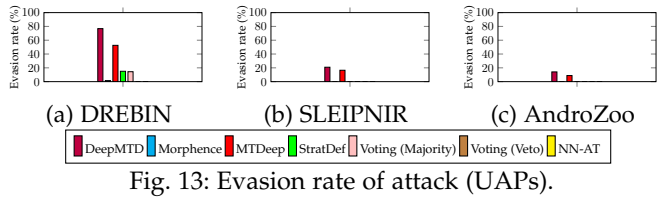
\begin{figure}[!htbp]
\vspace{-2.5mm}
\hspace{-0.25cm} 
	\centering
	\begin{subfigure}[b]{0.15\textwidth}
		\begin{tikzpicture}[scale=0.455]
			\begin{axis}[
					ybar,
					enlarge y limits=false,
					ymin=0, ymax=100,
					ylabel={Evasion rate (\%)},
					symbolic x coords={Our Diverse Ensemble (\(\Theta_{t} = 0.25\))},
					xtick=data,
					xticklabel style={text width=3cm,align=center},
				    xticklabels={},
					legend pos=north east,
					legend style={nodes={scale=0.5, transform shape}},
					legend to name=grayboxuapsevasionleg,
					legend columns=8,
                     height=3.6cm,
                    width=6.2cm,
            ylabel style=
            {
                yshift=-2mm, 
            },
					bar width=0.15cm
				]
				\addplot [fill=purple]
				coordinates {
                    (Our Diverse Ensemble (\(\Theta_{t} = 0.25\)),76.8)
				};
				\addplot [fill=cyan]
				coordinates {
(Our Diverse Ensemble (\(\Theta_{t} = 0.25\)),1.5)
				};
				\addplot [fill=red]
				coordinates {
(Our Diverse Ensemble (\(\Theta_{t} = 0.25\)),52.6)
				};
				
				\addplot [fill=green]
				coordinates {
(Our Diverse Ensemble (\(\Theta_{t} = 0.25\)),15)
				};
				\addplot [fill=pink]
				coordinates {
(Our Diverse Ensemble (\(\Theta_{t} = 0.25\)),14.4)
				};
				\addplot [fill=brown]
				coordinates {
					(Our Diverse Ensemble (\(\Theta_{t} = 0.25\)),0)
				};
				
                \addplot [fill=yellow]
				coordinates {
					(Our Diverse Ensemble (\(\Theta_{t} = 0.25\)),0)
				};

				\legend{DeepMTD, Morphence, MTDeep, StratDef, Voting (Majority), Voting (Veto), NN-AT}%
			\end{axis}
				   
		\end{tikzpicture}
\vspace{-1mm}
		\caption{DREBIN}
	\end{subfigure}
	\hspace{0.1cm} 
	\begin{subfigure}[b]{0.15\textwidth}
		\begin{tikzpicture}[scale=0.455]
			\begin{axis}[
					ybar,
					enlarge y limits=false,
					ymin=0, ymax=100,
					ylabel={Evasion rate (\%)},
					symbolic x coords={Our Diverse Ensemble (\(\Theta_{t} = 0.25\))},
					xtick=data,
					xticklabel style={text width=3cm,align=center},
				    xticklabels={},
					legend pos=north east,
					legend style={nodes={scale=0.5, transform shape}},
					legend to name=grayboxuapsevasionleg,
					legend columns=8,
                     height=3.6cm,
                   width=6.2cm,
            ylabel style=
            {
                yshift=-2mm, 
            },
					bar width=0.15cm
				]
				\addplot [fill=purple]
				coordinates {
(Our Diverse Ensemble (\(\Theta_{t} = 0.25\)),20.9)

				};
				\addplot [fill=cyan]
				coordinates {
(Our Diverse Ensemble (\(\Theta_{t} = 0.25\)),0)

				};
				\addplot [fill=red]
				coordinates {
(Our Diverse Ensemble (\(\Theta_{t} = 0.25\)),16.5)

				};
				\addplot [fill=green]
				coordinates {
(Our Diverse Ensemble (\(\Theta_{t} = 0.25\)),0)

				};
				\addplot [fill=pink]
				coordinates {
					(Our Diverse Ensemble (\(\Theta_{t} = 0.25\)),0)
					
				};
				\addplot [fill=brown]
				coordinates {
					(Our Diverse Ensemble (\(\Theta_{t} = 0.25\)),0)

				};
				
                \addplot [fill=yellow]
				coordinates {
					(Our Diverse Ensemble (\(\Theta_{t} = 0.25\)),0)

				};

				\legend{DeepMTD, Morphence, MTDeep, StratDef, Voting (Majority), Voting (Veto), NN-AT}%

			\end{axis}
				   
		\end{tikzpicture}
\vspace{-1mm}
		\caption{SLEIPNIR}
	\end{subfigure}
	\hspace{0.1cm} 
	\begin{subfigure}[b]{0.15\textwidth}
		\begin{tikzpicture}[scale=0.455]
			\begin{axis}[
					ybar,
					enlarge y limits=false,
					ymin=0, ymax=100,
					ylabel={Evasion rate (\%)},
					symbolic x coords={Our Diverse Ensemble (\(\Theta_{t} = 0.25\))},
					xtick=data,
					xticklabel style={text width=3cm,align=center},
				    xticklabels={},
					legend pos=north east,
					legend style={nodes={scale=0.5, transform shape}},
					legend to name=grayboxuapsevasionleg,
					legend columns=8,
                     height=3.6cm,
                   width=6.2cm,
            ylabel style=
            {
                yshift=-2mm, 
            },
					bar width=0.15cm
				]
				\addplot [fill=purple]
				coordinates {
(Our Diverse Ensemble (\(\Theta_{t} = 0.25\)),14.1)

				};
				\addplot [fill=cyan]
				coordinates {
(Our Diverse Ensemble (\(\Theta_{t} = 0.25\)),0)

				};
				\addplot [fill=red]
				coordinates {
(Our Diverse Ensemble (\(\Theta_{t} = 0.25\)),8.82)

				};
				\addplot [fill=green]
				coordinates {
(Our Diverse Ensemble (\(\Theta_{t} = 0.25\)),0)

				};
				\addplot [fill=pink]
				coordinates {
					(Our Diverse Ensemble (\(\Theta_{t} = 0.25\)),0)
					
				};
				\addplot [fill=brown]
				coordinates {
					(Our Diverse Ensemble (\(\Theta_{t} = 0.25\)),0)

				};
				
                \addplot [fill=yellow]
				coordinates {
					(Our Diverse Ensemble (\(\Theta_{t} = 0.25\)),0)

				};

				\legend{DeepMTD, Morphence, MTDeep, StratDef, Voting (Majority), Voting (Veto), NN-AT}%

			\end{axis}
				   
		\end{tikzpicture}
\vspace{-1mm}
		\caption{AndroZoo}
	\end{subfigure}
	\ref{grayboxuapsevasionleg}
\vspace{-2.5mm}
	\caption{Evasion rate of attack (UAPs).}
	\label{figure:grayboxuapsevasion}
\vspace{-3mm}
\end{figure}

Figure~\ref{figure:grayboxuapsevasion} shows the evasion rate of adversarial examples generated with UAPs. There is significantly less evasion compared with the transferability and query attacks, with an average evasion rate of 20\% for DREBIN, 4.7\% for SLEIPNIR, and 3.3\% for AndroZoo (with peak evasion rates of 77\%, 21\%, and 14.1\%, respectively). DeepMTD and MTDeep are evaded the most, especially for SLEIPNIR and AndroZoo. Similar to the transferability attack for these datasets, most defenses are not evaded by adversarial examples generated through UAPs, with a significantly lower average evasion rate overall.

\vspace{-3mm}
\section{Beyond Adversarial Robustness}
\label{sec:widerpicture}
\vspace{-1.5mm}
Some defenses consistently perform well against black-box and gray-box attacks. Defenses like Morphence, StratDef, and veto voting appear more resilient than others. Hence, it may seem appealing to deploy these defenses. However, metrics beyond adversarial robustness need to be considered, especially in the malware detection domain where the false positive rate (FPR) must remain low \cite{grosse2017adversarial, stokes2017attack, yang2017malware, 10.1145/3484491}. A high FPR means a less reliable and more frustrating service as analysts are flooded with false alarms and users find their benign queries misclassified as malware. Therefore, we simulate different system conditions with varying degrees of adversity to understand how each defense performs under these scenarios while considering other metrics. To do this, each defense is queried over 1000 times with different proportions of adversarial examples representing the attack intensity (\(q\)). For example, at \(q=0.5\), half of the queries to the defense are adversarial, while the remaining proportion is an equal number of benign and non-adversarial malware input samples. The adversarial examples are those produced previously under the gray-box transferability attack to maintain uniformity, as the black-box attack produces different adversarial examples for each oracle. We evaluate each defense with \(0.1 \leq q \leq 0.9\) to avoid the unrealistic cases of \(q=0\) (where there are no adversarial queries) and \(q=1\) (where all queries are adversarial). 

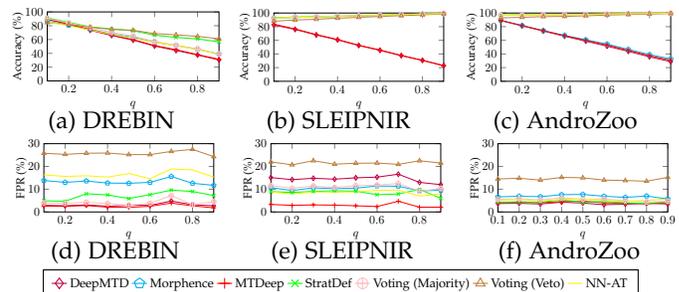
\begin{figure}[!htbp]
\vspace{-2.5mm}
\hspace{-0.25cm} 
	\centering
	\begin{subfigure}[b]{0.15\textwidth}
		\begin{tikzpicture}[scale=0.455]
			\begin{axis}[
					xlabel={\(q\)},
					ylabel={Accuracy (\%)}, 
					xmin=0.1, xmax=0.9,
					ymin=0, ymax=100,
					xtick={0.2,0.4,0.6,0.8},
					ytick={0,20,40,60,80,100},
					ymajorgrids=false,
					legend pos=south west,
					legend style={nodes={scale=0.5, transform shape}},
					cycle list name=color list,
					legend to name=grayboxfullfprlegend,
					legend columns=8,
            height=3.6cm,
           width=6.6cm,
            ylabel style=
            {
                yshift=-2mm, 
            }
				]

				\addplot[color=purple,mark=diamond]
				coordinates {
					(0.1,89.3)(0.2,82.1)(0.3,73.5)(0.4,65.9)(0.5,59.2)(0.6,50.6)(0.7,44.1)(0.8,37.6)(0.9,30.5)
				};
				\addlegendentry{DeepMTD}
						
				\addplot[color=cyan,mark=pentagon]
				coordinates {
					(0.1,86.7)(0.2,80.7)(0.3,74.7)(0.4,68.2)(0.5,63.1)(0.6,56.2)(0.7,51.4)(0.8,45.8)(0.9,39)
				};
				\addlegendentry{Morphence}

				\addplot[color=red,mark=+]
				coordinates {
					(0.1,89.7)(0.2,82.6)(0.3,74.4)(0.4,66.4)(0.5,59.8)(0.6,51.2)(0.7,45.3)(0.8,38.1)(0.9,31.4)
				};
				\addlegendentry{MTDeep}
				
				\addplot[color=green,mark=x]
				coordinates {
					(0.1,91.5)(0.2,84.9)(0.3,78.3)(0.4,75.2)(0.5,72.8)(0.6,66.3)(0.7,63.1)(0.8,60.9)(0.9,56.6)
				};
				\addlegendentry{StratDef}
				
				\addplot[color=pink,mark=oplus]
				coordinates {
					(0.1,90.7)(0.2,83.9)(0.3,77.4)(0.4,70.4)(0.5,65)(0.6,57.2)(0.7,51.7)(0.8,46.6)(0.9,38.9)
				};
				\addlegendentry{Voting (Majority)}
				
				\addplot[color=brown,mark=triangle]
				coordinates {
					(0.1,84.5)(0.2,82)(0.3,77.8)(0.4,74.8)(0.5,73.4)(0.6,68.7)(0.7,66.8)(0.8,64.6)(0.9,61.1)
				};
				\addlegendentry{Voting (Veto)}
            \addplot[color=yellow]
                coordinates {
                (0.1,85.7)(0.2,79.9)(0.3,73.9)(0.4,67.4)(0.5,61.9)(0.6,56)(0.7,51.2)(0.8,45.3)(0.9,39.2)
                };
            \addlegendentry{NN-AT}

			\end{axis}
		\end{tikzpicture}
		\vspace{-6.5mm}
		\caption{DREBIN}
	\end{subfigure}
	\hspace{0.1cm} 
	\begin{subfigure}[b]{0.15\textwidth}
		\begin{tikzpicture}[scale=0.45]
			\begin{axis}[
					xlabel={\(q\)},
					ylabel={Accuracy (\%)}, 
					xmin=0.1, xmax=0.9,
					ymin=0, ymax=100,
					xtick={0.2,0.4,0.6,0.8},
					ytick={0,20,40,60,80,100},
					ymajorgrids=false,
					legend pos=south west,
					legend style={nodes={scale=0.5, transform shape}},
					cycle list name=color list,
					legend to name=grayboxfullfprlegend,
					legend columns=8,
            height=3.6cm,
            width=6.6cm,
            ylabel style=
            {
                yshift=-2mm, 
            }
				]
				
				\addplot[color=purple,mark=diamond]
				coordinates {
					(0.1,82.4)(0.2,76)(0.3,67.9)(0.4,60.7)(0.5,52.5)(0.6,45.3)(0.7,37.6)(0.8,30.7)(0.9,22.9)
				};
				\addlegendentry{DeepMTD}
						
				\addplot[color=cyan,mark=pentagon]
				coordinates {
					(0.1,92.1)(0.2,93.5)(0.3,94.2)(0.4,95.2)(0.5,95.8)(0.6,96.5)(0.7,97.3)(0.8,98.4)(0.9,99.1)
				};
				\addlegendentry{Morphence}

				\addplot[color=red,mark=+]
				coordinates {
					(0.1,83.6)(0.2,76.7)(0.3,68.4)(0.4,61.2)(0.5,52.7)(0.6,45.8)(0.7,37.8)(0.8,30.6)(0.9,22.8)
				};
				\addlegendentry{MTDeep}
				
				\addplot[color=green,mark=x]
				coordinates {
					(0.1,93.2)(0.2,94.2)(0.3,94.6)(0.4,95)(0.5,96)(0.6,97)(0.7,97.9)(0.8,98.5)(0.9,99.4)
				};
				\addlegendentry{StratDef}
				
				\addplot[color=pink,mark=oplus]
				coordinates {
					(0.1,92.2)(0.2,93.5)(0.3,94.3)(0.4,95.4)(0.5,95.8)(0.6,96.7)(0.7,97.4)(0.8,98.5)(0.9,99.2)
				};
				\addlegendentry{Voting (Majority)}
				
				\addplot[color=brown,mark=triangle]
				coordinates {
					(0.1,88)(0.2,89.8)(0.3,90.8)(0.4,92.5)(0.5,93.6)(0.6,95)(0.7,96.2)(0.8,97.3)(0.9,98.8)
				};
				\addlegendentry{Voting (Veto)}
				
        \addplot[color=yellow]
            coordinates {
            (0.1,92.6)(0.2,93.9)(0.3,94.6)(0.4,95.5)(0.5,96)(0.6,96.9)(0.7,97.6)(0.8,98.6)(0.9,99.2)
            };
        \addlegendentry{NN-AT}

			\end{axis}
		\end{tikzpicture}
		\vspace{-6.5mm}
		\caption{SLEIPNIR}
	\end{subfigure}
	\hspace{0.1cm} 
	\begin{subfigure}[b]{0.15\textwidth}
		\begin{tikzpicture}[scale=0.45]
			\begin{axis}[
					xlabel={\(q\)},
					ylabel={Accuracy (\%)}, 
					xmin=0.1, xmax=0.9,
					ymin=0, ymax=100,
					xtick={0.2,0.4,0.6,0.8},
					ytick={0,20,40,60,80,100},
					ymajorgrids=false,
					legend pos=south west,
					legend style={nodes={scale=0.5, transform shape}},
					cycle list name=color list,
					legend to name=grayboxfullfprlegend,
					legend columns=8,
                    height=3.6cm,
                    width=6.6cm,
            ylabel style=
            {
                yshift=-2mm, 
            }
				]

				\addplot[color=purple,mark=diamond]
				coordinates {
					(0.1,89)(0.2,81.3)(0.3,73.7)(0.4,66.1)(0.5,58.8)(0.6,51.5)(0.7,43.9)(0.8,35.9)(0.9,28.8)
				};
				\addlegendentry{DeepMTD}
							
				\addplot[color=cyan,mark=pentagon]
				coordinates {
					(0.1,89)(0.2,81.7)(0.3,74.3)(0.4,67.2)(0.5,60.9)(0.6,54.5)(0.7,46.7)(0.8,39.1)(0.9,32.8)
				};
				\addlegendentry{Morphence}

				\addplot[color=red,mark=+]
				coordinates {
					(0.1,89.1)(0.2,81.5)(0.3,74)(0.4,66.7)(0.5,59.8)(0.6,53.1)(0.7,45.2)(0.8,37.5)(0.9,30.7)
				};
				\addlegendentry{MTDeep}
					
				\addplot[color=green,mark=x]
				coordinates {
					(0.1,97)(0.2,97.5)(0.3,97.7)(0.4,97.8)(0.5,98.3)(0.6,98.7)(0.7,99.1)(0.8,99.4)(0.9,99.7)
				};
				\addlegendentry{StratDef}
					
				\addplot[color=pink,mark=oplus]
				coordinates {
					(0.1,96.6)(0.2,97)(0.3,97.4)(0.4,97.5)(0.5,98)(0.6,98.5)(0.7,99)(0.8,99.4)(0.9,99.7)
				};
				\addlegendentry{Voting (Majority)}
					
				\addplot[color=brown,mark=triangle]
				coordinates {
					(0.1,92.9)(0.2,93.7)(0.3,94.7)(0.4,95.1)(0.5,95.9)(0.6,97)(0.7,97.8)(0.8,98.5)(0.9,99.2)
				};
				\addlegendentry{Voting (Veto)}
            \addplot[color=yellow]
                coordinates {
					(0.1,96.7)(0.2,97.1)(0.3,97.5)(0.4,97.7)(0.5,98.1)(0.6,98.6)(0.7,99)(0.8,99.3)(0.9,99.7)

                };
            \addlegendentry{NN-AT}

			\end{axis}
		\end{tikzpicture}
		\vspace{-6.5mm}
		\caption{AndroZoo}
	\end{subfigure}

\hspace{-0.15cm} 	
	\begin{subfigure}[b]{0.15\textwidth}
		\begin{tikzpicture}[scale=0.45]
			\begin{axis}[
					xlabel={\(q\)},
					ylabel={FPR (\%)}, 
					xmin=0.1, xmax=0.9,
					ymin=0, ymax=30,
					xtick={0.2,0.4,0.6,0.8},
					ytick={0,10,20,30,40},
					ymajorgrids=false,
					legend pos=south west,
					legend style={nodes={scale=0.5, transform shape}},
					cycle list name=color list,
					legend to name=grayboxfullfprlegend,
					legend columns=8,
            height=3.6cm,
           width=6.6cm,
            ylabel style=
            {
                yshift=-2mm, 
            }
				]

				\addplot[color=purple,mark=diamond]
				coordinates {
					(0.1,2.9)(0.2,2.7)(0.3,3.1)(0.4,2.5)(0.5,2.9)(0.6,2.9)(0.7,4.8)(0.8,3.2)(0.9,2.7)
				};
				\addlegendentry{DeepMTD}
							
				\addplot[color=cyan,mark=pentagon]
				coordinates {
					(0.1,13.8)(0.2,13)(0.3,13.6)(0.4,12.7)(0.5,12.6)(0.6,13)(0.7,15.6)(0.8,12.6)(0.9,11.7)
				};
				\addlegendentry{Morphence}

				\addplot[color=red,mark=+]
				coordinates {
					(0.1,2.6)(0.2,2.5)(0.3,2.8)(0.4,2.1)(0.5,2)(0.6,2.5)(0.7,3.9)(0.8,2.7)(0.9,1.8)
				};
				\addlegendentry{MTDeep}
					
				\addplot[color=green,mark=x]
				coordinates {
					(0.1,4.9)(0.2,4.7)(0.3,8)(0.4,7.5)(0.5,5.9)(0.6,7.6)(0.7,9.6)(0.8,9)(0.9,7.2)
				};
				\addlegendentry{StratDef}
					
				\addplot[color=pink,mark=oplus]
				coordinates {
					(0.1,3.9)(0.2,3.8)(0.3,4.2)(0.4,3.7)(0.5,2.9)(0.6,3.8)(0.7,7.2)(0.8,3.2)(0.9,4.5)
				};
				\addlegendentry{Voting (Majority)}
					
				\addplot[color=brown,mark=triangle]
				coordinates {
					(0.1,25.8)(0.2,25.3)(0.3,25.8)(0.4,25.9)(0.5,25.2)(0.6,25.2)(0.7,26.6)(0.8,27.5)(0.9,24.3)
				};
				\addlegendentry{Voting (Veto)}
					  \addplot[color=yellow]
                coordinates {
                (0.1,16.3)(0.2,15.6)(0.3,15.9)(0.4,15.3)(0.5,16.9)(0.6,14.6)(0.7,18.9)(0.8,18.5)(0.9,15.3)
                };
            \addlegendentry{NN-AT}

			\end{axis}
		\end{tikzpicture}
		\vspace{-6.5mm}
		\caption{DREBIN}
	\end{subfigure}
	\hspace{0.1cm} 
	\begin{subfigure}[b]{0.15\textwidth}
		\begin{tikzpicture}[scale=0.45]
			\begin{axis}[
					xlabel={\(q\)},
					ylabel={FPR (\%)}, 
					xmin=0.1, xmax=0.9,
					ymin=0, ymax=30,
					xtick={0.2,0.4,0.6,0.8},
					ytick={0,10,20,30,40},
					ymajorgrids=false,
					legend pos=south west,
					legend style={nodes={scale=0.5, transform shape}},
					cycle list name=color list,
					legend to name=grayboxfullfprlegend,
					legend columns=8,
            height=3.6cm,
            width=6.6cm,
            ylabel style=
            {
                yshift=-2mm, 
            }
				]
					
				\addplot[color=purple,mark=diamond]
				coordinates {
					(0.1,15.1)(0.2,14.2)(0.3,14.8)(0.4,14.4)(0.5,15)(0.6,15.3)(0.7,16.6)(0.8,13)(0.9,12)
				};
				\addlegendentry{DeepMTD}
							
				\addplot[color=cyan,mark=pentagon]
				coordinates {
					(0.1,10.7)(0.2,9.4)(0.3,10.6)(0.4,10.2)(0.5,10.5)(0.6,11.4)(0.7,11.3)(0.8,9.3)(0.9,9.5)
				};
				\addlegendentry{Morphence}

				\addplot[color=red,mark=+]
				coordinates {
					(0.1,3.3)(0.2,2.9)(0.3,3.1)(0.4,3)(0.5,2.7)(0.6,2.4)(0.7,4.7)(0.8,2.1)(0.9,2.1)
				};
				\addlegendentry{MTDeep}
					
				\addplot[color=green,mark=x]
				coordinates {
					(0.1,8.9)(0.2,8.6)(0.3,9)(0.4,9.3)(0.5,9.1)(0.6,7.6)(0.7,7.9)(0.8,9.9)(0.9,6)
				};
				\addlegendentry{StratDef}
					
				\addplot[color=pink,mark=oplus]
				coordinates {
					(0.1,11.6)(0.2,10.4)(0.3,11.4)(0.4,10.7)(0.5,11.6)(0.6,11.8)(0.7,12.5)(0.8,9.1)(0.9,10.3)
				};
				\addlegendentry{Voting (Majority)}
					
				\addplot[color=brown,mark=triangle]
				coordinates {
					(0.1,22)(0.2,20.7)(0.3,22.5)(0.4,21)(0.5,21.4)(0.6,21.5)(0.7,20.9)(0.8,22.5)(0.9,21.5)
				};
				\addlegendentry{Voting (Veto)}
					  \addplot[color=yellow]
                coordinates {
                (0.1,8.8)(0.2,7.9)(0.3,8.8)(0.4,8.7)(0.5,8.8)(0.6,9.5)(0.7,9.6)(0.8,7)(0.9,7.9)
                };
            \addlegendentry{NN-AT}

			\end{axis}
		\end{tikzpicture}
		\vspace{-6.5mm}
		\caption{SLEIPNIR}
	\end{subfigure}
	\hspace{0.1cm} 
	\begin{subfigure}[b]{0.15\textwidth}
		\begin{tikzpicture}[scale=0.45]
			\begin{axis}[
					xlabel={\(q\)},
					ylabel={FPR (\%)}, 
					xmin=0.1, xmax=0.9,
					ymin=0, ymax=30,
					xtick={0.1,0.2,0.3,0.4,0.5,0.6,0.7,0.8,0.9},
					ytick={0,10,20,30},
					ymajorgrids=false,
					legend pos=south west,
					legend style={nodes={scale=0.5, transform shape}},
					cycle list name=color list,
					legend to name=grayboxfullfprlegend,
					legend columns=8,
                    height=3.6cm,
                    width=6.6cm,
            ylabel style=
            {
                yshift=-2mm, 
            }
				]
				
				\addplot[color=purple,mark=diamond]
				coordinates {
					(0.1,3.7)(0.2,3.8)(0.3,3.5)(0.4,4.3)(0.5,3.9)(0.6,3.2)(0.7,3.4)(0.8,3.8)(0.9,3.5)

				};
				\addlegendentry{DeepMTD}
						
				\addplot[color=cyan,mark=pentagon]
				coordinates {
					(0.1,6.6)(0.2,6.9)(0.3,6.7)(0.4,7.6)(0.5,7.7)(0.6,6.9)(0.7,6.4)(0.8,7)(0.9,5.7)

				};
				\addlegendentry{Morphence}

				\addplot[color=red,mark=+]
				coordinates {
					(0.1,4.1)(0.2,4.4)(0.3,4.1)(0.4,4.5)(0.5,4.8)(0.6,4)(0.7,3.9)(0.8,3.6)(0.9,4.1)

				};
				\addlegendentry{MTDeep}
				
				\addplot[color=green,mark=x]
				coordinates {
					(0.1,4.3)(0.2,4.4)(0.3,4.3)(0.4,5.3)(0.5,4.3)(0.6,4.6)(0.7,4)(0.8,3.8)(0.9,4.4)

				};
				\addlegendentry{StratDef}
				
				\addplot[color=pink,mark=oplus]
				coordinates {
					(0.1,5.5)(0.2,5.8)(0.3,5.3)(0.4,6.2)(0.5,5.6)(0.6,5.7)(0.7,4.9)(0.8,4.4)(0.9,4.7)

				};
				\addlegendentry{Voting (Majority)}
				
				\addplot[color=brown,mark=triangle]
				coordinates {
					(0.1,14.5)(0.2,14.8)(0.3,14)(0.4,15.2)(0.5,15)(0.6,14)(0.7,13.8)(0.8,13.5)(0.9,15.2)

				};
				\addlegendentry{Voting (Veto)}
				
        \addplot[color=yellow]
            coordinates {
				(0.1,5.1)(0.2,5.5)(0.3,4.8)(0.4,5.5)(0.5,5.5)(0.6,5.2)(0.7,4.6)(0.8,5.1)(0.9,5.7)

			};
        \addlegendentry{NN-AT}

			\end{axis}
		\end{tikzpicture}
		\vspace{-6.5mm}
		\caption{AndroZoo}
	\end{subfigure}
	\ref{grayboxfullfprlegend}
\vspace{-2.5mm}
	\caption{Accuracy and FPR vs. $q$.}
	\label{figure:grayboxvsq}
\vspace{-4mm}
\end{figure}

Figure \ref{figure:grayboxvsq} shows the accuracy and false positive rate (FPR) of each defense when queried over 1000 times with different proportions of adversarial examples (see Appendix~\ref{appendix:extendedevaluationbigger} for F1 and AUC). While veto voting and Morphence offer good accuracy, particularly for SLEIPNIR, the greater FPR associated with these defenses can also be seen across all values of \(q\). The FPR of veto voting is much higher than most other defenses. If such defenses were deployed in a real-world setting, users would often receive incorrect predictions, which is unsuitable for the malware detection domain, as explained before. Conversely, MTDs like StratDef offer a more balanced performance in this regard, while also being more well-rounded against the adversarial threat.

\vspace{-3mm}

\section{Fingerprinting \& Reconnaissance}
\label{sec:otherfindings}
\vspace{-1mm}
An enhanced understanding of defenses can be gained through fingerprinting and reconnaissance. Recall that reconnaissance can yield useful information for \emph{future} attacks. If an attacker has a greater awareness of how an MTD operates, they could improve future attacks. We introduce two methods to demonstrate how this could be achieved, showing promising results and widening the scope for more elaborated attacks in future work.

\noindent{\textbf{Determining Predictive Nature of Defenses.}} As discussed in Section~\ref{sec:background}, an MTD is dynamic or hybrid in nature (whereas other defenses may be static). For defenses exhibiting static behavior (as well as hybrid defenses, to some degree), predictions will remain the same. This is ideal for an attacker, as it guarantees that an adversarial example can be reused, thereby ensuring its future success. Conversely, a defense behaving dynamically will exhibit less predictable behavior, leading to reduced repeat evasion (as seen before). 

Determining this predictive nature can be accomplished by querying each model with different input samples repeatedly (\(n\) times). Subsequently, a consensus mechanism determines whether predictions for the same input sample have changed across the queries. Fluctuations in predictions are an indication of a dynamic defense, while static defenses will always produce the same prediction for a single input sample. For instance, upon applying this technique with \(n=100\), several defenses such as DeepMTD, MTDeep, and voting (majority and veto) mostly produce predictions statically. This is understandable, as voting is static by nature. Meanwhile, the lack of variance in predictions indicates that DeepMTD may not be effectively regenerating student models, while MTDeep may be employing a pure strategy, which occurs when it computes the optimal strategy to be that only a single model from its ensemble serves the predictions. This is also why the evasion rate against these defenses is perhaps considerably higher in our previous experiments. Meanwhile, Morphence also delivers predictions statically until its query budget is exceeded. However, as we show in the following section, if the attacker discovers that the oracle produces different predictions after \(n\) queries, they can adapt their attack to use \(< n\) queries to ensure the oracle's static behavior. StratDef exhibits dynamic behavior; for example,  predictions for the same input samples vary by an average of 5.7\% for DREBIN and 2.7\% for SLEIPNIR on the test set described in Section~\ref{sec:expsetup}.

\noindent{\textbf{Determining Movement of Hybrid Defenses.}} Hybrid defenses provide static predictions until a condition is met, at which point predictions for may differ from before. %
For example, Morphence uses a query budget for this. However, this may be discoverable using the method described previously to determine the oracle's predictive nature. This is an example of stealing hyperparameters in ML \cite{wang2018stealing}. 

\begin{figure}[!htbp]
\vspace{-2.5mm}
\centering
    \begin{tikzpicture}[scale=0.65]
        \begin{axis}[
            height=2.5cm,
            width=12cm,
            tick style={draw=none},
            axis lines = left,
            xlabel={Number of queries (\(n\))},
            ylabel={Predicted class}, 
            xmin=100, xmax=1000,
            ymin=0, ymax=1,
            xtick={100,200,300,400,500,600,700,800,900,1000},
            xticklabels={1K,2K,3K,4K,5K,6K,7K,8K,9K,10K},
            ytick={0.0,1.0},
            ymajorgrids=false,
            legend pos=south west,
            legend style={nodes={scale=0.7, transform shape}}
            legend = []
        ]
        
        \addplot[
            line width=2pt,
            color=blue,
            ]
            coordinates {
            (100, 1)(200, 1)(299, 1)(300, 0)(399, 0)(400, 0)(499, 0)(500, 1)(600, 1)(699, 1)(700, 0)(799, 0)(800, 1)(900, 1)(1000, 1)
            };

        \end{axis}
    \end{tikzpicture}
\vspace{-2.5mm}
\caption{Prediction for the same input sample changes between 3-5K and 7-8K queries, implying that \(Q_{max} \leq 1K\).}
\label{figure:qmaxmorp}
\vspace{-3mm}
\end{figure}
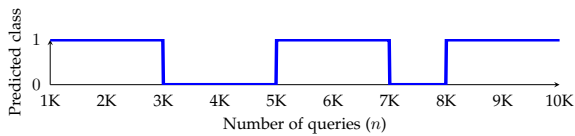

Across the queries, the specific point where the oracle modifies its predictions can indicate whether a query budget exists and allow for an estimation of its value. For example, Morphence's queuing system means that a longer waiting time implies greater system utilization. Therefore, when system utilization is low, we determine the predictive nature of the defense but with a single input sample and a much larger value of $n$ (e.g., 10,000). The intended result is that predictions vary across \(n\) to give an indication of the value of any query budget. If predictions do not change, it could be due to the input sample or because \(n\) is smaller than the query budget. In an experiment to discover the query budget for Morphence (\(Q_{max}\)), Figure~\ref{figure:qmaxmorp} shows that the prediction for the same input sample varies between 3-5K and 7-8K queries. This implies that the oracle (which is an example of a Morphence instance) changes after \(\approx1000\) queries (or \(Q_{max} \leq 1000\)), which can be confirmed by using another input sample. With this knowledge, we could develop \(\Delta\) for a black-box transferability attack without exceeding the query budget to ensure the static behavior of the oracle or perform a query attack with an \(n_{max}\) lower than the estimated \(Q_{max}\). Recall that this is an \emph{evaluation instance} of Morphence. In practice, one may find that a different $n$ is required to conduct this reconnaissance exercise.

\vspace{-2mm}
\section{Discussion \& Recommendations}
\label{sec:discussion}
\vspace{-1.5mm}
In this section, we discuss the overall findings from our evaluation. Based on these and our experimental results, we make key recommendations for developing effective MTDs against adversarial attacks in ML-based malware detection. 

In some cases, the evaluated MTDs are no better than the other ensemble or single-model defenses. %
Among the evaluated MTDs, DeepMTD and MTDeep perform poorly across the board, regardless of the type of attack strategy. Our results show that both seem to exhibit a consistent lack of variance in predictions, leading to almost static behavior in practice. As already observed in previous work \cite{qian2020ei}, DeepMTD's mechanism for regenerating its student models only works when the defense is idle for some time. Regarding MTDeep, we observe that in this domain, its optimizer only generates pure strategies; that is, it only uses a single model in practice to make the predictions. Therefore, these defenses are unable to take advantage of the main characteristic that an MTD should offer: dynamic behavior to increase the attacker's uncertainty.  

Meanwhile, Morphence and StratDef offer some degree of robustness in different attack scenarios. In particular, our results show that StratDef performs well for transferability attacks under both black-box and gray-box attack scenarios. Meanwhile, Morphence offers slightly better performance than other MTDs against query attacks, especially for lower values of \mbox{\(n_{max}\)} where the attacker faces more restrictions on the number of queries for their attack. The improved performance of these defenses can be attributed to more dynamic behavior (i.e., ``how to move''). We observe in practice that both Morphence and StratDef vary their behavior dynamically and use higher model diversity (i.e., ``what to move'') with diverse constituent models beyond DNNs, which makes attacks against these MTDs less successful. Despite this, their performance is far from perfect, and in some threat models and attacks, particularly the gray-box query attack, both Morphence and StratDef can be well-evaded. Also, and particularly in the case of Morphence, we have shown that it can be fingerprinted, leaving key hyperparameters that govern its behavior exposed to attackers. %

Yet, we believe that there is scope for MTDs to be a promising direction for defending against adversarial ML in the malware detection domain. If designed well, MTDs have the potential to introduce a layer of complexity and uncertainty for the attacker to thwart attacks. Hence, based on our results, we make key recommendations for developing effective MTDs to motivate further work in this area:

\noindent{\textbf{Don't use MTDs that behave statically.}} Static defenses have been found to be less effective in dealing with the adversarial ML threat when compared with more dynamic MTDs. As we have shown, defenses with inadequate movement mechanisms are no better than single-model defenses such as adversarial training. In some cases, using a dynamic or at least hybrid MTD such as Morphence or StratDef seems to offer better performance across threat models and attack strategies in terms of evasion rates and other metrics. In addition, high repeat evasion rates can be prevented as adversarial examples work \emph{inconsistently} against the oracle.

\noindent{\textbf{Increase the diversity of ``what to move''.}} %
Using a dynamic or hybrid MTD may still be ineffective if its constituent models can  be evaded through transferability. To minimize transferability between the constituent models, the constituent models need to be as diverse as possible. In fact, we have seen that MTDs that use different families (and not just DNNs) seem to perform better. %
This means that other model families (e.g., random forests, support vector machines, etc.) should also be included in the constituent models. %

\noindent{\textbf{Be independent on ``how to move''.}} The movement mechanism for MTDs should not be influenced solely by user behavior. Some defenses (e.g., Morphence) regenerate models to ``move'' their configuration, which is impacted exclusively by user activity (e.g., number of user queries), rather than by an independent mechanism (e.g., time or randomness). A user-influenced mechanism makes the oracle an easy target. Therefore, MTDs must ``move'' independently and cycle through configurations to prevent attacks.

\noindent{\textbf{Consider leakage of hyperparameters.}} Hyperparameters are carefully designed and chosen for a specific context and may compromise a defense if leaked. %
For example, in Section~\ref{sec:otherfindings}, we demonstrate how a query budget can be discovered. This could make it easy for the attacker to consider the available query budget in a future attack, allowing them to take advantage of the defense before it changes its behavior. %
Therefore, it is critical that the design of an MTD accommodates the potential leakage of hyperparameters. The defender should operate under the assumption that hyperparameters may be leaked. For example, in the scenario described earlier, hyperparameter leakage could be mitigated by cycling through models at prediction-time (randomly or strategically) so that it cannot be reliably ascertained %
how often predictions are changing.

\noindent{\textbf{Greater system awareness is vital.}} While an MTD has several advantages, it cannot identify when an attack is taking place. For instance, in our query attacks, this means that after several queries, most MTDs can be evaded as single-model defenses (like adversarially-trained models). Therefore, greater system awareness is vital, especially for protecting against query attacks given that particularly dynamic MTDs could vary their strategy at run-time. Prior work \cite{chen2020stateful, li2020blacklight} has shown that stateful detection methods could help detect attacks against single-model defenses in some cases. Therefore, such stateful detection methods as well as cyber-threat intelligence could be combined with an MTD to offer increased awareness and robustness. This would lead to an adaptive MTD system for detecting and reacting to potential \emph{attacks in progress} (e.g., by changing strategies or deciding to make a different move).

\vspace{-1.5mm}
\section{Related Work}
\label{sec:relatedwork}
\vspace{-1.5mm}
Szegedy et al. \cite{szegedy2013intriguing} introduced adversarial examples, demonstrating that they transfer across different models. Other work explored transferability attacks within other domains (e.g., image recognition) \cite{papernot2017practical, demontis2019adversarial, papernot2016transferability}. Papernot et al. \cite{papernot2017practical}  explored methods for attacking remotely-hosted single-model classifiers based on a black-box threat model introduced in \cite{laskov2014practical}, which was further explored in the malware detection domain \cite{rosenberg2018generic, brendel2017decision, demetrio2021functionality, ilyas2018black}. Liu et al. \cite{liu2016delving} then demonstrated that adversarial examples generated using ensembles of DNNs can lead to the evasion of a black-box image classifier. Our transferability attack improves on these approaches, as we use an ensemble of diverse substitute models and validate the transferability of adversarial examples across the substitute models before testing on the oracle. Prior work also developed query attacks (primarily for images) \cite{chen2020stateful, brendel2017decision, chen2020hopskipjumpattack, rosenberg2020query, ilyas2018black, li2020qeba, Bhagoji_2018_ECCV} to generate adversarial examples using techniques such as gradient and decision boundary estimation. These attacks cannot deal with discrete features and functionality preservation \cite{rosenberg2020query}. Compared to prior work \cite{rosenberg2020query}, our query attack does not assume access to prediction scores or use sliding windows, so we consider a threat model with even less information. Additionally, our gray-box query attack chooses which features to perturb in a heuristic manner rather than randomly.

Other work has explored model stealing, which is a different adversarial aim from ours \cite{tramer2016stealing, wang2018stealing, biggio2013evasion, biggio2018wild}. For instance, \cite{tramer2016stealing} shows how to obtain parameters from remote classifiers with partial knowledge of models. Several studies on adversarial ML have been conducted in other domains, such as network systems and website fingerprinting \cite{alatwi2021adversarial, 10.1145/3474369.3486875,10.1109/ASE.2019.00080,272174,schonherr2018adversarial}. Nonetheless, we show that the hyperparameters of MTDs and details about their behavior can be stolen in a black-box setting. 

A challenge within ML-based malware detection is model sustainability. As malware evolves, it becomes difficult to generalize models to detect unseen behavior, making classifiers unsustainable \mbox{\cite{203684, 10.1145/3183440.3195004, 8802672}}. Due to \emph{concept drift}, models may make poor decisions when facing the latest threats (but not always \mbox{\cite{10.1145/3503463}}). Prior work has suggested periodically retraining models \mbox{\cite{10.1007/978-3-642-04342-0_2, robotics8030050}}, though this may reduce their learning ability. Thus, the constant evolution of malware makes it a moving target in its own right. To deal with this, MTDs could use a detection system for concept drift \cite{203684} and then retire vulnerable constituent models.
 
Our work is part of a series of evaluations of defenses against adversarial ML attacks. Prior work has demonstrated that single-model defenses are ineffective at dealing with adversarial examples \cite{pods,shahzad2013comparative,athalye2018obfuscated,papernot2016limitations,papernot2018sok,goodfellow2014explaining}. He et al. \cite{he2017adversarial} showed that weak ensembles are insufficient against adversarial ML using image datasets. Work conducted in our domain has provided a grim view of the capabilities of available defenses \cite{pods,shahzad2013comparative}. %
We are the first to conduct a comparative evaluation of MTDs. Our work proposes novel attack strategies for increasing evasion against different models, especially MTDs. %
Moreover, past research on MTDs \mbox{\cite{cho2020toward, dasgupta2019survey, rashid2022stratdef, sengupta2018mtdeep, qian2020ei, song2019moving, amich2021morphence, additional1, additional2, 9833895, 9652915}} has not evaluated MTDs under different threat models, nor compared MTDs with each other, nor offered practical recommendations for developing effective MTD systems based on an evaluation of recent MTDs. Although \mbox{\cite{9652915}} makes some suggestions, they are different from ours, as we do not recommend randomization or using only DNNs.

\vspace{-1.5mm}

\section{Conclusion}
\label{sec:conclusion}
\vspace{-1.5mm}
We studied the effectiveness of several MTDs against evasion attacks in ML-based malware detection. For this, we used existing transferability and query attack strategies as well as novel strategies specifically tailored to MTDs. %
Under different scenarios, we demonstrated that our attack strategies achieve high evasion with minimal queries to the target model. Moreover, we demonstrated that it may be possible to understand how a target model operates through fingerprinting and reconnaissance methods.

Based on our findings, we provided recommendations for developing effective MTDs. MTDs that behave statically should be avoided, while MTDs that utilize diverse constituent models with effective movement should be favored. That is, an MTD should use models from different families (beyond just DNNs or just different DNN architectures) to maximize diversity for limiting the effectiveness of transferability attacks. Furthermore, these models should be used dynamically and moved strategically to maximize uncertainty and complexity for the attacker. In addition, we suggested that MTDs be coupled with greater system awareness to detect and react to \emph{attacks in progress}. This could pave the way for a promising way forward in dealing with attacks. This may be accomplished by developing a response that is both automated and stateful \cite{chen2020stateful}, and/or based on cyber-threat intelligence~\cite{shu2018threat,zhu2018chainsmith}.

\ifCLASSOPTIONcaptionsoff
  \newpage
\fi

\bibliographystyle{IEEEtran}
\bibliography{bib.bib}

\begin{appendices}
\vspace{-2mm}
\section{Evaluated Defenses}
\vspace{-1.5mm}
\label{appendix:evaldef}
Each defense has been evaluated using configuration and parameters as close as possible to its original paper. 
\vspace{-1.5mm}
\begin{table}[H]
\scalebox{0.8}{
\centering
\begin{tabular}{ll} 
\hline
Defense & Configuration/parameters/setup \\
\hline 
DeepMTD \cite{song2019moving} & \makecell[l]{\(w=0.3\), \(n=20\), \(T=0.6\)} \\
\hdashline
Morphence \cite{amich2021morphence} & \makecell[l]{\(n=4\), \(p=3\), \(Q_{max}=1000\)} \\
\hdashline
MTDeep \cite{sengupta2018mtdeep} & \makecell[l]{5 DNNs as constituent models. Assumed \(\alpha=1\)} \\
\hdashline
StratDef \cite{rashid2022stratdef} & \makecell[l]{Variety-GT using same models as paper. \\Assumed strong attacker and \(\alpha=1\)} \\
\hdashline
Majority \& Veto voting & \makecell[l]{Using same models as StratDef-Variety-GT} \\
\hdashline
NN-AT & \makecell[l]{\makecell[l]{4 fully-connected layers (128 (Relu),\\ 64 (Relu), 32 (Relu), 2 (Softmax)).\\ Adversarially-trained with up 25\% size of training data}} \\
\hline
\end{tabular}
}
\end{table}

\vspace{-2mm}

\section{Architectures of Substitute Models (\& Vanilla Models)}
\label{appendix:arcsubmodels}
\vspace{-1.5mm}
These model architectures are used for the substitute models in our transferability attack strategies. Additionally, we construct a set of vanilla models with these architectures that are used to generate a set of adversarial examples for Ensemble Adversarial Training (see \mbox{Section~\ref{sec:expsetup}}).
\begin{table}[H]
\scalebox{0.9}{
\centering
\begin{tabular}{ll} 
\hline
Model & Parameters \\
\hline 
Decision Tree & \makecell[l]{max\_depth=5, min\_samples\_leaf=1} \\
\hdashline
Neural Network & \makecell[l]{3 fully-connected layers (100 (Relu),\\ 50 (Relu), 2 (Softmax))} \\
\hdashline
Random Forest & \makecell[l]{max\_depth=100} \\
\hdashline
Support Vector Machine & \makecell[l]{LinearSVC with probability enabled} \\
\hline
\end{tabular}
}
\end{table}

\vspace{-2mm}

\section{Permitted Perturbations for DREBIN and AndroZoo}
\label{appendix:allowedperturbationsdrebin}
\vspace{-1.5mm}
DREBIN \mbox{\cite{arp2014drebin}} and AndroZoo \mbox{\cite{Allix:2016:ACM:2901739.2903508}} are Android datasets, both of which can be divided into eight feature families comprised of extracted static features such as permissions, API calls, hardware requests, and URL requests.  According to industry literature and prior work (e.g., \mbox{\cite{li2021framework, 8171381, li2020enhancing, al2018adversarial, pierazzi2020problemspace, labaca2021universal, rashid2022stratdef}}), features may be added or removed during attacks to traverse the decision boundary, based on the feature family. 

However, malicious functionality must be preserved as a core constraint in this domain. As we operate in the feature-space, we offer a lower bound of functionality preservation. For example, attacks cannot remove features from the manifest file nor intent filter, and component names must be consistently named. Therefore, the table below enumerates the perturbations for each feature family that are allowed. For example, if a feature belonging to the S2 family is removed by an attack, then its original value is restored as it is not permitted to be removed (see \mbox{Section~\ref{sec:expsetup}}).

\begin{table}[H]
\centering
\scalebox{0.9}{
\begin{tabular}{llcc}
    \hline
     & Feature families & \multicolumn{1}{l}{Addition} & \multicolumn{1}{l}{Removal} \\ \hline
    \multirow{4}{*}{manifest} & S1 Hardware & \cmark & \xmark \\
     & S2 Requested permissions & \cmark & \xmark \\
     & S3 Application components & \cmark & \cmark \\
     & S4 Intents & \cmark & \xmark \\ \hline
    \multirow{4}{*}{dexcode} & S5 Restricted API Calls & \cmark & \cmark \\
     & S6 Used permission & \xmark & \xmark \\
     & S7 Suspicious API calls & \cmark & \cmark \\
     & S8 Network addresses & \cmark & \cmark \\ \hline
    \end{tabular}
}
\caption{Permitted perturbations for Android datasets. These are determined by consulting industry documentation and prior work \cite{li2021framework, 8171381, li2020enhancing, al2018adversarial, pierazzi2020problemspace, labaca2021universal, rashid2022stratdef}.}
\label{table:permittedperturbations}
\end{table}

\vspace{-2mm}

\section{Effectiveness of Attacks Against Models With Fewer Features}
\label{appendix:lessfeaturesexp}
We evaluate the performance of our gray-box query attack strategy against two vanilla neural network models to demonstrate the effectiveness of attacks when the feature-space is \emph{drastically} reduced. That is, for each dataset, we train two vanilla neural network models: one with the full set of features (described in Section~\ref{sec:expsetup}); and another with only 500 features. For the neural network with reduced features, the features are selected by using the \verb|SelectKBest| function of the scikit-learn library (with the chi2 scoring function that computes the chi-squared stats between each non-negative feature and class). The table below shows that the attack performs just as well when the feature-space is significantly decreased for DREBIN and SLEIPNIR.

\begin{table}[!htbp]
\centering
\begin{tabular}{lll}
\hline
 & \textbf{500 features} & \textbf{Full features} \\
\hline
DREBIN & 100\% & 98.6\% \\
\hdashline
SLEIPNIR & 100\% & 99.5\% \\
\hline
\label{table:reducedfeaturesexpvanillann}
\end{tabular}
\caption{Evasion rate achieved by gray-box query attack against vanilla neural networks with different numbers of features.}
\end{table}

\vspace{-2mm}

\section{Extended Results}
\label{appendix:extendedresults}
\vspace{-1.5mm}
The extended results are located in the following anonymous repository: \url{https://osf.io/nym5a/?view_only=4ba9b399086c4f7cadc65a6a4e8da83e}

\section{F1 and AUC (Section~\ref{sec:widerpicture})}
\label{appendix:extendedevaluationbigger}
\vspace{-1.5mm}
\begin{figure}[H]
\hspace{-0.25cm} 
	\centering
	\begin{subfigure}[b]{0.15\textwidth}
		\begin{tikzpicture}[scale=0.45]
			\begin{axis}[
					xlabel={\(q\)},
					ylabel={F1 (\%)}, 
					xmin=0.1, xmax=0.9,
					ymin=0, ymax=100,
					xtick={0.2,0.4,0.6,0.8},
					ytick={0,20,40,60,80,100},
					ymajorgrids=false,
					legend pos=south west,
					legend style={nodes={scale=0.5, transform shape}},
					cycle list name=color list,
					legend to name=grayboxfullauclegend,
					legend columns=8,
                    height=3.6cm,
                    width=6.6cm,
            ylabel style=
            {
                yshift=-2mm, 
            }
				]

				\addplot[color=purple,mark=diamond]
				coordinates {
					(0.1,89.4)(0.2,82.8)(0.3,74.9)(0.4,68.3)(0.5,63.1)(0.6,55.8)(0.7,51.6)(0.8,47.2)(0.9,42.5)
				};
				\addlegendentry{DeepMTD}
							
				\addplot[color=cyan,mark=pentagon]
				coordinates {
					(0.1,87.8)(0.2,82.6)(0.3,77.9)(0.4,72.6)(0.5,69.1)(0.6,63.9)(0.7,61.4)(0.8,57.8)(0.9,53.2)
				};
				\addlegendentry{Morphence}

				\addplot[color=red,mark=+]
				coordinates {
					(0.1,89.9)(0.2,83.3)(0.3,75.9)(0.4,68.8)(0.5,63.8)(0.6,56.5)(0.7,53.1)(0.8,47.8)(0.9,43.5)
				};
				\addlegendentry{MTDeep}
					
				\addplot[color=green,mark=x]
				coordinates {
					(0.1,92)(0.2,86.1)(0.3,81)(0.4,79.3)(0.5,78.3)(0.6,74)(0.7,72.8)(0.8,72.6)(0.9,70.5)
				};
				\addlegendentry{StratDef}
					
				\addplot[color=pink,mark=oplus]
				coordinates {
					(0.1,91.1)(0.2,84.9)(0.3,79.5)(0.4,73.7)(0.5,70)(0.6,64)(0.7,61)(0.8,58)(0.9,52.7)
				};
				\addlegendentry{Voting (Majority)}
					
				\addplot[color=brown,mark=triangle]
				coordinates {
					(0.1,86.9)(0.2,85.2)(0.3,82.4)(0.4,80.7)(0.5,80.5)(0.6,77.4)(0.7,77.1)(0.8,76.5)(0.9,74.7)
				};
				\addlegendentry{Voting (Veto)}
            \addplot[color=yellow]
                coordinates {
                (0.1,87)(0.2,82.1)(0.3,77.3)(0.4,72)(0.5,68.3)(0.6,63.9)(0.7,61.5)(0.8,57.6)(0.9,53.5)

                };
            \addlegendentry{NN-AT}

			\end{axis}
		\end{tikzpicture}
		\vspace{-6.5mm}
		\caption{DREBIN}
	\end{subfigure}
	\hspace{0.1cm} 
	\begin{subfigure}[b]{0.15\textwidth}
		\begin{tikzpicture}[scale=0.45]
			\begin{axis}[
					xlabel={\(q\)},
					ylabel={F1 (\%)}, 
					xmin=0.1, xmax=0.9,
					ymin=0, ymax=100,
					xtick={0.2,0.4,0.6,0.8},
					ytick={0,20,40,60,80,100},
					ymajorgrids=false,
					legend pos=south west,
					legend style={nodes={scale=0.5, transform shape}},
					cycle list name=color list,
					legend to name=grayboxfullauclegend,
					legend columns=8,
                    height=3.6cm,
                    width=6.6cm,
            ylabel style=
            {
                yshift=-2mm, 
            }
				]
					
				\addplot[color=purple,mark=diamond]
				coordinates {
					(0.1,83.5)(0.2,77.6)(0.3,70.4)(0.4,64.1)(0.5,56.8)(0.6,50.9)(0.7,44.6)(0.8,38.8)(0.9,32.4)
				};
				\addlegendentry{DeepMTD}
							
				\addplot[color=cyan,mark=pentagon]
				coordinates {
					(0.1,92.9)(0.2,94.6)(0.3,95.6)(0.4,96.6)(0.5,97.2)(0.6,97.8)(0.7,98.4)(0.8,99.1)(0.9,99.5)
				};
				\addlegendentry{Morphence}

				\addplot[color=red,mark=+]
				coordinates {
					(0.1,83)(0.2,76.4)(0.3,68.6)(0.4,62.3)(0.5,54.6)(0.6,49.2)(0.7,43.1)(0.8,37.5)(0.9,31.7)
				};
				\addlegendentry{MTDeep}
					
				\addplot[color=green,mark=x]
				coordinates {
					(0.1,93.9)(0.2,95.2)(0.3,95.9)(0.4,96.5)(0.5,97.4)(0.6,98.1)(0.7,98.8)(0.8,99.1)(0.9,99.7)
				};
				\addlegendentry{StratDef}
					
				\addplot[color=pink,mark=oplus]
				coordinates {
					(0.1,93.1)(0.2,94.7)(0.3,95.7)(0.4,96.7)(0.5,97.2)(0.6,97.9)(0.7,98.5)(0.8,99.2)(0.9,99.6)
				};
				\addlegendentry{Voting (Majority)}
					
				\addplot[color=brown,mark=triangle]
				coordinates {
					(0.1,89.9)(0.2,92)(0.3,93.3)(0.4,94.8)(0.5,95.9)(0.6,96.9)(0.7,97.8)(0.8,98.5)(0.9,99.4)
				};
				\addlegendentry{Voting (Veto)}
				  \addplot[color=yellow]
                coordinates {
                (0.1,93.3)(0.2,94.9)(0.3,95.8)(0.4,96.8)(0.5,97.4)(0.6,98)(0.7,98.6)(0.8,99.2)(0.9,99.6)
                };
            \addlegendentry{NN-AT}

			\end{axis}
		\end{tikzpicture}
		\vspace{-6.5mm}
		\caption{SLEIPNIR}
	\end{subfigure}
	\hspace{0.1cm} 
	\begin{subfigure}[b]{0.15\textwidth}
		\begin{tikzpicture}[scale=0.45]
			\begin{axis}[
					xlabel={\(q\)},
					ylabel={F1 (\%)}, 
					xmin=0.1, xmax=0.9,
					ymin=0, ymax=100,
					xtick={0.2,0.4,0.6,0.8},
					ytick={0,20,40,60,80,100},
					ymajorgrids=false,
					legend pos=south west,
					legend style={nodes={scale=0.5, transform shape}},
					cycle list name=color list,
					legend to name=androzooandrozoograyboxfullauclegend,
					legend columns=8,
                    height=3.6cm,
                    width=6.6cm,
            ylabel style=
            {
                yshift=-2mm, 
            }
				]
					
				\addplot[color=purple,mark=diamond]
				coordinates {
					(0.1,89.2)(0.2,82.1)(0.3,75.2)(0.4,68.8)(0.5,62.8)(0.6,57)(0.7,51.2)(0.8,45)(0.9,40.2)
				};
				\addlegendentry{DeepMTD}
							
				\addplot[color=cyan,mark=pentagon]
				coordinates {
					(0.1,89.5)(0.2,82.9)(0.3,76.5)(0.4,70.7)(0.5,65.9)(0.6,61.2)(0.7,55.1)(0.8,49.4)(0.9,45.6)
				};
				\addlegendentry{Morphence}

				\addplot[color=red,mark=+]
				coordinates {
					(0.1,89.4)(0.2,82.4)(0.3,75.6)(0.4,69.5)(0.5,64.1)(0.6,59.1)(0.7,52.9)(0.8,47.2)(0.9,42.8)
				};
				\addlegendentry{MTDeep}
					
				\addplot[color=green,mark=x]
				coordinates {
					(0.1,97.3)(0.2,97.9)(0.3,98.3)(0.4,98.4)(0.5,98.9)(0.6,99.2)(0.7,99.5)(0.8,99.7)(0.9,99.9)
				};
				\addlegendentry{StratDef}
					
				\addplot[color=pink,mark=oplus]
				coordinates {
					(0.1,96.9)(0.2,97.5)(0.3,98)(0.4,98.3)(0.5,98.7)(0.6,99.1)(0.7,99.4)(0.8,99.7)(0.9,99.8)
				};
				\addlegendentry{Voting (Majority)}
					
				\addplot[color=brown,mark=triangle]
				coordinates {
					(0.1,93.9)(0.2,95)(0.3,96)(0.4,96.6)(0.5,97.3)(0.6,98.1)(0.7,98.7)(0.8,99.2)(0.9,99.6)
				};
				\addlegendentry{Voting (Veto)}
				  \addplot[color=yellow]
                coordinates {
					(0.1,97.1)(0.2,97.6)(0.3,98.1)(0.4,98.4)(0.5,98.7)(0.6,99.1)(0.7,99.4)(0.8,99.6)(0.9,99.8)
                };
            \addlegendentry{NN-AT}

			\end{axis}
		\end{tikzpicture}
		\vspace{-6.5mm}
		\caption{AndroZoo}
	\end{subfigure}
	
	\hspace{-0.15cm} 
	\begin{subfigure}[b]{0.15\textwidth}
		\begin{tikzpicture}[scale=0.45]
			\begin{axis}[
					xlabel={\(q\)},
					ylabel={AUC (\%)}, 
					xmin=0.1, xmax=0.9,
					ymin=0, ymax=100,
					xtick={0.2,0.4,0.6,0.8},
					ytick={0,20,40,60,80,100},
					ymajorgrids=false,
					legend pos=south west,
					legend style={nodes={scale=0.5, transform shape}},
					cycle list name=color list,
					legend to name=grayboxfullauclegend,
					legend columns=8,
                    height=3.6cm,
                    width=6.2cm
				]

				\addplot[color=purple,mark=diamond]
				coordinates {
					(0.1,48.3)(0.2,48.7)(0.3,48.6)(0.4,48.1)(0.5,48.3)(0.6,48.6)(0.7,47.4)(0.8,47.6)(0.9,47.2)

				};
				\addlegendentry{DeepMTD}
						
				\addplot[color=cyan,mark=pentagon]
				coordinates {
					(0.1,87.1)(0.2,76.7)(0.3,68.4)(0.4,60)(0.5,54.9)(0.6,48.5)(0.7,45.5)(0.8,41.3)(0.9,36.8)

				};
				\addlegendentry{Morphence}

				\addplot[color=red,mark=+]
				coordinates {
					(0.1,86.9)(0.2,76.5)(0.3,67)(0.4,58.6)(0.5,53.2)(0.6,46.7)(0.7,43.4)(0.8,38.5)(0.9,35)

				};
				\addlegendentry{MTDeep}
				
				\addplot[color=green,mark=x]
				coordinates {
					(0.1,98.2)(0.2,96.8)(0.3,76.3)(0.4,72.9)(0.5,72.3)(0.6,65.9)(0.7,64.2)(0.8,62.7)(0.9,60.2)

				};
				\addlegendentry{StratDef}
				
				\addplot[color=pink,mark=oplus]
				coordinates {
					(0.1,93.9)(0.2,89.3)(0.3,84.5)(0.4,81)(0.5,79.4)(0.6,75.5)(0.7,73)(0.8,71.8)(0.9,70.3)

				};
				\addlegendentry{Voting (Majority)}
				
				\addplot[color=brown,mark=triangle]
				coordinates {
					(0.1,88.7)(0.2,82.6)(0.3,75.5)(0.4,70.8)(0.5,69)(0.6,62.9)(0.7,61)(0.8,59)(0.9,55.8)

				};
				\addlegendentry{Voting (Veto)}
            \addplot[color=yellow]
                coordinates {
					(0.1,87.2)(0.2,76.9)(0.3,68.4)(0.4,60)(0.5,54.8)(0.6,48.6)(0.7,45.7)(0.8,41)(0.9,36.6)

				};
            \addlegendentry{NN-AT}

			\end{axis}
		\end{tikzpicture}
		\vspace{-6.5mm}
		\caption{DREBIN}
	\end{subfigure}
	\hspace{0.1cm} 
	\begin{subfigure}[b]{0.15\textwidth}
		\begin{tikzpicture}[scale=0.45]
			\begin{axis}[
					xlabel={\(q\)},
					ylabel={AUC (\%)}, 
					xmin=0.1, xmax=0.9,
					ymin=0, ymax=100,
					xtick={0.2,0.4,0.6,0.8},
					ytick={0,20,40,60,80,100},
					ymajorgrids=false,
					legend pos=south west,
					legend style={nodes={scale=0.5, transform shape}},
					cycle list name=color list,
					legend to name=grayboxfullauclegend,
					legend columns=8,
                    height=3.6cm,
                    width=6.2cm
				]
				
				\addplot[color=purple,mark=diamond]
				coordinates {
					(0.1,59)(0.2,58.9)(0.3,59.9)(0.4,60.2)(0.5,59.1)(0.6,60.2)(0.7,61.9)(0.8,59.2)(0.9,58)

				};
				\addlegendentry{DeepMTD}
						
				\addplot[color=cyan,mark=pentagon]
				coordinates {
					(0.1,98.2)(0.2,98.5)(0.3,98.8)(0.4,99.2)(0.5,99.3)(0.6,99.5)(0.7,99.5)(0.8,99.8)(0.9,99.9)

				};
				\addlegendentry{Morphence}

				\addplot[color=red,mark=+]
				coordinates {
					(0.1,82.7)(0.2,71.2)(0.3,60.4)(0.4,51.2)(0.5,42.9)(0.6,36.4)(0.7,30.8)(0.8,25.4)(0.9,20.4)

				};
				\addlegendentry{MTDeep}
				
				\addplot[color=green,mark=x]
				coordinates {
					(0.1,98.6)(0.2,98.9)(0.3,99)(0.4,99.2)(0.5,99.4)(0.6,99.5)(0.7,99.7)(0.8,99.8)(0.9,99.9)

				};
				\addlegendentry{StratDef}
				
				\addplot[color=pink,mark=oplus]
				coordinates {
					(0.1,92.8)(0.2,93.3)(0.3,93)(0.4,93)(0.5,92.4)(0.6,91.8)(0.7,91.6)(0.8,94.3)(0.9,93)

				};
				\addlegendentry{Voting (Majority)}
				
				\addplot[color=brown,mark=triangle]
				coordinates {
					(0.1,96.4)(0.2,96.7)(0.3,97)(0.4,97.3)(0.5,97.3)(0.6,97.6)(0.7,97.7)(0.8,97.8)(0.9,98.1)

				};
				\addlegendentry{Voting (Veto)}
				
        \addplot[color=yellow]
            coordinates {
				(0.1,98)(0.2,98.5)(0.3,98.7)(0.4,99.1)(0.5,99.2)(0.6,99.5)(0.7,99.5)(0.8,99.7)(0.9,99.7)

			};
        \addlegendentry{NN-AT}

			\end{axis}
		\end{tikzpicture}
		\vspace{-6.5mm}
		\caption{SLEIPNIR}
	\end{subfigure}
	\hspace{0.1cm} 
	\begin{subfigure}[b]{0.15\textwidth}
		\begin{tikzpicture}[scale=0.45]
			\begin{axis}[
					xlabel={\(q\)},
					ylabel={AUC (\%)}, 
					xmin=0.1, xmax=0.9,
					ymin=0, ymax=100,
					xtick={0.2,0.4,0.6,0.8},
					ytick={0,20,40,60,80,100},
					ymajorgrids=false,
					legend pos=south west,
					legend style={nodes={scale=0.5, transform shape}},
					cycle list name=color list,
					legend to name=androzooandrozoograyboxfullauclegend,
					legend columns=8,
                    height=3.6cm,
                    width=6.2cm
				]

				\addplot[color=purple,mark=diamond]
				coordinates {
					(0.1,47.6)(0.2,47.7)(0.3,48)(0.4,48.6)(0.5,48.1)(0.6,48.2)(0.7,48.3)(0.8,48.3)(0.9,48.3)

				};
				\addlegendentry{DeepMTD}
						
				\addplot[color=cyan,mark=pentagon]
				coordinates {
					(0.1,85.7)(0.2,74.2)(0.3,64.3)(0.4,56.5)(0.5,50.4)(0.6,44.8)(0.7,38.5)(0.8,33)(0.9,29.5)

				};
				\addlegendentry{Morphence}

				\addplot[color=red,mark=+]
				coordinates {
					(0.1,85.6)(0.2,74.1)(0.3,64.2)(0.4,56.2)(0.5,50)(0.6,44.4)(0.7,38.3)(0.8,32.8)(0.9,29.3)

				};
				\addlegendentry{MTDeep}
				
				\addplot[color=green,mark=x]
				coordinates {
					(0.1,99.1)(0.2,99.1)(0.3,99.2)(0.4,99.3)(0.5,99.4)(0.6,99.5)(0.7,99.6)(0.8,99.7)(0.9,99.9)

				};
				\addlegendentry{StratDef}
				
				\addplot[color=pink,mark=oplus]
				coordinates {
					(0.1,98.1)(0.2,98.3)(0.3,98.5)(0.4,98.4)(0.5,98.4)(0.6,98.6)(0.7,98.8)(0.8,98.8)(0.9,98.8)

				};
				\addlegendentry{Voting (Majority)}
				
				\addplot[color=brown,mark=triangle]
				coordinates {
					(0.1,97)(0.2,97.4)(0.3,97.8)(0.4,97.7)(0.5,97.9)(0.6,98.1)(0.7,98.4)(0.8,98.9)(0.9,98.6)

				};
				\addlegendentry{Voting (Veto)}
            \addplot[color=yellow]
                coordinates {
					(0.1,98.6)(0.2,98.9)(0.3,99.1)(0.4,99.2)(0.5,99.3)(0.6,99.6)(0.7,99.7)(0.8,99.7)(0.9,99.9)

				};
            \addlegendentry{NN-AT}

			\end{axis}
		\end{tikzpicture}
		\vspace{-6.5mm}
		\caption{AndroZoo}
	\end{subfigure}
	\ref{grayboxfullauclegend}
	\caption{F1 \& AUC vs. $q$.}
	\label{figure:grayboxfullauc}
\end{figure}
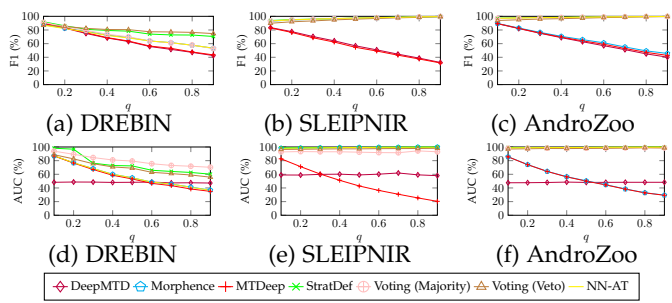

\vspace{-2mm}

\section{Varying Training Data Size in Gray-box Transferability Attack (Section~\ref{sec:grayboxevaltransfattack})}
\label{appendix:additionalgrayboxtransf}
\vspace{-1.5mm}
We evaluate how the attack performs when the substitute models are trained with different sizes of training data (\% of the original size). For DREBIN, the average evasion rate decreases between 25\% to 75\% sizes, after which it increases once 100\% of the training data is used to construct substitute models. Meanwhile, for SLEIPNIR, we observe peak attack performance at 25\% training data, with this decreasing as the training set size increases. The evasion rate against DeepMTD and MTDeep remains consistent, with only the performance against other models decreasing dramatically. This largely supports the idea that overfitting may be occurring. The disparity between the trends in DREBIN and SLEIPNIR can be attributed to the differences and nuances of each dataset relating to the different-sized attack surfaces of each. The black-box transferability attack still performs better. We primarily attribute this phenomenon to the suggestions made earlier in the paper: the substitute models capture the oracle's traits and behavior better, which means that adversarial examples for the black-box substitute models transfer better to the oracle.

\begin{figure}[!htbp]
	\centering
	\begin{subfigure}[b]{0.25\textwidth}
		\begin{tikzpicture}[scale=0.45]
			\begin{axis}[
					ybar,
					enlarge y limits=false,
					width=9cm, height=3.6cm,
					ymin=0, ymax=100,
					ylabel={Evasion rate (\%)},
					symbolic x coords={Our Diverse Ensemble (\(\Theta_{t} = 0.75\)), Our Diverse Ensemble (\(\Theta_{t} = 0.25\)), Single DNN},
					xticklabels={Our Diverse Ensemble (\(\Theta_{t} = 0.75\)), Our Diverse Ensemble (\(\Theta_{t} = 0.25\)), Single DNN (Papernot et al.)},
					xtick=data,
					xticklabel style={text width=2.5cm,align=center},
					nodes near coords align={horizontal},
					legend pos=north east,
					legend style={nodes={scale=0.5, transform shape}},
					legend to name=grayboxevasionratesextendedleg,
					legend columns=8,
					bar width=0.15cm,
					enlarge x limits={abs=1cm}
				]
				\addplot [fill=purple]
				coordinates {
					(Our Diverse Ensemble (\(\Theta_{t} = 0.75\)), 83.3)
(Our Diverse Ensemble (\(\Theta_{t} = 0.25\)), 17.3)
(Single DNN, 19.2)

				};
				\addplot [fill=cyan]
				coordinates {
					(Our Diverse Ensemble (\(\Theta_{t} = 0.75\)), 79.6)
(Our Diverse Ensemble (\(\Theta_{t} = 0.25\)), 11.4)
(Single DNN, 13.8)

				};
				\addplot [fill=red]
				coordinates {
					(Our Diverse Ensemble (\(\Theta_{t} = 0.75\)), 79.6)
(Our Diverse Ensemble (\(\Theta_{t} = 0.25\)), 12)
(Single DNN, 13.8)

				};
				
				\addplot [fill=green]
				coordinates {
                    (Our Diverse Ensemble (\(\Theta_{t} = 0.75\)), 85.2)
(Our Diverse Ensemble (\(\Theta_{t} = 0.25\)), 21.8)
(Single DNN, 23.6)

				};
				\addplot [fill=pink]
				coordinates {
                    (Our Diverse Ensemble (\(\Theta_{t} = 0.75\)), 87)
(Our Diverse Ensemble (\(\Theta_{t} = 0.25\)), 14.6)
(Single DNN, 18.7)

				};
				\addplot [fill=brown]
				coordinates {
                    (Our Diverse Ensemble (\(\Theta_{t} = 0.75\)), 63)
(Our Diverse Ensemble (\(\Theta_{t} = 0.25\)), 9)
(Single DNN, 9.9)

				};
				
                \addplot [fill=yellow]
				coordinates {
					(Our Diverse Ensemble (\(\Theta_{t} = 0.75\)), 77.8)
(Our Diverse Ensemble (\(\Theta_{t} = 0.25\)), 11.2)
(Single DNN, 13.3)
				};

				\legend{DeepMTD, Morphence, MTDeep, StratDef, Voting (Majority), Voting (Veto), NN-AT}%
			\end{axis}
				   
		\end{tikzpicture}
		\caption{25\% size, DREBIN}
	\end{subfigure}
	\begin{subfigure}[b]{0.22\textwidth}
		\begin{tikzpicture}[scale=0.45]
			\begin{axis}[
					ybar,
					enlarge y limits=false,
					width=9cm, height=3.6cm,
					ymin=0, ymax=100,
					ylabel={Evasion rate (\%)},
					symbolic x coords={Our Diverse Ensemble (\(\Theta_{t} = 0.75\)), Our Diverse Ensemble (\(\Theta_{t} = 0.25\)), Single DNN},
					xticklabels={Our Diverse Ensemble (\(\Theta_{t} = 0.75\)), Our Diverse Ensemble (\(\Theta_{t} = 0.25\)), Single DNN (Papernot et al.)},
					xtick=data,
					xticklabel style={text width=2.5cm,align=center},
					nodes near coords align={horizontal},
					legend pos=north east,
					legend style={nodes={scale=0.5, transform shape}},
					legend to name=grayboxevasionratesextendedleg,
					legend columns=8,
					bar width=0.15cm,
					enlarge x limits={abs=1cm}
				]
				\addplot [fill=purple]
				coordinates {
					(Our Diverse Ensemble (\(\Theta_{t} = 0.75\)), 59.3)
	(Our Diverse Ensemble (\(\Theta_{t} = 0.25\)), 17.6)
	(Single DNN, 22.5)
	
				};
				\addplot [fill=cyan]
				coordinates {
					(Our Diverse Ensemble (\(\Theta_{t} = 0.75\)), 40.7)
	(Our Diverse Ensemble (\(\Theta_{t} = 0.25\)), 8.8)
	(Single DNN, 11.7)
	
				};
				\addplot [fill=red]
				coordinates {
					(Our Diverse Ensemble (\(\Theta_{t} = 0.75\)), 50)
	(Our Diverse Ensemble (\(\Theta_{t} = 0.25\)), 10.5)
	(Single DNN, 13.9)
	
				};
				
				\addplot [fill=green]
				coordinates {
					(Our Diverse Ensemble (\(\Theta_{t} = 0.75\)), 66.3)
	(Our Diverse Ensemble (\(\Theta_{t} = 0.25\)), 19.9)
	(Single DNN, 21.6)
	
				};
				\addplot [fill=pink]
				coordinates {
					(Our Diverse Ensemble (\(\Theta_{t} = 0.75\)), 64)
	(Our Diverse Ensemble (\(\Theta_{t} = 0.25\)), 13.7)
	(Single DNN, 18.2)
	
				};
				\addplot [fill=brown]
				coordinates {
					(Our Diverse Ensemble (\(\Theta_{t} = 0.75\)), 31.4)
	(Our Diverse Ensemble (\(\Theta_{t} = 0.25\)), 6.6)
	(Single DNN, 7.8)
	
				};
				
				\addplot [fill=yellow]
				coordinates {
					(Our Diverse Ensemble (\(\Theta_{t} = 0.75\)), 40.7)
	(Our Diverse Ensemble (\(\Theta_{t} = 0.25\)), 9.3)
	(Single DNN, 12.6)

				};

				\legend{DeepMTD, Morphence, MTDeep, StratDef, Voting (Majority), Voting (Veto), NN-AT}%
			\end{axis}
				   
		\end{tikzpicture}
		\caption{50\% size, DREBIN}
	\end{subfigure}
	\begin{subfigure}[b]{0.25\textwidth}
		\begin{tikzpicture}[scale=0.45]
			\begin{axis}[
					ybar,
					enlarge y limits=false,
					width=9cm, height=3.6cm,
					ymin=0, ymax=100,
					ylabel={Evasion rate (\%)},
					symbolic x coords={Our Diverse Ensemble (\(\Theta_{t} = 0.75\)), Our Diverse Ensemble (\(\Theta_{t} = 0.25\)), Single DNN},
					xticklabels={Our Diverse Ensemble (\(\Theta_{t} = 0.75\)), Our Diverse Ensemble (\(\Theta_{t} = 0.25\)), Single DNN (Papernot et al.)},
					xtick=data,
					xticklabel style={text width=2.5cm,align=center},
					nodes near coords align={horizontal},
					legend pos=north east,
					legend style={nodes={scale=0.5, transform shape}},
					legend to name=grayboxevasionratesextendedleg,
					legend columns=8,
					bar width=0.15cm,
					enlarge x limits={abs=1cm}
				]
				\addplot [fill=purple]
				coordinates {
					(Our Diverse Ensemble (\(\Theta_{t} = 0.75\)), 53.8)
	(Our Diverse Ensemble (\(\Theta_{t} = 0.25\)), 14.9)
	(Single DNN, 17.1)
	
				};
				\addplot [fill=cyan]
				coordinates {
					(Our Diverse Ensemble (\(\Theta_{t} = 0.75\)), 45)
	(Our Diverse Ensemble (\(\Theta_{t} = 0.25\)), 9.6)
	(Single DNN, 13)
	
				};
				\addplot [fill=red]
				coordinates {
					(Our Diverse Ensemble (\(\Theta_{t} = 0.75\)), 52.5)
	(Our Diverse Ensemble (\(\Theta_{t} = 0.25\)), 10.6)
	(Single DNN, 13.9)
	
				};
				
				\addplot [fill=green]
				coordinates {
					(Our Diverse Ensemble (\(\Theta_{t} = 0.75\)), 61.3)
	(Our Diverse Ensemble (\(\Theta_{t} = 0.25\)), 18.2)
	(Single DNN, 19.9)
	
				};
				\addplot [fill=pink]
				coordinates {
					(Our Diverse Ensemble (\(\Theta_{t} = 0.75\)), 60)
	(Our Diverse Ensemble (\(\Theta_{t} = 0.25\)), 12.2)
	(Single DNN, 15.3)
	
				};
				\addplot [fill=brown]
				coordinates {
					(Our Diverse Ensemble (\(\Theta_{t} = 0.75\)), 33.8)
	(Our Diverse Ensemble (\(\Theta_{t} = 0.25\)), 6.8)
	(Single DNN, 8.8)
	
				};
				
				\addplot [fill=yellow]
				coordinates {
					(Our Diverse Ensemble (\(\Theta_{t} = 0.75\)), 45)
	(Our Diverse Ensemble (\(\Theta_{t} = 0.25\)), 9.4)
	(Single DNN, 13.4)

				};

				\legend{DeepMTD, Morphence, MTDeep, StratDef, Voting (Majority), Voting (Veto), NN-AT}%
			\end{axis}
				   
		\end{tikzpicture}
		\caption{75\% size, DREBIN}
	\end{subfigure}
	\begin{subfigure}[b]{0.22\textwidth}
		\begin{tikzpicture}[scale=0.45]
			\begin{axis}[
					ybar,
					enlarge y limits=false,
					width=9cm, height=3.6cm,
					ymin=0, ymax=100,
					ylabel={Evasion rate (\%)},
					symbolic x coords={Our Diverse Ensemble (\(\Theta_{t} = 0.75\)), Our Diverse Ensemble (\(\Theta_{t} = 0.25\)), Single DNN},
					xticklabels={Our Diverse Ensemble (\(\Theta_{t} = 0.75\)), Our Diverse Ensemble (\(\Theta_{t} = 0.25\)), Single DNN (Papernot et al.)},
					xtick=data,
					xticklabel style={text width=2.5cm,align=center},
					nodes near coords align={horizontal},
					legend pos=north east,
					legend style={nodes={scale=0.5, transform shape}},
					legend to name=grayboxevasionratesextendedleg,
					legend columns=8,
					bar width=0.15cm,
					enlarge x limits={abs=1cm}
				]
				\addplot [fill=purple]
				coordinates {
					(Our Diverse Ensemble (\(\Theta_{t} = 0.75\)), 95.1)
(Our Diverse Ensemble (\(\Theta_{t} = 0.25\)), 54.9)
(Single DNN, 55.6)	
				};
				\addplot [fill=cyan]
				coordinates {
					(Our Diverse Ensemble (\(\Theta_{t} = 0.75\)), 91.3)
(Our Diverse Ensemble (\(\Theta_{t} = 0.25\)), 50.5)
(Single DNN, 50.4)	
				};
				\addplot [fill=red]
				coordinates {
					(Our Diverse Ensemble (\(\Theta_{t} = 0.75\)), 92.7)
(Our Diverse Ensemble (\(\Theta_{t} = 0.25\)), 51.4)
(Single DNN, 51.8)
				};
				
				\addplot [fill=green]
				coordinates {
				(Our Diverse Ensemble (\(\Theta_{t} = 0.75\)), 64.5)
(Our Diverse Ensemble (\(\Theta_{t} = 0.25\)), 39.5)
(Single DNN, 43.5)	
				};
				\addplot [fill=pink]
				coordinates {
					(Our Diverse Ensemble (\(\Theta_{t} = 0.75\)), 94.1)
(Our Diverse Ensemble (\(\Theta_{t} = 0.25\)), 52.3)
(Single DNN, 52.7)	
				};
				\addplot [fill=brown]
				coordinates {
					(Our Diverse Ensemble (\(\Theta_{t} = 0.75\)), 57.8)
(Our Diverse Ensemble (\(\Theta_{t} = 0.25\)), 32)
(Single DNN, 34.8)	
				};

                \addplot [fill=yellow]
				coordinates {
					(Our Diverse Ensemble (\(\Theta_{t} = 0.75\)), 90.467)
(Our Diverse Ensemble (\(\Theta_{t} = 0.25\)), 50.056)
(Single DNN, 49.777)

				};

				\legend{DeepMTD, Morphence, MTDeep, StratDef, Voting (Majority), Voting (Veto), NN-AT}%
			\end{axis}
				   
		\end{tikzpicture}
		\caption{100\% size, DREBIN}
	\end{subfigure}
	\begin{subfigure}[b]{0.25\textwidth}
		\begin{tikzpicture}[scale=0.45]
			\begin{axis}[
					ybar,
					enlarge y limits=false,
					width=9cm, height=3.6cm,
					ymin=0, ymax=100,
					ylabel={Evasion rate (\%)},
					symbolic x coords={Our Diverse Ensemble (\(\Theta_{t} = 0.75\)), Our Diverse Ensemble (\(\Theta_{t} = 0.25\)), Single DNN},
					xticklabels={Our Diverse Ensemble (\(\Theta_{t} = 0.75\)), Our Diverse Ensemble (\(\Theta_{t} = 0.25\)), Single DNN (Papernot et al.)},
					xtick=data,
					xticklabel style={text width=2.5cm,align=center},
					nodes near coords align={horizontal},
					legend pos=north east,
					legend style={nodes={scale=0.5, transform shape}},
					legend to name=grayboxevasionratesextendedleg,
					legend columns=8,
					bar width=0.15cm,
					enlarge x limits={abs=1cm}
				]
				\addplot [fill=purple]
				coordinates {
                    (Our Diverse Ensemble (\(\Theta_{t} = 0.75\)), 95.5)
(Our Diverse Ensemble (\(\Theta_{t} = 0.25\)), 51)
(Single DNN, 51.6)

				};
				\addplot [fill=cyan]
				coordinates {
                    (Our Diverse Ensemble (\(\Theta_{t} = 0.75\)), 0.5)
(Our Diverse Ensemble (\(\Theta_{t} = 0.25\)), 0)
(Single DNN, 0)

				};
				\addplot [fill=red]
				coordinates {
                    (Our Diverse Ensemble (\(\Theta_{t} = 0.75\)), 96)
(Our Diverse Ensemble (\(\Theta_{t} = 0.25\)), 51.5)
(Single DNN, 51.6)

				};
				\addplot [fill=green]
				coordinates {
                    (Our Diverse Ensemble (\(\Theta_{t} = 0.75\)), 23.6)
(Our Diverse Ensemble (\(\Theta_{t} = 0.25\)), 12.9)
(Single DNN, 1.4)

				};
				\addplot [fill=pink]
				coordinates {
                    (Our Diverse Ensemble (\(\Theta_{t} = 0.75\)), 18.6)
(Our Diverse Ensemble (\(\Theta_{t} = 0.25\)), 9.7)
(Single DNN, 0)

				};
				\addplot [fill=brown]
				coordinates {
                    (Our Diverse Ensemble (\(\Theta_{t} = 0.75\)), 0)
(Our Diverse Ensemble (\(\Theta_{t} = 0.25\)), 0)
(Single DNN, 0)

				};
				
                \addplot [fill=yellow]
				coordinates {
                    (Our Diverse Ensemble (\(\Theta_{t} = 0.75\)), 44.2)
(Our Diverse Ensemble (\(\Theta_{t} = 0.25\)), 23.8)
(Single DNN, 2.3)
				};

				\legend{DeepMTD, Morphence, MTDeep, StratDef, Voting (Majority), Voting (Veto), NN-AT}%
			\end{axis}
				   
		\end{tikzpicture}
		\caption{25\% size, SLEIPNIR}
	\end{subfigure}
	\begin{subfigure}[b]{0.22\textwidth}
		\begin{tikzpicture}[scale=0.45]
			\begin{axis}[
					ybar,
					enlarge y limits=false,
					width=9cm, height=3.6cm,
					ymin=0, ymax=100,
					ylabel={Evasion rate (\%)},
					symbolic x coords={Our Diverse Ensemble (\(\Theta_{t} = 0.75\)), Our Diverse Ensemble (\(\Theta_{t} = 0.25\)), Single DNN},
					xticklabels={Our Diverse Ensemble (\(\Theta_{t} = 0.75\)), Our Diverse Ensemble (\(\Theta_{t} = 0.25\)), Single DNN (Papernot et al.)},
					xtick=data,
					xticklabel style={text width=2.5cm,align=center},
					nodes near coords align={horizontal},
					legend pos=north east,
					legend style={nodes={scale=0.5, transform shape}},
					legend to name=grayboxevasionratesextendedleg,
					legend columns=8,
					bar width=0.15cm,
					enlarge x limits={abs=1cm}
				]
				\addplot [fill=purple]
				coordinates {
					(Our Diverse Ensemble (\(\Theta_{t} = 0.75\)), 93.1)
	(Our Diverse Ensemble (\(\Theta_{t} = 0.25\)), 48)
	(Single DNN, 45.2)
	
				};
				\addplot [fill=cyan]
				coordinates {
					(Our Diverse Ensemble (\(\Theta_{t} = 0.75\)), 0)
	(Our Diverse Ensemble (\(\Theta_{t} = 0.25\)), 0)
	(Single DNN, 0)
	
				};
				\addplot [fill=red]
				coordinates {
					(Our Diverse Ensemble (\(\Theta_{t} = 0.75\)), 93.1)
	(Our Diverse Ensemble (\(\Theta_{t} = 0.25\)), 48.2)
	(Single DNN, 45.2)
	
				};
				\addplot [fill=green]
				coordinates {
					(Our Diverse Ensemble (\(\Theta_{t} = 0.75\)), 0.5)
	(Our Diverse Ensemble (\(\Theta_{t} = 0.25\)), 0.2)
	(Single DNN, 0)
	
				};
				\addplot [fill=pink]
				coordinates {
					(Our Diverse Ensemble (\(\Theta_{t} = 0.75\)), 0)
	(Our Diverse Ensemble (\(\Theta_{t} = 0.25\)), 0)
	(Single DNN, 0)
	
				};
				\addplot [fill=brown]
				coordinates {
					(Our Diverse Ensemble (\(\Theta_{t} = 0.75\)), 0)
	(Our Diverse Ensemble (\(\Theta_{t} = 0.25\)), 0)
	(Single DNN, 0)
	
				};
				
				\addplot [fill=yellow]
				coordinates {
					(Our Diverse Ensemble (\(\Theta_{t} = 0.75\)), 0.5)
	(Our Diverse Ensemble (\(\Theta_{t} = 0.25\)), 0.2)
	(Single DNN, 0.4)

				};

				\legend{DeepMTD, Morphence, MTDeep, StratDef, Voting (Majority), Voting (Veto), NN-AT}%
			\end{axis}
				   
		\end{tikzpicture}
		\caption{50\% size, SLEIPNIR}
	\end{subfigure}
	\begin{subfigure}[b]{0.25\textwidth}
		\begin{tikzpicture}[scale=0.45]
			\begin{axis}[
					ybar,
					enlarge y limits=false,
					width=9cm, height=3.6cm,
					ymin=0, ymax=100,
					ylabel={Evasion rate (\%)},
					symbolic x coords={Our Diverse Ensemble (\(\Theta_{t} = 0.75\)), Our Diverse Ensemble (\(\Theta_{t} = 0.25\)), Single DNN},
					xticklabels={Our Diverse Ensemble (\(\Theta_{t} = 0.75\)), Our Diverse Ensemble (\(\Theta_{t} = 0.25\)), Single DNN (Papernot et al.)},
					xtick=data,
					xticklabel style={text width=2.5cm,align=center},
					nodes near coords align={horizontal},
					legend pos=north east,
					legend style={nodes={scale=0.5, transform shape}},
					legend to name=grayboxevasionratesextendedleg,
					legend columns=8,
					bar width=0.15cm,
					enlarge x limits={abs=1cm}
				]
				\addplot [fill=purple]
				coordinates {
					(Our Diverse Ensemble (\(\Theta_{t} = 0.75\)), 96.3)
	(Our Diverse Ensemble (\(\Theta_{t} = 0.25\)), 52.6)
	(Single DNN, 53.2)
	
				};
				\addplot [fill=cyan]
				coordinates {
					(Our Diverse Ensemble (\(\Theta_{t} = 0.75\)), 1.4)
	(Our Diverse Ensemble (\(\Theta_{t} = 0.25\)), 0)
	(Single DNN, 0)
	
				};
				\addplot [fill=red]
				coordinates {
					(Our Diverse Ensemble (\(\Theta_{t} = 0.75\)), 95.3)
	(Our Diverse Ensemble (\(\Theta_{t} = 0.25\)), 52.9)
	(Single DNN, 53.2)
	
				};
				\addplot [fill=green]
				coordinates {
					(Our Diverse Ensemble (\(\Theta_{t} = 0.75\)), 2.8)
	(Our Diverse Ensemble (\(\Theta_{t} = 0.25\)), 1.7)
	(Single DNN, 4.1)
	
				};
				\addplot [fill=pink]
				coordinates {
					(Our Diverse Ensemble (\(\Theta_{t} = 0.75\)), 1.4)
	(Our Diverse Ensemble (\(\Theta_{t} = 0.25\)), 0.7)
	(Single DNN, 1.4)
	
				};
				\addplot [fill=brown]
				coordinates {
					(Our Diverse Ensemble (\(\Theta_{t} = 0.75\)), 0)
	(Our Diverse Ensemble (\(\Theta_{t} = 0.25\)), 0)
	(Single DNN, 0)
	
				};
				
				\addplot [fill=yellow]
				coordinates {
					(Our Diverse Ensemble (\(\Theta_{t} = 0.75\)), 3.3)
	(Our Diverse Ensemble (\(\Theta_{t} = 0.25\)), 2.4)
	(Single DNN, 4.5)

				};

				\legend{DeepMTD, Morphence, MTDeep, StratDef, Voting (Majority), Voting (Veto), NN-AT}%
			\end{axis}
				   
		\end{tikzpicture}
		\caption{75\% size, SLEIPNIR}
	\end{subfigure}
	\begin{subfigure}[b]{0.22\textwidth}
		\begin{tikzpicture}[scale=0.45]
			\begin{axis}[
					ybar,
					enlarge y limits=false,
					width=9cm, height=3.6cm,
					ymin=0, ymax=100,
					ylabel={Evasion rate (\%)},
					symbolic x coords={Our Diverse Ensemble (\(\Theta_{t} = 0.75\)), Our Diverse Ensemble (\(\Theta_{t} = 0.25\)), Single DNN},
					xticklabels={Our Diverse Ensemble (\(\Theta_{t} = 0.75\)), Our Diverse Ensemble (\(\Theta_{t} = 0.25\)), Single DNN (Papernot et al.)},
					xtick=data,
					xticklabel style={text width=2.5cm,align=center},
					nodes near coords align={horizontal},
					legend pos=north east,
					legend style={nodes={scale=0.5, transform shape}},
					legend to name=grayboxevasionratesextendedleg,
					legend columns=8,
					bar width=0.15cm,
					enlarge x limits={abs=1cm}
				]
				\addplot [fill=purple]
				coordinates {
				(Our Diverse Ensemble (\(\Theta_{t} = 0.75\)), 96.8)
                (Our Diverse Ensemble (\(\Theta_{t} = 0.25\)), 55.3)
                (Single DNN, 57.3)
				};
				\addplot [fill=cyan]
				coordinates {
				(Our Diverse Ensemble (\(\Theta_{t} = 0.75\)), 0)
                (Our Diverse Ensemble (\(\Theta_{t} = 0.25\)), 0)
                (Single DNN, 0)
				};
				\addplot [fill=red]
				coordinates {
				(Our Diverse Ensemble (\(\Theta_{t} = 0.75\)), 96.5)
                (Our Diverse Ensemble (\(\Theta_{t} = 0.25\)), 55.4)
                (Single DNN, 56.3)
				};
				\addplot [fill=green]
				coordinates {
				(Our Diverse Ensemble (\(\Theta_{t} = 0.75\)), 0)
                (Our Diverse Ensemble (\(\Theta_{t} = 0.25\)), 0)
                (Single DNN, 0)
				};
				\addplot [fill=pink]
				coordinates {
					(Our Diverse Ensemble (\(\Theta_{t} = 0.75\)), 0)
                    (Our Diverse Ensemble (\(\Theta_{t} = 0.25\)), 0.2)
                    (Single DNN, 0)
				};
				\addplot [fill=brown]
				coordinates {
				(Our Diverse Ensemble (\(\Theta_{t} = 0.75\)), 0)
                (Our Diverse Ensemble (\(\Theta_{t} = 0.25\)), 0)
                (Single DNN, 0)
				};
				
                \addplot [fill=yellow]
				coordinates {
					(Our Diverse Ensemble (\(\Theta_{t} = 0.75\)), 0)
                    (Our Diverse Ensemble (\(\Theta_{t} = 0.25\)), 0)
                    (Single DNN, 0)

				};

				\legend{DeepMTD, Morphence, MTDeep, StratDef, Voting (Majority), Voting (Veto), NN-AT}%
			\end{axis}
				   
		\end{tikzpicture}
		\caption{100\% size, SLEIPNIR}
	\end{subfigure}
	\ref{grayboxevasionratesextendedleg}
	\caption{Evasion rate vs. sizes of training data.}
	\label{figure:grayboxevasionratesleipnirextended}
\end{figure}
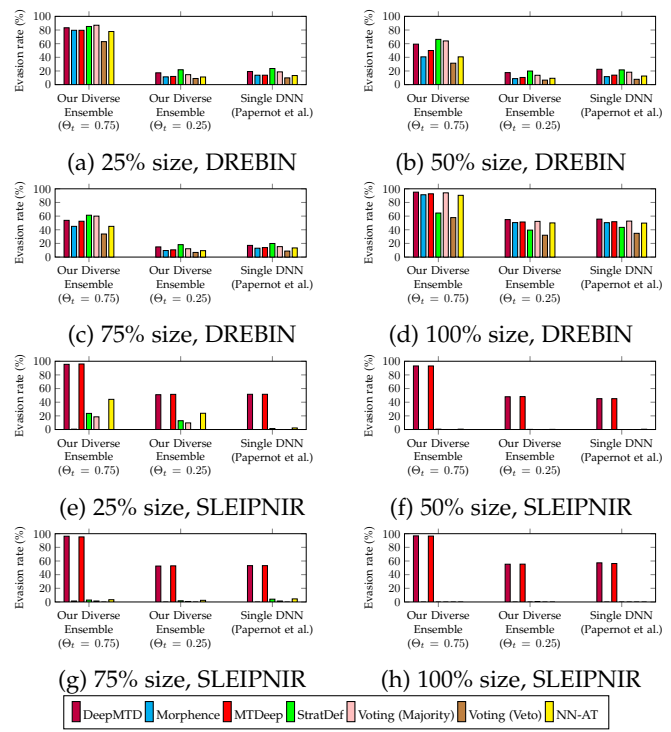

\end{appendices}

\end{document}